\documentclass[11pt]{article}
\RequirePackage{amsthm,amsmath,amsfonts,amssymb}
\RequirePackage[colorlinks,citecolor=blue,urlcolor=blue]{hyperref}
\RequirePackage{graphicx}
\RequirePackage[authoryear]{natbib}

\usepackage{fullpage}

\usepackage{url}
\usepackage{algpseudocode}
\usepackage{algorithm}
\usepackage{caption}
\usepackage{subcaption}
\usepackage[title]{appendix}
\usepackage{multirow}
\usepackage{colortbl}
\usepackage{xcolor}
\usepackage{enumitem}
\usepackage[ruled,vlined,algo2e]{algorithm2e}

\newtheorem{theorem}{Theorem}[section]
\newtheorem{corollary}[theorem]{Corollary}

\newtheorem{lemma}[theorem]{Lemma}
\newtheorem{assumption}{Assumption}
\newtheorem{remark}{Remark}
\DeclareMathOperator*{\argmax}{arg\,max}
\DeclareMathOperator*{\argmin}{arg\,min}
\newcommand\numberthis{\addtocounter{equation}{1}\tag{\theequation}}

\newcommand\blfootnote[1]{
  \begingroup
  \renewcommand\thefootnote{}\footnote{#1}
  \addtocounter{footnote}{-1}
  \endgroup
}

\definecolor{DSgray}{cmyk}{0,1,0,0}

\newcommand{\Vhat}{\widehat{V}}
\newcommand{\Uhat}{\widehat{U}}
\newcommand{\bU}{U}
\newcommand{\bV}{V}
\newcommand{\Usgd}{\mathcal{U}}
\newcommand{\Vsgd}{\mathcal{V}}
\newcommand{\Msgd}{\widehat{M}^{\mathbf{sgd}}}
\newcommand{\Mlr}{\widehat{M}^{\mathbf{proj}}}

\newcommand{\Uorg}{U_{\bot}}
\newcommand{\Vorg}{V_{\bot}}
\newcommand{\bX}{X}

\newcommand{\Minit}{\widehat{M}^{\mathbf{init}}}
\newcommand{\Munbs}{\widehat{M}^{\mathbf{unbs}}}

\newcommand{\Vsp}{\widehat{\mathsf{V}}}
\newcommand{\Usp}{\widehat{\mathsf{U}}}
\newcommand{\bP}{\mathbb{P}}
\newcommand{\bE}{\mathbb{E}}
\newcommand{\pmin}{p_0}

\newcommand{\Df}{\Delta_{\mathrm{diff}}}

\begin{document}

\title{Online Statistical Inference in Decision-Making with Matrix Context}
\author{Qiyu Han \and Will Wei Sun \and Yichen Zhang\blfootnote{Daniels School of Business, Purdue University.}
}
\date{}
\maketitle
\setcounter{page}{1}
\vspace{-.5em}
\begin{abstract}
    The study of online decision-making problems that leverage contextual information has drawn notable attention due to their significant applications in fields ranging from healthcare to autonomous systems. In modern applications, contextual information can be rich and is often represented as a matrix. Moreover, while existing online decision algorithms mainly focus on reward maximization, less attention has been devoted to statistical inference. To address these gaps, in this work, we consider an online decision-making problem with a matrix context where the true model parameters have a low-rank structure. We propose a {\it fully online} procedure to conduct statistical inference with adaptively collected data. The low-rank structure of the model parameter and the adaptive nature of the data collection process make this difficult: standard low-rank estimators are biased and cannot be obtained in a sequential manner while existing inference approaches in sequential decision-making algorithms fail to account for the low-rankness and are also biased. To overcome these challenges, we introduce a new online debiasing procedure to simultaneously handle both sources of bias. Our inference framework encompasses both parameter inference and optimal policy value inference. In theory, we establish the asymptotic normality of the proposed online debiased estimators and prove the validity of the constructed confidence intervals for both inference tasks. Our inference results are built upon a newly developed low-rank stochastic gradient descent estimator and its convergence result, which are also of independent interest.
\end{abstract}

\noindent
{\it Keywords}: online inference, online decision-making, low-rank matrix, reinforcement learning, stochastic gradient descent.

\section{Introduction}
\label{sec: intro}

\begin{figure}[b]
    \centering
    \includegraphics[width = 0.9\linewidth]{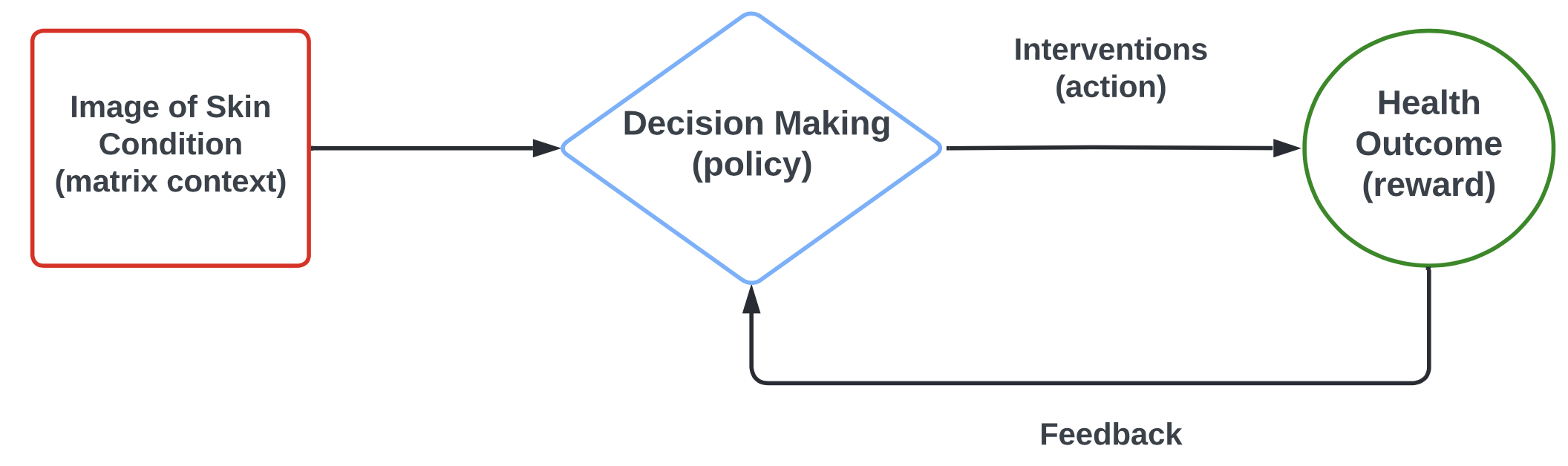}
    \caption{An illustration of our online decision-making framework with matrix context.}
    \label{fig:chicken pox}
\end{figure}

From personalized medicine to recommendation systems, exploiting personalized information in decision-making has gained popularity during the last decades \citep{kosorok2019precision,fang2022fairness, qi2022robustness}. In the widely studied framework of online decision-making with contextual information, decisions are sequentially made for users based on the current context and historical interactions \citep{li2010contextual,agrawal2013thompson,li2017provably, lattimore2020bandit}. In traditional settings, the context is typically formulated in a vector. However, contextual information in modern online decision-making problems is often in a matrix form. In the skin treatment example shown in Figure \ref{fig:chicken pox}, the decision-making policy determines whether an immediate intervention should be applied based on the patient's current image of skin condition (a matrix context) and the health outcomes of historical interventions \citep{akrout2019improving}. The inspiration for this example can be traced to the recently growing application of mobile Health, which targets to deliver immediate interventions, such as motivational messages, to individuals through mobile devices according to their current health condition \citep{istepanian2007m,deliu2022reinforcement}. In such examples, the context is an image that can be formulated as a matrix. The goal of the decision-making policy is to decide the best action at each time based on the current matrix context and all historical interactions. 

In this paper, we consider an online decision-making problem with matrix contexts. In particular, at time $t$, given a matrix context $X_t \in \mathbb{R}^{d_1 \times d_2}$, the policy takes an action $a_t \in \{ 0,1\}$ and observes a noisy reward $y_t \in \mathbb{R}$ as 
\begin{equation}
        \label{eq: model}
         y_t = a_t \langle M_1, \bX_t \rangle + (1-a_t)\langle M_0, \bX_t \rangle + \xi_t,
\end{equation}
where $\xi_t \in \mathbb{R}$ is the random noise and $ \left \langle M_i, X_t \right \rangle = tr(X_t^\top M_i)$, for $i \in \{0,1\}$, denotes the matrix inner product. The true matrix parameter $M_i$ is assumed to be of low rank with a rank $r \ll \min \{d_1,d_2 \}$. In our motivation example, a group of pixels in the image that form a region can impose a collaborative effect on describing the health outcome, allowing the matrix parameter to have a low-rank structure \citep{chen2019inference,xia2019confidence,xia2021statistical}. In addition, such a low-rank structure is crucial in online decision-making due to its high dimensionality compared to its limited sample size. In $(\ref{eq: model})$, when $a_t = 1$ (with intervention), the reward is given by $\langle M_1, X_t\rangle + \xi_t$ (health outcome with intervention); when $a_t = 0$ (without intervention), the reward is given by $\langle M_0, X_t\rangle + \xi_t$ (health outcome without intervention). Without loss of generality, our work mainly focuses on a binary action, i.e., $a_t \in \{ 0,1\}$ at each time $t$, and it can be easily extended to multiple actions in a discrete action space.

While existing sequential decision-making algorithms mainly focused on choosing the best action to maximize the cumulative reward \citep{li2010contextual,agrawal2013thompson,li2017provably, lattimore2020bandit}, less attention has been paid to statistical inference in sequential decision-making frameworks. In real-world applications, we are often not just interested in obtaining the point estimate of the reward function but also a measure of the statistical uncertainty associated with the estimate. This is especially relevant in fields such as personalized medicine, mobile health, and automated driving, where it is often risky to run a policy without a statistically sound estimate of its quality. For example, online randomized experiments like A/B testing have been widely conducted by technological/pharmaceutical companies to compare a new product with an old one. Recent studies \citep{li2021unifying, shi2021online, shi2022dynamic} have used various bandit or reinforcement learning methods to form sequential testing procedures. In these online evaluation tasks, it is important to quantify the uncertainty of the point estimate for constructing valid hypothesis testing. 

Statistical inference significantly enhances scientific knowledge by applying insights from prior experiments to improve future research designs, extending beyond the immediate objectives of in-experiment learning aimed at optimizing decision-making performance. This knowledge is crucial for capturing the extensive, long-term consequences of actions and associated rewards. For example, if an inference result learns that certain variables have a significant impact on the outcomes, this insight can be used to improve the design of future experiments \citep{shi2021statistical, zhang2021statistical, zhang2022statistical, shi2022off}. Different from in-experiment learning focusing on maximizing reward within the trial, statistical inference can lead to more strategic and informed decision-making over time \citep{simchi2023multi}. Therefore, our work aims to provide a comprehensive online inferential framework applicable throughout a wide range of sequential decision-making algorithms.

Motivated by the importance of statistical inference, we first provide a procedure to conduct entry-wise inference on the true matrix parameter $M_i$ under the sequential decision-making framework. We introduce a matrix $T \in \mathbb{R}^{d_1 \times d_2}$ such that $\langle M_i, T\rangle$ characterizes the entries of interest for hypothesis testing. For example, setting $T = e_{j_1}e_{j_2}^\top$, where $\{e_{j_1}\}_{j_1 \in [d_1]}$ and $\{e_{j_2}\}_{j_2 \in [d_2]}$ denote the canonical basis vector in $\mathbb{R}^{d_1}$ and $\mathbb{R}^{d_2}$, respectively, our work allows a valid confidence interval of $\langle M_i, T\rangle = M_i(j_1,j_2)$ for hypothesis testing on whether the $(j_1,j_2)$-th entry of the matrix $M_i$ is zero, i.e., 
\begin{equation}
\label{eq: hype test 1}
        H_0 : M_i(j_1,j_2) = 0 \hspace{3mm}  \text{v.s.} \hspace{3mm} H_1: M_i(j_1,j_2) \ne 0,
\end{equation}
where $M_i(j_1,j_2)$ denotes the $(j_1,j_2)$ entry of $M_i$. In this case, we can test the effectiveness of a certain entry in the matrix context for describing the reward. It is worth pointing out that the form of $T$ is flexible. For example, setting $T = e_{j_1}e_{j_2}^\top - e_{j_3}e_{j_4}^\top$ can test whether $M_i(j_1,j_2)$ and $M_i(j_3,j_4)$ are significantly different. Moreover, our work also enables us to check whether different actions result in different effectiveness of a certain context entry by testing
\begin{equation}
\label{eq: hype test 2}
        H_0 : M_1(j_1,j_2) - M_0(j_1,j_2) = 0 \hspace{3mm}  \text{v.s.} \hspace{3mm} H_1: M_1(j_1,j_2) - M_0(j_1,j_2) \ne 0.
\end{equation}
As \cite{poldrack2011handbook} introduced in their neuroimaging book, statistical inference on the pixel level is able to test whether an individual pixel in an image has a significant effect on measuring the outcome. In our motivational example in Figure \ref{fig:chicken pox}, hypothesis test \eqref{eq: hype test 1} provides the answer of whether a certain pixel is significant in determining the reward, while hypothesis test \eqref{eq: hype test 2} helps us understand if the intervention causes a significant difference in the patient's health outcome.

In addition to the parameter inference, we further extend our online inference framework to the optimal policy value. This value represents the best-expected reward a decision-maker can achieve given complete knowledge of the environment. The need to infer this optimal value becomes crucial in real-world applications whenever the experimenters need to assess the best possible reward they can achieve given the currently available interventions. Such assessment determines the adequacy of current actions in achieving desirable outcomes or necessitates refinement of the action set. In particular, the optimal policy value attainable under the current environment is defined as 
\begin{equation}
    \label{eq: optimal value}
    V^* = \bE\left[\left \langle M_{a^*(X)},X\right \rangle \right], ~~ \text{with}~~ a^*(X) = I\{\langle M_1 - M_0, X \rangle > 0 \},
\end{equation}
where $a^*(X)$ indicates the optimal policy for a given context $X$ under our reward function described in \eqref{eq: model}. To provide additional clarification, experimenters can assess whether the current best treatment outcome surpasses a certain threshold $(V_0)$ by conducting the following one-sided statistical test:
\begin{equation}
    \label{eq: hype test 3}
    H_0 : V^* \le V_0  \hspace{3mm} \text{v.s.} \hspace{3mm} H_1 : V^* > V_0.
\end{equation}

\begin{figure}
        \centering
        \includegraphics[width = 12cm, height = 3.5cm]{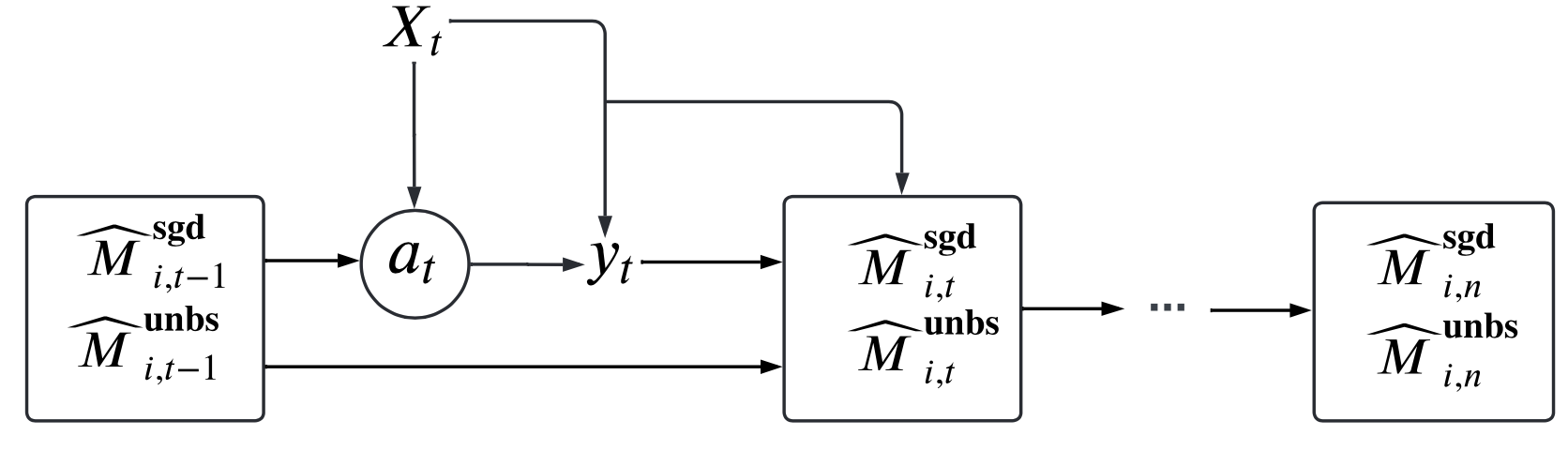}
        \caption{The flow chart of the proposed sequential procedure for a total of $n$ iterations.}
        \label{fig:algo flow}
\end{figure}

After exploring the essential aspects of both parameter inference and optimal policy value inference, we now present our proposed methodology, a procedural framework specifically designed to address these key areas of statistical estimation and inference in online decision-making. In particular, we iteratively update a low-rank estimation of $M_i$ under a sequential decision-making framework with low computational cost. Meanwhile, we simultaneously maintain an unbiased estimator in an online fashion for inference purposes. We briefly illustrate this online procedure in Figure \ref{fig:algo flow} where the low-rank estimation of $M_i$ is denoted as $\Msgd_{i,t}$, and the unbiased estimator for the inference purpose is denoted as $\Munbs_{i,t}$. We summarize the role and properties of both estimators below.
\begin{itemize}
    \item $\Msgd_{i,t}$: Low-rank but biased, sequentially updated low-rank estimation for $M_i$. \vspace{2mm}
    
    \item $\Munbs_{i,t}$: Unbiased but not low-rank, designed for conducting inference of $M_i$.
\end{itemize}

In our problem, it is important to maintain both estimators to handle the two tasks of sequential decision-making and online inference. The methodological contributions of our proposed procedure can be viewed from three aspects. First, in existing low-rank literature, a low-rank estimator is typically obtained by solving nuclear-norm penalized optimization using offline samples \citep{candes2011tight, koltchinskii2015optimal, chen2019inference,xia2019confidence}. However, the offline methods become impractical when handling large-scale matrices due to the substantial storage costs. For instance, storing a single $500 \times 500$ single-precision matrix requires about one megabyte, underscoring the significant storage demands in an offline setting where thousands of such matrices are necessary. In contrast, our proposed online estimation method exhibits distinct advantages in terms of data storage efficiency by eliminating the need for local storage of the complete dataset. Our online estimation procedure uses a single observation at a time and then discards it, which makes this technique particularly well-suited for high-dimensional datasets. In our method, we sequentially update the low-rank factorization of $\Msgd_{i,t}$ via stochastic gradient descent (SGD) to preserve its low-rankness. While it is suitable for sequential decision-making, $\Msgd_{i,t}$ is not directly applicable for statistical inference due to its bias. This motivates our new design of an unbiased estimator $\Munbs_{i,t}$ by sequentially debiasing $\Msgd_{i,t}$ for online inference.

\begin{figure}
    \centering
     \begin{subfigure}{0.31\linewidth}
         \includegraphics[width=\linewidth]{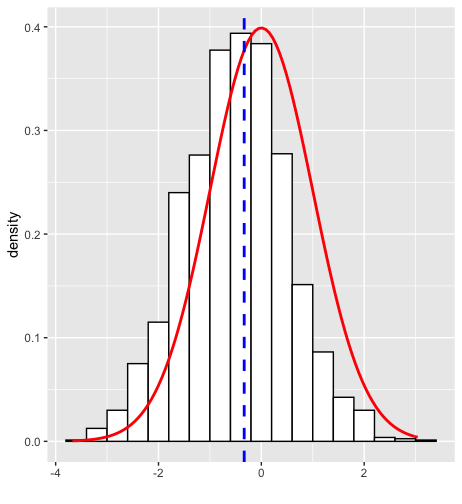}
         \caption{Bias of the estimator caused by adaptively collected data.}
         \label{fig: a}
     \end{subfigure}
     \hfill
     \begin{subfigure}{0.31\linewidth}
         \includegraphics[width=\linewidth]{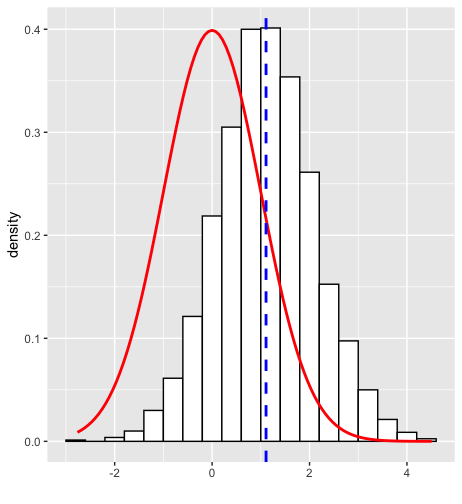}
         \caption{Bias of the estimator caused by the low-rankness.}
         \label{fig: b}
     \end{subfigure}
     \hfill
     \begin{subfigure}{0.31\linewidth}
         \includegraphics[width=\linewidth]{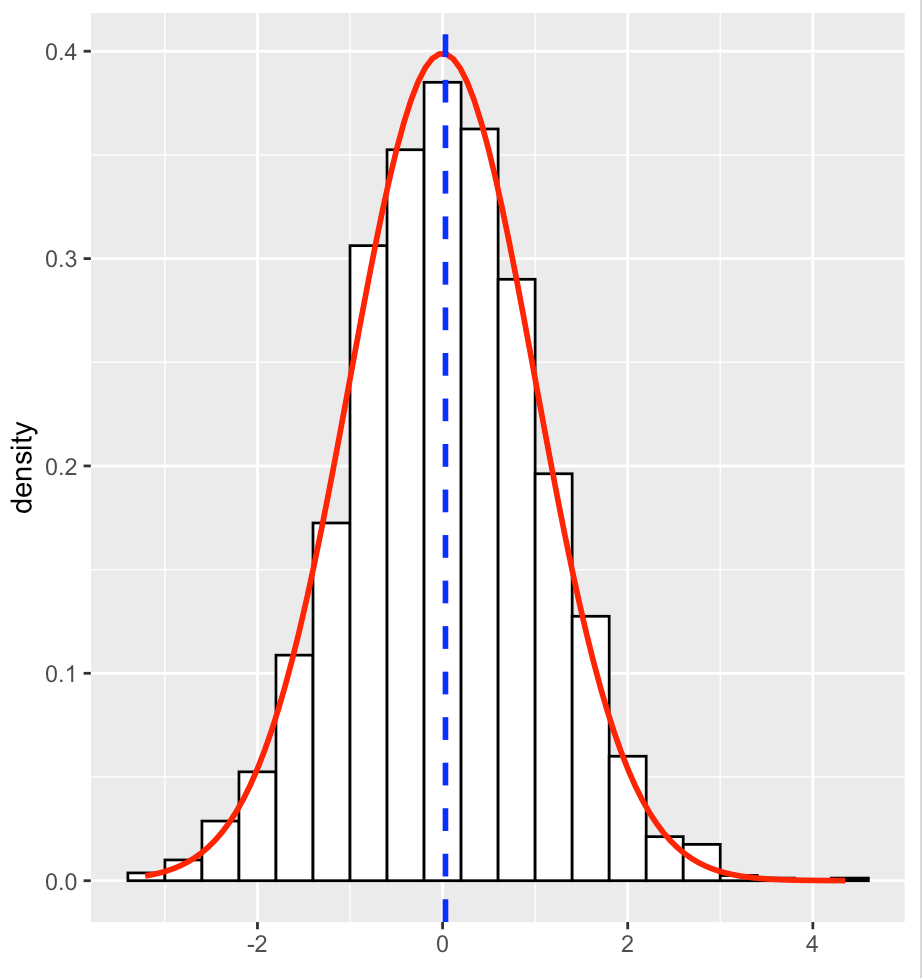}
         \caption{Our proposed debiased estimator}
         \label{fig: c}
     \end{subfigure}
    \caption{The empirical distributions of two biased estimators and our debiased method. The center of each empirical distribution is shown in the blue dashed line, and the standard normal curve is shown in red.}
    \label{fig: bias comparison}
\end{figure}

Second, the debiasing procedure to obtain $\Munbs_{i,t}$ also requires delicate design since it needs to compensate for two sources of bias: (1) the bias in $\Msgd_{i,t}$ caused by preserving the low-rankness, and (2) the bias in adaptive sample collection due to the fact that the samples are not collected randomly, but rather through the distribution of $a_t$ which is determined by the historical information. To illustrate these two types of bias, Figure \ref{fig: a} demonstrates the bias of the estimator caused by adaptive sample collection, and Figure \ref{fig: b} demonstrates the bias of the estimator caused by the low-rankness. To fill in the gap, we introduce a new debiasing approach to handle both sources of bias simultaneously in a sequential manner. Figure \ref{fig: c} shows that our proposed estimator is unbiased and enables a valid statistical inference.

Third, we further introduce an online estimator tailored for optimal policy value inference. While most of the existing literature focuses on offline value inference, our proposed estimator for the optimal policy value equips the experimenters with the ability to monitor the confidence interval of the optimal policy value in a timely manner. Unlike the approach for parameter inference, which requires a sufficient sample size for both action $1$ and action $0$ to ensure adequate information is collected for $M_1$ and $M_0$, the optimal policy value estimator only leverages samples obtained through the estimated optimal action at each time. As a result, our approach to inferring the optimal policy value enables the exploration probability to gradually decrease over time. Additionally, our framework is adaptable to handle scenarios in which the probabilities of selecting each action, as determined by the decision-making policy, are unknown and estimated empirically. 

In addition to the aforementioned methodological contributions, we further summarize our theoretical contributions and discuss the technical challenges in our analysis. 

\begin{itemize}
    \item We provide a non-asymptotic convergence result for the sequentially updated low-rank estimator $\Msgd_{i,t}$ in Theorem \ref{thm: sgd consistent}. That is, with high probability,
    \begin{equation*}
        \|\Msgd_{i,t} - M_i \|_{\mathrm{F}} \le C \sigma_i\sqrt{\frac{dr\log^2d}{t^\varsigma}},
    \end{equation*}
   for some positive constant $C$, where $d = \max\{d_1,d_2\}$, and $\varsigma \in (0.5,1)$. The existing SGD literature for the low-rank estimation is limited except  \cite{jin2016provable} considers a noiseless matrix completion problem with i.i.d. samples. Our work, on the other hand, deals with noisy reward and the adaptive sampling in the sequential decision-making setting. In the noiseless scenario, stochastic objective functions share the same minimizer, with each gradient descent iteration steadily progressing toward this common minimizer. However, the introduction of noise leads to the steps of SGD targeting varying minimizers, causing the SGD updates to oscillate or move away from the optimal solution's local region. To prevent this from happening, it is crucial to add stabilization measures to ensure the optimization trajectory consistently advances toward the right direction. 
    \item We establish the asymptotic normality of $\widehat{m}_T^{(i)}$ for estimating $m_T^{(i)} = \langle M_i,T\rangle$ in Theorem \ref{thm1}. Due to the fact that our data are collected adaptively and sequentially, the analysis based on offline \emph{i.i.d.} samples is no longer applicable in our case. Traditional debiasing approach in the offline low-rank literature \citep{xia2021statistical} involves splitting the dataset into two independent sets, using one to correct biases in the low-rank estimator derived from the other one. However, in online decision-making, where data is passed only once, a sequential debiasing method is necessary. Gathering all data for debiasing at the end is computationally infeasible and renders existing methods ineffective. Our sequential method eliminates the need to store historical data, allowing efficient debiasing at each step in the online decision-making process. Due to these significant differences, new proof techniques are necessary to address the dependency on data. In addition, due to both low-rankness and data adaptivity, our proof involves controlling the additional variance introduced by our debiasing procedure. As an important step, the convergence result of $\Msgd_{i,t}$ shown in Theorem \ref{thm: sgd consistent} ensures this additional variance is well controlled. 
    \item For the purpose of statistical inference of the parameter, we propose a fully online estimator for the variance of $\widehat{m}_T^{(i)}$ without storing historical data. We prove the consistency of this estimator, which provides the guarantees that the asymptotic normality in Theorem \ref{thm2} holds with the estimated standard deviation. This ensures the validity of our constructed confidence interval for the true matrix parameter. 
    \item Finally, we establish the asymptotic normality of our optimal policy value estimator in Theorem \ref{thm:value inference}, showing that the asymptotic bias of the estimator approaches zero with data accumulation. We additionally propose a variance estimator for constructing confidence intervals, and Theorem \ref{thm:consistent} demonstrates the reliability of this estimator, affirming the empirical validity of the generated confidence intervals. Besides addressing the theoretical challenges posed by non-\emph{i.i.d.} data collection and the low-rank structure, establishing the asymptotic normality of the optimal policy value estimator also involves ensuring convergence of the estimated optimal action towards the true optimal action. This is crucial for controlling the bias resulting from the accumulation of differences between the estimated and true optimal actions, which is shown to be sufficiently small compared to the variance of the optimal policy value estimator.  
\end{itemize}

\subsection{Related Literature}
This section discusses three lines of related work, including online inference based on SGD, statistical inference in bandit and Reinforcement Learning (RL) settings, and statistical inference for low-rank matrices. The literature review presents the fundamental differences compared to our work in terms of motivation and problem settings, which end up with different algorithms and technical tools for theoretical analysis. 

\textbf{Online Inference Based on SGD.} Our work is related to a recent growing literature on statistical inference based on SGD. \cite{fang2018online} proposed an online bootstrap procedure for the estimation of confidence intervals of the SGD estimator. \cite{chen2020statistical} studied the statistical inference of the true model parameters by proposing two consistent estimators of the asymptotic covariance of the averaged SGD estimator, extended by \cite{zhu2021online} to a fully online scenario. \cite{shi2021statisticalHigh} developed an online estimation procedure for high-dimensional statistical inference. \cite{chen2021online} studied the online inference when the gradient information is unavailable and \cite{tang2023acceleration} extends the analysis to SGD with momentum. All of these works consider i.i.d. samples and are not applicable to adaptively collected data. Recently, \cite{chen2021sgd,chen2022online} conducted the statistical inference of the model parameters via SGD under online decision-making settings. \cite{ramprasad2022online, liu2023online} studied the online inference in linear stochastic optimization with Markov noise. However, none of these works handles the low-rankness in a matrix estimation.

\textbf{Statistical Inference in Bandit and RL Settings.} \cite{chen2021statistical} studied the asymptotic behavior of the parameters under the traditional linear contextual bandit framework. \cite{bibaut2021post} studied the asymptotic behavior of the treatment effect with contextual adaptive data collection. \cite{zhan2021off} and \cite{hadad2021confidence} developed adaptive weighting methods to construct estimators that are suitable for policy value inference with adaptive collected data. \cite{deshpande2021online} and \cite{khamaru2021near} considered the adaptive linear regression. \cite{zhang2021statistical, zhang2022statistical} provided statistical inference for the M-estimators in the contextual bandit and non-Markovian environment. \cite{shen2021doubly} employed a doubly robust estimator for the optimal policy value inference within an online decision-making framework. In addition to these references, there are also related inference works in RL. For example, \cite{shi2021statistical} constructed the confidence interval for the policy value in the Markov decision process, and \cite{shi2022off,bian2024off} further extended the statistical inference to the confounded Markov decision processes and doubly inhomogeneous environments, respectively. The above works are tailored for vector contexts and not for matrix contexts.

\textbf{Statistical Inference for Low-Rank Matrix.} With the sample splitting procedure for obtaining an unbiased estimator, \cite{carpentier2019uncertainty} constructed confidence sets for the matrix of interest with regard to its Frobenius norm. \cite{xia2019confidence} conducted the inference on the matrix's singular subspace, reflecting the information about matrix geometry. To conduct inference on matrix entries, \cite{carpentier2018iterative} proposed a new estimator that was established using the iterative thresholding method. \cite{chen2019inference} proposed a debiased estimator for a matrix completion problem. \cite{xia2021statistical} studied the inference of a matrix linear form, which established the entry-level confidence intervals. However, none of the above works is applicable when the data are adaptively collected. As shown in Figure \ref{fig: bias comparison}, we need to handle two sources of bias in our setting, which demands a new debiasing procedure.

\subsection{Notations and Organization}
\label{sec: notations}
For a matrix $M \in \mathbb{R}^{d_1 \times d_2}$, we use $\| M \|_{\mathrm{F}}$ to denote its Frobenius norm, $\| M\|$ to denote its matrix operator norm, and $\| M\|_{\ell_1}$ to denote its vectorized $\ell_1$ norm. We use $M(i,j)$ to denote the entry of $M$ at row $i$ and column $j$. Assume a matrix has rank $r$, then we denote the $\lambda_1, \lambda_r$ as its largest and smallest singular values, respectively, and we denote $\kappa(M) = \lambda_1/\lambda_r$ as the condition number of $M$. Given a matrix $A \in \mathbb{R}^{d_1 \times d_2}$, we denote $\langle M, A \rangle$ as the matrix inner product, i.e., $\langle M, A \rangle = \text{tr}(M^\top A)$. For a matrix $U \in \mathbb{R}^{d \times r}$, then we denote its orthogonal complement as $U_{\bot} \in \mathbb{R}^{d \times (d-r)}$. We use the notation $C_1,C_2, \ldots$ to represent the absolute constants, and we use $ a\lesssim b$ to represent $a \le Cb$ for some absolute constant $C$. We denote $\xrightarrow{p}$ and $\xrightarrow{d}$ as convergence in probability and in distribution, respectively. Finally, we use $I\{\cdot\}$ to denote the indicator function.

The rest of the paper is organized as follows. In Section \ref{sec: decision making}, we introduce our problem setting and decision-making procedure under the online decision-making framework. In Section \ref{sec: Inference}, we propose the online debiasing procedure to construct an unbiased estimator for inference purposes. We also present the asymptotic normality of the proposed estimator and prove the validity of the proposed statistical inference procedure. In Section \ref{sec:optimal policy value}, we outline a procedure for inferring the value of the optimal policy. In Section \ref{sec: experiment}, we present numerical experiments to demonstrate the merit of our proposed method. Finally, the supplementary material includes additional numerical studies, further discussions on assumptions, and comprehensive proofs of main theorems and technical lemmas.

\section{Online Decision Making and Low-Rank Estimation}
\label{sec: decision making}
In this section, we first present the online decision-making procedure designed to address the exploration-exploitation dilemma. Subsequently, we propose a sequential low-rank estimation for $M_i$, denoted as $\Msgd_{i,t}$ for $i=0,1$ and $t=1, 2, \dots$. The convergence properties of the proposed SGD estimator are discussed in the later part of this section.

    \subsection{Sequential Decision Making}
    \label{sec: sequential decision making}
    In sequential decision-making, the objective is to select a series of actions over time aiming to maximize the cumulative reward. As described by our reward model, denoted by \eqref{eq: model}, the reward, represented by $y_t$ at time $t$, is observed after the execution of an action $a_t$. Let $\mathcal{F}_{t}$ denote the filtration generated by all the historical randomness up to time $t$, i.e.,  $\mathcal{F}_t = \sigma ( X_1, a_1, y_1,...,X_t,a_t, y_t )$. Then the policy function, denoted as $\pi_t$, can be formally expressed as
    \begin{align*}
        \bP(a_t = 1 | \mathcal{F}_{t-1}, X_t) = \pi_t(X_t, \Msgd_{1,t-1}, \Msgd_{0,t-1}),
    \end{align*}
     and correspondingly, $\bP(a_t = 0 | \mathcal{F}_{t-1}, X_t) = 1- \pi_t(X_t, \Msgd_{1,t-1}, \Msgd_{0,t-1})$. Here, the domain and range of policy function can be specified as $\pi_t: \mathbb{R}^{d_1 \times d_2} \times  \mathbb{R}^{d_1 \times d_2} \times  \mathbb{R}^{d_1 \times d_2} \rightarrow [0,1]$. To streamline notation, we employ $\pi_t$ to represent the probability of selecting action $a_t=1$ at time $t$, while $1-\pi_t$ denotes the probability associated with selecting $a_t = 0$ accordingly. 
     
     The estimation and inference procedure introduced in this work is applicable to a wide range of randomized bandit policies, 
     and here we list three examples.
    \begin{itemize}
        \item \textbf{$\varepsilon$-Greedy.}
        One widely used policy demonstrating the exploration-exploitation tradeoff is the $\varepsilon$-greedy approach \citep{lattimore2020bandit} which allocates $\varepsilon_t/2$ as the exploration probability while $1-\varepsilon_t/2$ for exploitation at each iteration. With any pre-specified $\varepsilon_t \in (0,1)$, $\pi_t$ can be explicitly expressed using $\varepsilon_t$. Specifically, probability of taking action $a_t = 1$ at time $t$ is described as
        \begin{equation*}
            \bP(a_t = 1|\mathcal{F}_{t-1}, X_t) = (1-\varepsilon_t)I\left\{\langle \Msgd_{1,t-1} - \Msgd_{0,t-1}, X_t\rangle>0\right\} + \frac{\varepsilon_t}{2}.
        \end{equation*}
         
        \item \textbf{Softmax Policy.} Our proposed method can also be employed effectively with softmax policies that utilize exponential weighting schemes to balance exploration and exploitation. Consider the following probability model for choosing action $a_t = 1$,
        \begin{equation*}
            \bP(a_t = 1|\mathcal{F}_{t-1}, X_t) = \frac{\exp(\langle \Msgd_{1,t-1}, X_t\rangle )}{ \exp(\langle \Msgd_{0,t-1}, X_t\rangle)+ \exp(\langle \Msgd_{1,t-1}, X_t\rangle)}.
        \end{equation*}
        The action with a higher estimated reward is assigned with a higher probability through a softmax transformation. Popular applications include EXP3, EXP4 \citep{auer2002nonstochastic}, and softmax policy gradient \citep{mei2020global, boutilier2020differentiable, agarwal2021theory}.\vspace{2mm}

        \item \textbf{Thompson Sampling.} Thompson Sampling \citep{lattimore2020bandit} balances the exploration-exploitation trade-off by sampling from the posterior distribution over the expected reward for each action. At time $t$, the algorithm samples the matrix parameter $\bar{M}_{i,t}$ from the posterior distribution $\mathcal{P}^{(i)}(\cdot | \mathcal{F}_{t-1})$, and chooses the action to be the one that gives the maximum reward, i.e.,  $a_t = \argmax_i ~ \langle \bar{M}_{i,t}, X_t \rangle$. As the posterior distribution may not have an explicit form, approximate sampling could be employed and we discuss an adapted approach in the supplementary material.  
    \end{itemize}

Although our focus in the main paper remains on the aforementioned randomized policies with known action probabilities to enhance clarity, we also detail a methodology and accompanying theoretical analysis for scenarios where action probabilities are unknown. This discussion is provided in the supplementary material. These popular bandit algorithms typically select actions at time $t$ based on current estimations of model parameters. Therefore, an accurate estimation of $M_i$ enables more precise reward predictions, thereby enhancing the decision-making performance. In the following section, we introduce the methodology for deriving a sequential and sample-efficient estimator for $M_i$. 
    
    \subsection{Online Low-Rank Estimation via SGD}
    \label{sec: low rank estimation}
    In this section, we introduce the procedure to obtain the online low-rank estimator $\Msgd_{i,t}$. The estimation method needs to meet two requirements: (1) the estimator should be updated sequentially under the online decision-making framework, and (2) the estimator should leverage the inherent low-rank structure to ensure sample efficiency. To accomplish these tasks, we apply SGD to iteratively update the estimation of the low-rank factorization of $M_i$. Specifically, for $i = 0,1$, we solve the following stochastic optimization problem via SGD,
    \begin{equation}
    \label{eq: pop loss}
        \min_{\Usgd_i\in \mathbb{R}^{d_1 \times r},\Vsgd_i\in \mathbb{R}^{d_2 \times r}}F\left(\Usgd_i,\Vsgd_i\right) ~=~  \mathbb{E}\Big[ f\left(\Usgd_i, \Vsgd_i; \{X,y\}\right)\Big], 
    \end{equation}
    where the expectation is taken with respect to the randomness of $\{X,y\}$, and the individual loss function is defined as
    \begin{equation}
    \label{eq: indiv loss}
        f\left(\Usgd_i, \Vsgd_i; \{X,y\}\right) = \frac{1}{2} \left(y - \left \langle \Usgd_i\Vsgd^\top_i, X\right\rangle \right)^2.
    \end{equation}
    If we denote $\Usgd_{i,t}$ and $\Vsgd_{i,t}$ as the estimated $\Usgd_i$ and $\Vsgd_i$ at time $t$, respectively, a naive SGD approach for implementing the update at time $t$ with learning rate $\eta_t$ is given by
    \begin{equation}
    \label{eq: naive update}
        \left(\begin{array}{l}
        \Usgd_{i,t}\\
        \Vsgd_{i,t}
        \end{array}\right) = \left(\begin{array}{l}
        \Usgd_{i,t-1}\\
        \Vsgd_{i,t-1}
        \end{array}\right) - \eta_t I\{a_t =i \}\nabla f(\Usgd_{i,t-1}, \Vsgd_{i,t-1};\{X_t, y_t\}),
    \end{equation}
    where $\nabla f$ is the gradient of the individual loss function in \eqref{eq: indiv loss}, i.e.,
    \begin{equation*}
        \nabla f(\Usgd_{i,t-1}, \Vsgd_{i,t-1};\{X_t, y_t\}) = \left(\begin{array}{l}
         (\langle\Usgd_{i,t-1}\Vsgd_{i,t-1}^\top , X_t\rangle -y_t )X_t\Vsgd_{i,t-1} \\
        (\langle \Usgd_{i,t-1}\Vsgd_{i,t-1}^\top , X_t\rangle  - y_t )X^\top_t\Usgd_{i,t-1}
        \end{array}\right).
    \end{equation*}
    
    However, this naive implementation is not applicable to our analysis for two reasons. First, the stochastic gradient given in the above form is no longer an unbiased estimator of the population gradient $\nabla F(\Usgd_{i,t-1},\Vsgd_{i,t-1})$ because this stochastic gradient depends on the adaptive distribution of $a_t$ while the population gradient does not. Second, our analysis requires that $\Usgd_{i,t}$ and $\Vsgd_{i,t}$ stay in a neighborhood such that $F(\Usgd_{i,t}, \Vsgd_{i,t})$ enjoys the smoothness and strong convexity, but this naive approach may destroy this geometric property of $F$ as discussed later in Section \ref{sec: explain stochastic gradient}. To address the aforementioned two concerns, we propose our stochastic gradient as
    \begin{align*}
    \label{eq: practice gradient}
         & g(\Usgd_{i,t-1}, \Vsgd_{i,t-1};\{X_t, y_t,a_t, \pi_t\}) \numberthis \\
         =&   \frac{I\{ a_t = i\}}{i\pi_t + (1-i)(1-\pi_t)}\left(\begin{array}{l}
       (\langle \Usgd_{i,t-1}\Vsgd_{i,t-1}^\top , X_{t}\rangle - y_t)X_{t}\Vsgd_{i,t-1}R_{\Vsgd}D_{\Vsgd}^{-\frac{1}{2}}Q_{\Vsgd}Q_{\Usgd}^\top D_{\Usgd}^{\frac{1}{2}}R_{\Usgd}^\top \\
        (\langle \Usgd_{i,t-1}\Vsgd_{i,t-1}^\top , X_{t}\rangle-y_t)X_{t}\Usgd_{i,t-1}R_{\Usgd}D_{\Usgd}^{-\frac{1}{2}}Q_{\Usgd}Q_{\Vsgd}^\top D_{\Vsgd}^{\frac{1}{2}}R_{\Vsgd}^\top
        \end{array}\right).
     \end{align*}
     We describe the procedure of obtaining the above auxiliary matrices at each iteration in Algorithm \ref{alg:sgd update practice}. The inverse weight $1/[i\pi_t + (1-i)(1-\pi_t)]$ is applied to compensate for the bias in the naive stochastic gradient in \eqref{eq: naive update} caused by the adaptive distribution of $a_t$, where we recall that $\pi_t$ is the shorthand notation for $\bP(a_t=1|\mathcal{F}_{t-1},X_t)$. Besides the inverse weighting, our form of $g$ also serves as a computationally efficient method for re-normalizing $\Usgd_{i,t-1}$ and $\Vsgd_{i,t-1}$ to ensure that each iterate stays in a neighborhood. We provide more explanations and benefits of choosing $g$ as our stochastic gradient in Section \ref{sec: explain stochastic gradient}. Given the designed stochastic gradient $g$, our updating rule is 
     \begin{equation}
     \label{eq: practice update}
        \left(\begin{array}{l}
        \Usgd_{i,t}\\
        \Vsgd_{i,t}
        \end{array}\right) = \left(\begin{array}{l}
        \Usgd_{i,t-1}\\
        \Vsgd_{i,t-1}
        \end{array}\right) - \eta_t g(\Usgd_{i,t-1}, \Vsgd_{i,t-1};\{X_t, y_t,a_t, \pi_t\}),
    \end{equation}
    where we require the learning rate $\eta_t$ to decay as $t$ grows to diminish the effect of the noise in the convergence analysis. We defer the discussion of the learning rate to Section \ref{sec: convergence analysis sgd}. To further clarify this updating rule, we take $a_t = 1$ at time $t$ for example, then $g(\Usgd_{0,t-1}, \Vsgd_{0,t-1};\{X_t, y_t,a_t, \pi_t\}) =(0,0)^\top$, which implies $\Usgd_{0,t}$, $\Vsgd_{0,t}$ (for the action $a_t=0$) are not updated. Meanwhile, the singular value decomposition (SVD) is applied to $\Usgd_{1,t-1}^\top\Usgd_{1,t-1}$ and $\Vsgd_{1,t-1}^\top\Vsgd_{1,t-1}$ after $\Usgd_{1,t-1}$ and $\Vsgd_{1,t-1}$ are updated according to \eqref{eq: practice update}. The one-step update at time $t$ is summarized in Algorithm \ref{alg:sgd update practice}. Finally, we set $\Msgd_{i,t} = \Usgd_{i,t}\Vsgd_{i,t}^\top$, which will be used for the decision policy in the next iteration.
    \begin{algorithm}[t]
     \caption{One-Step SGD Update at time $t$}
        \label{alg:sgd update practice}
        \begin{algorithmic}[1]
        \State \textbf{Input}: $\Usgd_{i,t-1}$, $\Vsgd_{i,t-1}$ for $i = 0,1$, $X_t$, $y_t$, $a_t$, $\pi_t$, $\eta_t$ \vspace{2mm}
        
         \State \hspace{4mm} $R_{\Usgd}D_{\Usgd}R^\top_{\Usgd} \leftarrow$ SVD $\left (\Usgd_{a_t,t-1}^\top \Usgd_{a_t,t-1}\right)$ , $R_{\Vsgd}D_{\Vsgd}R^\top_{\Vsgd} \leftarrow$ SVD $\left (\Vsgd_{a_t,t-1}^\top \Vsgd_{a_t,t-1}\right)$. \vspace{2mm}

         \State \hspace{4mm} $Q_{\Usgd}DQ_{\Vsgd} \leftarrow$ SVD$\left (D_{\Usgd}^{\frac{1}{2}}R^\top_{\Usgd}R_{\Vsgd}D_{\Vsgd}^{\frac{1}{2}}\right)$. \vspace{2mm}

        \State \hspace{4mm} For $i=0,1$, update $\Usgd_{i,t}$, $\Vsgd_{i,t}$ using \eqref{eq: practice update}.\vspace{2mm} 
        
        \State \textbf{Output}: $\Usgd_{i,t}$, $\Vsgd_{i,t}$, $R_{\Usgd}$, $D_{\Usgd}$, $R_{\Vsgd}$, $D_{\Vsgd}$
        \end{algorithmic}
    \end{algorithm}

\subsection{Explanation of the Form of Stochastic Gradient}
    \label{sec: explain stochastic gradient}
    We first discuss the necessity of applying the inverse weighting to compensate for the bias caused by the adaptive distribution of $a_t$. Then we discuss the necessity of renormalizing $\Usgd_{i,t-1}$ and $\Vsgd_{i,t-1}$ at each time $t$. Finally, we demonstrate that Algorithm \ref{alg:sgd update practice} only requires computing the SVD for an $r \times r$ matrix instead of a $d_1 \times d_2$ matrix at each iteration for re-normalization, which makes our algorithm computationally efficient. 
     
     As the SGD update is implemented under the online decision-making setting, the samples are collected through the action $a_t$ according to our decision-making policy at each time. This implies that the sample used for each update is not collected randomly but based on the “past experience'' inherited in the distribution of $a_t$. Since the action $a_t$ determines either $(\Usgd_{1,t},\Vsgd_{1,t})$, or $(\Usgd_{0,t},\Vsgd_{0,t})$ to be updated at time $t$, we need to eliminate this bias so that the estimation for both $i=0$ and $1$ can be treated equally. Inspired by \cite{chen2021sgd}, we apply the inverse weight that serves as a distribution correction that compensates for the aforementioned bias using the fact $\mathbb{E}\left [I\{a_t = i\} | X_t, \mathcal{F}_{t-1} \right] = i\pi_t + (1-i)(1 - \pi_t)$.
     
    To ensure the convergence of our algorithm, we need $\Usgd_{i,t}$ and $\Vsgd_{i,t}$ to stay in a local region. The naive implementation of SGD such as \eqref{eq: naive update} might end up with an estimator $\Usgd_{i,t}$ very large and $\Vsgd_{i,t}$ very small or vice versa even though $\Usgd_{i,t}\Vsgd_{i,t}^\top$ is a reasonable estimate of $M_i$ \citep{jin2016provable}. To see it, assuming we have matrices $A \in \mathbb{R}^{d_1 \times r}$ and $B \in \mathbb{R}^{d_2 \times r}$, then $AB^\top = \tilde{A}\tilde{B}^\top$ even if $\tilde{A}$ is very small while $\tilde{B}$ very large, e.g. $\tilde{A} = \delta A$ and $\tilde{B} = \delta^{-1}B$ for some very small scalar $\delta$. To avoid this situation, we can apply re-normalization at the beginning of each iteration by setting $\tilde{\Usgd}_{a_t,t-1} = W_{\Usgd}D^{\frac{1}{2}}$ and $\tilde{\Vsgd}_{a_t,t-1} = W_{\Vsgd}D^{\frac{1}{2}}$, where $W_{\Usgd}D W_{\Vsgd}^\top$ is the top-$r$ SVD of $\Usgd_{a_t,t-1}\Vsgd_{a_t,t-1}^\top$, meaning that $W_\Usgd$ and $W_{\Vsgd}$ are the top-$r$ singular vectors. On the other hand, we leave $(\tilde{\Usgd}_{1-a_t,t-1}, \tilde{\Vsgd}_{1-a_t,t-1})$ unchanged from the last iteration, i.e.,  $(\tilde{\Usgd}_{1-a_t,t-1}, \tilde{\Vsgd}_{1-a_t,t-1}) = (\Usgd_{1-a_t,t-1}, \Vsgd_{1-a_t,t-1})$. Then a straightforward way to deal with this concern is to plug the renormalized version $\tilde{\Usgd}_{a_t,t-1}$ and $\tilde{\Vsgd}_{a_t,t-1}$ into \eqref{eq: naive update} with the inverse weighting 
     \begin{equation}
     \label{eq: normalize update}
        \left(\begin{array}{l}
        \Usgd_{i,t}\\
        \Vsgd_{i,t}
        \end{array}\right) = \left(\begin{array}{l}
        \tilde{\Usgd}_{i,t-1}\\
        \tilde{\Vsgd}_{i,t-1}
        \end{array}\right) - \eta_t \frac{I\{a_t = i\}}{i\pi_t + (1-i)(1 - \pi_t)}\nabla f(\tilde{\Usgd}_{i,t-1}, \tilde{\Vsgd}_{i,t-1};\{X_t, y_t\}).
    \end{equation}
    In this case, the strong convexity and smoothness of $F$ can be guaranteed within the neighborhood of $(\tilde{\Usgd}_{i,t-1},\tilde{\Vsgd}_{i,t-1})$. Unfortunately, this naive approach requires computing the SVD of a $d_1\times d_2$ matrix at each iteration, which incurs a huge computational cost. Nonetheless, the low-rankness of $\Usgd_{i,t}$ and $\Vsgd_{i,t}$ allows us to compute a cheaper SVD on $r\times r$ matrices $\Usgd^\top_{i,t}\Usgd_{i,t}$ and $\Vsgd^\top_{i,t}\Vsgd_{i,t}$ instead. The resulting alternative approach, described in Algorithm \ref{alg:sgd update practice} using \eqref{eq: practice gradient} as the stochastic gradient, handles the re-normalization issue in a computationally efficient way. It only remains to show the equivalency between \eqref{eq: practice update} and \eqref{eq: normalize update}, which demonstrates that the re-normalization can be done by applying the SVD of $r \times r$ matrices.
    \begin{lemma}[\citealt{jin2016provable}]
    \label{prop: equivalency}
     The updating rules given by \eqref{eq: practice update} and \eqref{eq: normalize update} are equivalent in the sense that, at any time $t$, the updates $\Usgd_{i,t}$, $\Vsgd_{i,t}$  from \eqref{eq: practice update}, and $\Usgd'_{i,t}$ and $\Vsgd'_{i,t}$ from \eqref{eq: normalize update}, satisfy the relation $\Usgd'_{i,t}\Vsgd'^\top_{i,t} = \Usgd_{i,t}\Vsgd^\top_{i,t}$.
    \end{lemma}
    Lemma \ref{prop: equivalency} follows directly from Lemma 3.2 in \cite{jin2016provable}, establishing computational equivalence between two SVD procedures. While the renormalization technique is adapted for computational efficiency, our statistical convergence analysis for stochastic gradient descent differs due to two reasons. Firstly, our framework encompasses noisy observations, where each stochastic gradient descent iteration does not progress toward a common minimizer. Secondly, our approach requires the integration of decision-making policies throughout data collection. These differences call for new tools to analyze the convergence of our low-rank estimation.

    \subsection{Convergence Analysis of Low-Rank Estimation}
    \label{sec: convergence analysis sgd}
     Before presenting the convergence results, we introduce the following assumptions for our true model.
     \begin{assumption}
        \label{assum: noise} 
        We consider the reward model \eqref{eq: model}. For $i \in \{ 0,1\}$,
        \begin{enumerate}[label=(\roman*)]
            \item  The noise $\xi_t$ given $a_t = i$ are \emph{i.i.d.} sub-Gaussian random variables with parameter $\sigma_i$,
            \begin{equation*}
            \mathbb{E}[\xi_t|a_t=i] = 0, \hspace{2mm}\mathbb{E}[\xi_t^2|a_t = i] = \sigma_i^2, \hspace{2mm} \mathbb{E}[e^{s\xi_t}|a_t = i] \le e^{s^2 \sigma_i^2},\quad \forall s \in \mathbb{R}.
        \end{equation*}
        
        \item The context matrix $X_t$ has \emph{i.i.d} standard Gaussian entries, i.e., $X_t(j_1,j_2) \sim \mathcal{N}(0,1)$. Moreover, $X_t$ is independent from $\mathcal{F}_{t-1}$ and $\xi_t$, and $\{X_t\}$ are \emph{i.i.d.} across all $t$.\vspace{2mm}
        
        \item The true matrix parameter  $M_i$ is low-rank with rank $r \ll \min\{ d_1,d_2\}$, and its condition number is $\kappa(M_i) \le \kappa$ for a positive constant $\kappa$. 
        \end{enumerate}
    \end{assumption}
    Assumption \ref{assum: noise} indicates that the observed $y_t$ after taking action is corrupted by a sub-Gaussian noise with parameter $\sigma_i$, which is a common assumption in online decision-making literature \citep{lattimore2020bandit}. Additionally, we assume the context matrix $X_t$ has \emph{i.i.d.} standard Gaussian entries, which is a typical and convenient assumption in the low-rank matrix regression literature \citep{xia2019confidence}, and this contextual information received at each time is \emph{i.i.d.} and independent from the noise. We note that the Gaussian condition is not exclusive and can be extended to include other distributions. For instance, in the supplementary material, we discuss an alternative design of the contextual matrix that can broaden the scope of our inference framework, moving beyond online low-rank regression to include the case of online low-rank matrix completion.
    Finally, we assume that the matrix is well conditioned with a known rank $r$, which is common in existing low-rank literature \citep{ xia2021statistical,zhu2022learning, chen2019inference,chen2020semiparametric}. A theoretical analysis for the case of unknown $r$ remains unclear even in the traditional matrix regression problems and deserves a careful investigation in future works.
     
     We then discuss the initialization of $\Usgd_i$ and $\Vsgd_i$ for $i = 0,1$. Given a low-rank initialization $\Minit_i$ (i.e., $\Msgd_{i,0}$), we can obtain $\Usgd_{i,0}$ and $\Vsgd_{i,0}$ by applying the SVD on $\Minit_i$. We denote $W^{\textbf{init}}_{\Usgd}$ and $W^{\textbf{init}}_{\Vsgd}$ as the top-$r$ left and right singular vectors of $\Minit_i$, along with a diagonal matrix containing top-$r$ singular values denoted as $D^{\textbf{init}}$. Then we set
     \begin{equation}
     \label{eq:UV init}
         \Usgd_{i,0} =  W_{\Usgd}^{\textbf{init}}(D^{\textbf{init}})^{\frac{1}{2}}, \quad \text{and} \quad \Vsgd_{i,0} =  W_{\Vsgd}^{\textbf{init}}(D^{\textbf{init}})^{\frac{1}{2}}.
     \end{equation}
     For theoretical analysis, we require the following assumption on initialization. 
     \begin{assumption}
     \label{assum: init}
    With $\sigma_i$ specified in Assumption \ref{assum: noise}, the initialization $\Minit_i$ satisfies
     $
         \big \|\Minit_i - M_i \big\|_{\mathrm{F}} \le C \sigma_i$ 
     for $i = 0,1$, and some constant $C>0$.
     \end{assumption}
     The procedure of obtaining such initialization can be seen as the random exploration phase in the bandit problem. Since the samples are independent in the random exploration phase, such initialization condition is mild and can be satisfied by existing low-rank estimation literature \citep{xia2019confidence}. 
     \begin{assumption}
         \label{assum: decay rate}
         The probabilities $\pi_t$ and $1-\pi_t$, defined in Section \ref{sec: sequential decision making}, satisfy
         \begin{equation*}
             \min\{\pi_t, 1-\pi_t\} \ge t^{-\beta}\pmin,
         \end{equation*}
         { for some $ 0 \leq \beta < 1$} and $\pmin \in (0,1)$.
     \end{assumption}
     This assumption ensures sufficient exploration by preventing the exploration probability from decaying too rapidly. When $\beta=0$, it requires a constant lower bound $\pmin$ for exploration, which is a common assumption in SGD-based inference \citep{chen2021sgd,chen2022online}. However, for estimation, Assumption \ref{assum: decay rate} provides flexibility by allowing the lower bound of the exploration probability to decay over time for any $\beta > 0$ for the estimation resuls in this section and the policy value inference in Section 4. 
     
    With all these assumptions, we are ready to present the convergence result of our online low-rank estimation obtained through Algorithm \ref{alg:sgd update practice}. Recall that we define $d = \max \{ d_1, d_2\}$ and set $\Msgd_{i,t} = \Usgd_{i,t}\Vsgd_{i,t}^\top$ at each iteration. To simplify the notations, we assume $\|M_0\|=\|M_1\|=1$, and define $\lambda_r = \min\{\lambda_r(M_1), \lambda_r(M_0) \}$ with the condition number $\kappa\leq 1/\lambda_r$. 
    \begin{theorem}
    \label{thm: sgd consistent}
    Define the learning rate $\eta_{t} = c\cdot (\max\{t, t^\star \})^{-\alpha}$, and $t^\star = \left ( \gamma^2dr\log^2d\right )^{\frac{1}{\alpha-\beta}}$ for some constant $c>0$ and {$\alpha\in(\beta,1)$}. Assume the signal-to-noise ratio $\frac{\lambda_r}{\sigma_i} \ge 10C$ for some constant $C>0$ and Assumptions \ref{assum: noise}--\ref{assum: decay rate} hold. For any large enough $\gamma > 0$, with probability at least $1 - \frac{4n}{d^{\gamma}}$, we have for $1\leq t\leq n$,
    \begin{equation*}
        \left\| \Msgd_{i,t} - M_i \right \|_{\mathrm{F}} \le C_1 \gamma \sigma_i \sqrt{\frac{dr\log^2 d}{t^{\alpha-\beta}}},
    \end{equation*}
    for some positive constant $C_1$.
    \end{theorem}

    \begin{remark}{\hspace{-1em} Theorem \ref{thm: sgd consistent} can be generalized to accommodate a relaxed initial condition $ \|\Minit_i - M_i \|_{\mathrm{F}}\leq C\lambda_r$. This generalization is formally stated in Theorem D.1 of the supplementary material. Specifically, if the initialization falls outside original region defined in Assumption \ref{assum: init} but within the relaxed one, a burn-in phase of estimation ensures that the same convergence rate can be achieved for sufficiently large $t$.}\end{remark}

    When $\beta = 0$, the estimation error rate in Theorem \ref{thm: sgd consistent} reduces to $\tilde{O}(\sqrt{dr/t^\alpha})$, ignoring the logarithm factors, which closely aligns with the statistically optimal rate in the offline setting \citep{xia2019confidence} as one specifies $\alpha$ to be close to $1$. For $\beta > 0$, the decision-making policy allows for a decaying exploration probability, 
    {which may increase the estimation error but could benefit the decision-making objectives. Specifically, under an $\varepsilon$-greedy policy with $\varepsilon_t = p_0 t^{-\beta}$, the cumulative regret over a time horizon of $n$ is bounded by $
    \tilde O(n^{1 - \frac{\alpha - \beta}{2}} +n^{1 - \beta})
    $, ignoring logarithmic terms and dimensionality, where the two terms correspond to the regret due to exploitation and exploration, respectively. The parameter $\beta$ represents a tradeoff between online decision-making and the estimation error. Setting $\beta=\frac13\alpha$ with $\alpha$ approaches $1$, the cumulative regret is of the order $n^{2/3}$. A similar tradeoff in online decision making and parameter estimation has also been observed in \cite{simchi2023multi}. }

   Having developed our online estimation method along with its associated error rate, we now proceed to present the framework for statistical inference. Section \ref{sec: Inference} details the methodology and theoretical foundation for parameter inference, while Section \ref{sec:optimal policy value} focuses on inferring the optimal policy value.

    \section{Parameter Inference}
    \label{sec: Inference}
    In this section, we propose an online framework for conducting entry-wise statistical inference on the parameter $M_i$, which leverages the low-rank estimation from the earlier section. 
    Particularly, we propose a sequential debiasing procedure that can obtain an unbiased estimator by removing the two types of bias inherited in $\Msgd_{i,t}$ simultaneously as shown in Figure \ref{fig: bias comparison}. We first introduce our proposed online debiasing procedure. We then present the asymptotic normality of our proposed unbiased estimator, which serves as the theoretical foundation for conducting the inference. Finally, we propose the estimation of the variance of this unbiased estimator and show the consistency of the estimator. It is worth pointing out that our estimation can be obtained in a fully online fashion without storing historical data.

     \subsection{Online Debiasing Procedure}
     \label{sec: online de-bias}
       As discussed in the existing low-rank matrix inference literature \citep{xia2019confidence,chen2019inference, xia2021statistical}, debiasing is a commonly used method that handles the bias caused by preserving the low-rankness. Unlike existing debiasing approaches, our debiasing procedure needs to deal with two sources of bias. First, even though the estimation method via SGD in Section \ref{sec: low rank estimation} ensures that $\Usgd_{i,t}$ and $\Vsgd_{i,t}$ are unbiased estimators for the corresponding low-rank factorization of $M_i$, there is no guarantee that $\Usgd_{i,t}\Vsgd_{i,t}^\top$ is an unbiased estimator for $M_i$. Second, because the data collection is adaptive through the action $a_t$, we also need to handle the bias introduced by the adaptive samples in the bandit setting. To fill in the gap, we introduce a new debiasing procedure to eliminate both types of bias due to low-rankness and data adaptivity. The unbiased estimator obtained from our proposed online debiasing procedure is described as follows: taking $i = 1$ for example, we define
     \begin{equation*}
         \widetilde{M}_{1,t} = \Msgd_{1,t-1} +  \frac{I\{a_t = 1\}}{\pi_t} (y_t - \langle \Msgd_{1,t-1}, X_t \rangle )X_t,
     \end{equation*}
     at time $t$, and then update an online unbiased estimator
     \begin{equation*}
         \Munbs_{1,t} = ( \widetilde{M}_{1,t} + (t-1) \Munbs_{1,t-1} )/t,
     \end{equation*}
      as the running average of $\widetilde{M}_{1,t}$. We apply the inverse weighting in $\widetilde{M}_{1,t}$ to compensate for the bias caused by the adaptive distribution of $a_t$. Additionally, $(y_t - \langle \Msgd_{1,t-1}, X_t \rangle )X_t$ in the second term of $\widetilde{M}_{1,t}$ can be seen as the gradient of $f(M) = \frac{1}{2}(y_t - \langle M, X_t\rangle)^2$ at $\Msgd_{1,t-1}$. This gradient does not impose low-rank constraint and thus pushes $\Msgd_{1,t-1}$ towards the direction of an unbiased estimation of $M_1$.  Moreover, it is important to note that we use $\Msgd_{1,t-1}$ instead of $\Msgd_{1,t}$ to obtain $\widetilde{M}_{1,t}$. Otherwise, $\widetilde{M}_{i,t}$ would no longer be an unbiased estimator of $M_i$ because updating $\Msgd_{1,t}$ uses the observation $X_t$, causing the dependence between $\Msgd_{1,t}$ and $X_t$. Finally, we obtain our unbiased estimator for the inference purpose as
     \begin{equation}
     \label{eq: Munbs 1}
         \Munbs_{1,n} = \frac{1}{n}\sum_{t=1}^n\Msgd_{1,t-1} + \frac{1}{n} \sum_{t=1}^n \frac{I\{a_t = 1\}}{\pi_t} (y_t - \langle \Msgd_{1,t-1}, X_t \rangle )X_t,
     \end{equation}
      which is essentially the average over $\widetilde{M}_{1,t}$. To see the unbiasness of $\Munbs_{1,n}$ more formally, we define $\Delta_{t-1} =M_1 - \Msgd_{1,t-1} $, and rewrite equation \eqref{eq: Munbs 1} by adding and subtracting $M_1$. With the definition of $y_t$ from \eqref{eq: model}, we then have 
     \begin{equation*}
         \Munbs_{1,n} = M_1 +\underbrace{ \frac{1}{n} \displaystyle \sum^n_{t=1} I\{a_t = 1\} \xi_t \bX_t / \pi_t}_{\widehat{Z}_1} +  \underbrace{\frac{1}{n} \displaystyle \sum^n_{t=1}\left(\frac{I\{a_t = 1\} \langle \Delta_{t-1}, \bX_t \rangle \bX_t}{\pi_t} - \Delta_{t-1}\right)}_{\widehat{Z}_2}.
     \end{equation*}
      Then both $\widehat{Z}_1$ and $\widehat{Z}_2$ are sum of martingale difference sequence by noting that for $\widehat{Z}_1$ 
    \begin{equation*}
         \mathbb{E}\left [ \frac{I\{ a_t = 1\} }{\pi_t}\xi_t \bX_t \Big | \mathcal{F}_{t-1}\right ]
         =  \mathbb{E}\left [ \mathbb{E} \left[ \frac{I\{ a_t = 1\} }{\pi_t}\xi_t \bX_t \Big | \mathcal{F}_{t-1}, \bX_t \right] \Big | \mathcal{F}_{t-1}\right] 
         = 0, 
    \end{equation*}
    and similarly for $\widehat{Z}_2$, Assumption \ref{assum: noise} implies that
    \begin{align*}
         & \mathbb{E}\left [ \frac{I\{a_t = 1\} \langle \Delta_{t-1}, X_t \rangle X_t}{\pi_t} - \Delta_{t-1} \Big | \mathcal{F}_{t-1} \right ] \\
         = & \mathbb{E}\left [ \frac{\langle \Delta_{t-1}, \bX_t \rangle \bX_t}{\pi_t} \mathbb{E} \left [ I\{a_t = 1\} \Big | \mathcal{F}_{t-1}, \bX_t \right ] - \Delta_{t-1} \Big | \mathcal{F}_{t-1} \right ] = 0.
    \end{align*}
    A similar debiasing procedure also applies to the case when $i = 0$ by replacing the $\pi_t$ by $(1-\pi_t)$ due to the fact that $\mathbb{E}[I\{a_t = 0\}|X_t,\mathcal{F}_{t-1}] = 1 - \pi_t$. We summarize the online debiasing procedure at each time $t$ in Algorithm \ref{alg:online de-bias}.
    \begin{algorithm}[t]
    \caption{One-Step Online Debiasing Update}
        \label{alg:online de-bias}
        \begin{algorithmic}[1]
        \State \textbf{Input}: $\Munbs_{i,t-1}$, $\Msgd_{i,t-1}$, for $i=0,1$, $X_t$,  $y_t$,  $\pi_t, a_t$ \vspace{2mm}
        
        \State \indent For $i = 0,1$, $\widetilde{M}_{i,t} \leftarrow \Msgd_{i,t-1} +  \frac{I\{a_t = i\}}{i \pi_t +(1-i)(1-\pi_t)} (y_t - \langle \Msgd_{i,t-1}, \bX_t \rangle )\bX_t$. \vspace{2mm}
        
        \State \indent $\Munbs_{i,t} \leftarrow ( \widetilde{M}_{i,t} + (t-1) \Munbs_{i,t-1} )/t$.\vspace{2mm}
        
        \State \textbf{Output}: $\Munbs_{1,t}$, $\Munbs_{0,t}$
     \end{algorithmic}
    \end{algorithm}
    
    As we mentioned earlier, the debiasing procedure eliminates both sources of bias simultaneously disregarding maintaining the low-rankness. In this case, $\Munbs_{i,n}$ obtained after $n$-iterations is not low-rank. Since the true parameter $M_i$ has a low-rank structure, we can apply a low-rank projection on the $\Munbs_{i,n}$ by its left and right top-$r$ singular vectors to yield an improved estimate for the inference purpose, which is denoted as $\Mlr_{i,n}$. Recall that we target to conduct the statistical inference on $m_T^{(i)} = \left \langle M_i,T\right \rangle$ that we discussed in Section \ref{sec: intro}, the corresponding estimator for the inference purpose is defined as 
    \begin{equation}
    \label{eq: small m hat}
        \widehat{m}_T^{(i)} = \left \langle \Mlr_{i,n}, T  \right \rangle.
    \end{equation}

     While $\Munbs_{i,n}$ serves as an unbiased estimator for $M_i$, it should be noted that $ \Mlr_{i,n}$ does not necessarily possess this property. In theory, we can show that this additional bias in $\widehat{m}_T^{(i)}$ is quantifiable and negligible under mild assumptions that we introduce in Section \ref{sec: asym norm}.  Moreover, to obtain $\Mlr_{i,n}$, we need to compute the SVD for a $d_1 \times d_2$ matrix $\Munbs_{i,n}$, and this computation is only required once after $n$-iterations. Because of its heavy computation cost, $\Mlr_{i,t}$ is not suitable for replacing the online estimator $\Msgd_{i,t}$ for the decision-making purpose as $\Msgd_{i,t}$ only requires computing the SVD of an $r \times r$ matrix at each iteration.
\subsection{Asymptotic normality of \texorpdfstring{$\widehat{m}^{(i)}_T$}{Lg}}
\label{sec: asym norm}
     We start the discussion on asymptotic normality by introducing several assumptions for the theoretical analysis. We denote $U_i$ and $V_i$ as the left and right singular vectors of the true matrix parameter $M_i$.
    \begin{assumption}
        \label{assum: null space}
        There exists a constant $\alpha_T > 0$ such that 
        \begin{equation*}
              \alpha_T \|T\|_{\mathrm{F}} \sqrt{\frac{r}{d_1}} \le \|U_i^\top T\|_{\mathrm{F}} ,\quad  \alpha_T \|T\|_{\mathrm{F}} \sqrt{\frac{r}{d_2}} \le \|TV_i \|_{\mathrm{F}}.
        \end{equation*}
    \end{assumption} 
    To perform statistical inference for $m_T^{(i)}=\langle M_i,T\rangle$, Assumption \ref{assum: null space} ensures that $T$ does not lie entirely in the null space of $M_i$ by imposing a lower bound on $\|U_i^\top T\|_{\mathrm{F}}$ and $\|TV_i \|_{\mathrm{F}}$. \begin{assumption}
        \label{assum: inco_assum}
        There exists a constant $\mu > 0$ such that, for $i \in \{0,1\}$,
        \begin{equation*}
            \max \Big \{ \sqrt{\frac{d_1}{r}}  \max_{j\in [d_1]} \| e_j^\top U_i \| , \sqrt{\frac{d_2}{r}} \ \max_{j\in [d_2]} \| e_j^\top V_i \|\Big \} \le \mu.
        \end{equation*}
    \end{assumption}
    Assumption \ref{assum: inco_assum} imposes an incoherence condition on the spectral space of the true matrix parameters $M_0,M_1$, indicating that their singular vectors should not be overly sparse. While not required to establish asymptotic normality, it simplifies the expression of the asymptotic distribution. Further discussion is provided in Section E.13 of the supplementary material.
    \begin{assumption}
    \label{assum: SNR condition}
        As $n, d_1, d_2 \rightarrow \infty$, assume
        \begin{equation*}
            \max \Big \{ \sqrt{\frac{dr\log^2d}{n^{\alpha}}},~~\frac{\sigma_i}{\lambda_r}\sqrt{\frac{d^2 r}{n}}
            \Big\} \rightarrow 0,
        \end{equation*}
        where $\sigma_i$ is defined in Assumption \ref{assum: noise}, and $\alpha\in(0,1)$ is specified in Theorem \ref{thm: sgd consistent}. In addition, there exist constants $\gamma$, $\gamma_d, \underline{\lambda}> 0$ such that $n = o(d^\gamma)$, $\lambda_r\geq \underline\lambda$, and $d_1/d_2+d_2/d_1 \le \gamma_d$.
    \end{assumption}
    { Assumption \ref{assum: SNR condition} requires conditions on the sample size and signal-to-noise ratio for reliable entry-level parameter inference. Under the additional assumption that the matrix $T$, which specifies the linear form under inference, is low-rank, the second condition may be relaxed to $(\sigma_i/\lambda_r)\sqrt{dr/n} = o(1)$. Section E.13 of the supplementary material outlines key supporting arguments for this relaxation, while a rigorous analysis is deferred to future work.}

    \begin{theorem}
    \label{thm1}
    Under Assumptions \ref{assum: noise}--\ref{assum: SNR condition} with $\beta=0$, and if we denote $\pi_t(X):= \mathbb{P}(a_t=1|\mathcal{F}_{t-1}, X_t=X)$ with $\pi_t(X) \xrightarrow{p}\pi_\infty(X)$ for any $X$. As $n,d_1,d_2 \rightarrow \infty$, we have
    \begin{equation*}
         \frac{\widehat{m}^{(i)}_T - m_T^{(i)}}{\sigma_iS_i/\sqrt{n}} \xrightarrow{d} \mathcal{N}\left(0, 1 \right), \quad i = 0,1,
    \end{equation*}
    where
    \begin{equation*}
        S^2_i = \int \frac{\Big \langle U_{i,\bot}U_{i,\bot}^\top \bX V_iV_i^\top+  U_iU_i^\top \bX V_{i,\bot} V^\top_{i,\bot}, T \Big \rangle ^2}{i\pi_{\infty}(X) + (1-i)(1-\pi_{\infty}(X))} dP_X,
    \end{equation*}
    \end{theorem}
    Theorem \ref{thm1} assumes $\beta=0$ in Assumption \ref{assum: decay rate}, requiring the policy to maintain a constant lower bound $\pmin$ for exploration. To ensure asymptotic normality of the parameter for each action, it mandates that each action is pulled sufficiently often to gather enough information for reliable parameter inference. As we will discuss in Section \ref{sec:optimal policy value}, the restriction on $\beta=0$ can be relaxed for the inference of optimal policy value. 
    
    { Theorem \ref{thm1} provides a key insight: incorporating a debiasing step improves the estimation rate to  $n^{-1/2}$. This improvement stems from the additional averaging performed during the debiasing procedure, which mitigates fluctuations across multiple iterates. As a result, the variance of the averaged sequence is reduced, leading to faster convergence. This acceleration behavior is analogous to the vector case studied in \cite{polyak1992acceleration}. }
     
    The above result allows us to derive the asymptotic normality of the difference between two estimators. The following corollary demonstrates the asymptotic behavior of the difference between $\widehat{m}_T^{(1)} - \widehat{m}_T^{(0)}$, and thus provides the theoretical guarantee for the hypothesis testing mentioned in \eqref{eq: hype test 2}.
    \begin{corollary}
    \label{co: the difference}
     Under Assumptions of Theorem \ref{thm1}, as $n,d_1,d_2 \rightarrow \infty$, we have
    \begin{equation*}
        \frac{\big( \widehat{m}_T^{(1)} - \widehat{m}_T^{(0)} \big) - \big (m_T^{(1)} - m_T^{(0)} \big)}{\sqrt{(\sigma_1^2S_1^2 + \sigma_0^2S_0^2)/n}} \xrightarrow{d} \mathcal{N}\left (0, 1 \right).
    \end{equation*}
    \end{corollary}
    The intuition of proving Corollary \ref{co: the difference} is that the main terms in $\widehat{m}_T^{(i)} - m_T^{(i)}$, $i=0,1$, are uncorrelated while the remainder terms are negligible. Therefore, the asymptotic variance of $( \widehat{m}_T^{(1)} - \widehat{m}_T^{(0)} ) -  (m_T^{(1)} - m_T^{(0)})$ is given by the sum of two individual variances. 
    
    \subsection{Parameter Inference}
    \label{sec: parameter inference}
    With the asymptotic normality shown in Theorem \ref{thm1}, we are in a position to answer the inferential question about $m_T^{(i)}$ by constructing an online data-dependent confidence interval. In this section, we show that the asymptotic normality of $\widehat{m}_T^{(i)}$ remains valid after we replace $S_i^2$ and $\sigma_i^2$ by their estimators. To achieve this goal, we only need to prove the consistency of the proposed variance estimator. 
     
    Throughout this section, we use $\Usp_{i,t}$ and $\Vsp_{i,t}$ to denote the left and right top-$r$ singular vectors of $\Msgd_{i,t}$, and $\Usp_{i,t \bot}$, $\Vsp_{i,t \bot}$ as their orthogonal complements. To obtain a consistent estimator of $S^2_i$ in Theorem \ref{thm1}, we need first to demonstrate that the $\Usp_{i,t}\Usp_{i,t}^\top$ and $\Vsp_{i,t}\Vsp_{i,t}^\top$ are consistent estimators for $U_iU_i^\top$ and $V_iV_i^\top$, where $U_i$ and $V_i$ denote the left and right top-$r$ singular vectors of $M_i$ respectively. Indeed, by the matrix perturbation theorem \citep{davis1970rotation, wedin1972perturbation},  for some positive constant $C$ we have
    \begin{equation*}
        \max\left\{ \|\Usp_{i,t}\Usp^{\top}_{i,t} - U_iU_i^\top\|_{\mathrm{F}},\|\Vsp_{i,t}\Vsp^{\top}_{i,t} - V_iV_i^\top\|_{\mathrm{F}} \right \} \le C \cdot \frac{\|\Msgd_{i,t} -M_i \|_{\mathrm{F}}}{\lambda_r}.
    \end{equation*}
    The convergence rate of $\Msgd_{i,t}$ shown in Theorem \ref{thm: sgd consistent} enables us to prove the consistency of the variance estimator, which leads to the following asymptotic normality of $\widehat{m}_T^{(i)}$ with the estimated $S^2_i$ and $\sigma_i^2$.
    \begin{theorem}
    \label{thm2}
    Under Assumptions of Theorem \ref{thm1}, as $n, d_1, d_2 \rightarrow \infty$, we have
    \begin{equation*}
        \frac{\widehat{m}_T^{(i)} - m_T^{(i)}}{\hat{\sigma}_i \hat{S}_i /\sqrt{n}} \xrightarrow{d} \mathcal{N}(0,1), \quad i = 0,1,
    \end{equation*}
    where
    \begin{equation}
    \label{eq: sigma estimate}
        \hat{\sigma}_i^2 =\frac{1}{n} \sum_{t = 1}^{n} \frac{I\{a_t = i \}}{i\pi_t + (1-i)(1 - \pi_t)}(y_t - \langle \Msgd_{i,t-1}, X_t \rangle)^2, 
    \end{equation}\vspace{-1.5em}
    \begin{equation}
    \label{eq: S estimate}
        \hat{S}_i^2\hspace{-.1em}=\hspace{-.1em} \frac{1}{n}\sum_{t=1}^n\hspace{-.18em} \frac{I\{ a_t =i\} \Big\langle \Usp_{i,t-1\bot}\Usp^{\top}_{i,t-1\bot} X_t \Vsp_{i,t-1}\Vsp^{\top}_{i,t-1}+\Usp_{i,t-1}\Usp^{\top}_{i,t-1} X_t \Vsp_{i,t-1\bot}\Vsp^{\top}_{i,t-1\bot} ,T\Big\rangle^2\hspace{-.18em}}{i\pi_t^2 + (1-i)(1-\pi_t)^2}.
    \end{equation}
    \end{theorem}
    It is worth pointing out that acquiring estimators $\hat{S}^2_i$ and $\hat{\sigma}^2_i$ only requires storing the partial sums instead of all historical data. At time $t$, estimators $\hat{S}^2_{i}$ and $\hat{\sigma}^2_i$ get updated by computing the running average of \eqref{eq: sigma estimate} and \eqref{eq: S estimate} for both $i = 0$ and $1$, and note that only $\Usp_{a_t,t-1}\Usp_{a_t,t-1}^\top$ and $\Vsp_{a_t,t-1}\Vsp_{a_t,t-1}^\top$ need to be calculated at each iteration. We present the method of obtaining $\Usp_{a_t,t-1}\Usp_{a_t,t-1}^\top$ and $\Vsp_{a_t,t-1}\Vsp_{a_t,t-1}^\top$ in the fourth to the last line inside the \emph{for} loop of Algorithm \ref{alg: online procedure}. Meanwhile, we can obtain the corresponding orthogonal complements used in \eqref{eq: S estimate} via 
    \begin{equation*}
         \Usp_{a_t,t-1\bot}\Usp^{\top}_{a_t,t-1\bot} = I - \Usp_{a_t,t-1}\Usp^{\top}_{a_t,t-1},
        \quad \text{and}\quad \Vsp_{a_t,t-1\bot}\Vsp^{\top}_{a_t,t-1\bot} = I - \Vsp_{a_t,t-1}\Vsp^{\top}_{a_t,t-1},
    \end{equation*}
    where $I$ denotes the identity matrix.
    
    Given the result of Theorem \ref{thm2}, we can thus construct the data-dependent confidence interval for the true parameter $m_T^{(i)}$. In particular, at any confidence level $\alpha\in (0,1)$ we can construct the confidence interval 
    \begin{equation}
    \label{eq: interval}
        \left[\widehat{m}^{(i)}_T - z_{\alpha/2} \hat{\sigma}_i \hat{S}_i/\sqrt{n},~ \widehat{m}^{(i)}_T + z_{\alpha/2} \hat{\sigma}_i \hat{S}_i/\sqrt{n}\right],
    \end{equation}
    where $z_{\alpha/2}$ denotes the standard score of normal distribution for the upper $\alpha/2$-quantile. The whole procedure of conducting the inference for $m_T^{(i)}$ is summarized in Algorithm \ref{alg: online procedure}.
    It is also worth pointing out that due to Corollary \ref{co: the difference}, we extend the result of Theorem \ref{thm2} to 
    \begin{equation*}
         \frac{(\widehat{m}_T^{(1)} - \widehat{m}_T^{(0)}) - (m_T^{(1)} - m_T^{(0)})}{\sqrt{(\hat{\sigma}^2_0 \hat{S}^2_0 + \hat{\sigma}^2_1 \hat{S}^2_1) /n}} \xrightarrow{d} \mathcal{N}(0,1),
    \end{equation*}
    which allows us to test the difference in effectiveness between the actions. 
    
    \begin{algorithm}[ht!]
    \caption{Online Inference of $m_T^{(i)}$}
        \label{alg: online procedure}
        \begin{algorithmic}[1]
             \State \textbf{Input}: $\Minit_1$, $\Minit_0$, $\Usgd_{i,0}$, $\Vsgd_{i,0}$ $r$.
             
             \State  \textbf{Initialization}: $\Munbs_{i,0} \leftarrow \Minit_i$, $\Msgd_{i,0} \leftarrow \Minit_i$, for $i = 0,1$.
            
            \State \For{$t\gets 1 $ to $n$}{
            
             Observe a contextual matrix $X_t$.
        
             Compute $\pi_t$ according to the policy. 
             
             Decide the action $a_t$ by $Ber(\pi_t)$.
        
             Receive reward $y_t$ according to \eqref{eq: model}.

             For $i = 0,1$, $\Munbs_{i,t} \leftarrow$ Algorithm \ref{alg:online de-bias} ($\Munbs_{i,t-1}$, $\Msgd_{i,t-1}$, $X_t$, $y_t$, $a_t$, $\pi_t$).
             
             $\Usgd_{i,t}$, $\Vsgd_{i,t}, R_{\Usgd}$, $D_{\Usgd}$, $R_{\Vsgd}$, $D_{\Vsgd} \leftarrow$  Algorithm \ref{alg:sgd update practice} ($\Usgd_{i,t-1}$, $\Vsgd_{i,t-1}$, $X_t$, $y_t$, $a_t$, $\pi_t$).

             $\Usp_{a_t,t-1}\Usp_{a_t,t-1}^\top  \leftarrow R_{\Usgd}D_\Usgd^{-1}R_{\Usgd}^\top$, ~~ $\Vsp_{a_t,t-1}\Vsp_{a_t,t-1}^\top  \leftarrow R_{\Vsgd}D_\Vsgd^{-1}R_{\Vsgd}^\top$.
             
             $\Usp_{a_t,t-1\bot}\Usp^{\top}_{a_t,t-1\bot} \leftarrow I - \Usp_{a_t,t-1}\Usp^{\top}_{a_t,t-1}$, ~~ 
             $\Vsp_{a_t,t-1\bot}\Vsp^{\top}_{a_t,t-1\bot} \leftarrow I - \Vsp_{a_t,t-1}\Vsp^{\top}_{a_t,t-1}$.
             
              Update $\hat{\sigma}^2_i$ and $\hat{S}^2_i$ by computing the running average of \eqref{eq: sigma estimate} and \eqref{eq: S estimate}.  
             
             $\Msgd_{i,t} \leftarrow \Usgd_{i,t}\Vsgd_{i,t}^\top$.}
        \State Compute the top-$r$ singular vectors of $\Munbs_{i,n}$ to obtain $\Mlr_{i,n}$, and then we calculate $\widehat{m}^{(i)}_T$ by \eqref{eq: small m hat}.
        \State Obtain the confidence interval as \eqref{eq: interval}.
        \end{algorithmic}
    \end{algorithm}

\section{Inference for Optimal Policy Value}
\label{sec:optimal policy value}

In this section, we investigate the statistical inference of optimal policy value as defined in \eqref{eq: optimal value}. In contrast with Section \ref{sec: Inference}, which requires the exploration probability to be lower bounded by constant, we relax this condition by permitting the exploration probability to gradually diminish over time for optimal policy value inference. Echoing the debiasing technique outlined in Equation \eqref{eq: Munbs 1} from Section \ref{sec: online de-bias}, we adopt a similar strategy to develop an estimator for inferring the optimal policy value. The construction of this estimator also incorporates a correction term designed for bias reduction. Due to space limitations, this section focuses on scenarios where exploration probabilities are known. We defer the optimal policy value inference procedure when these probabilities are unknown yet estimated to Section A of the supplementary material.

\subsection{Estimator for Optimal Policy Value}

We now present our estimator for the optimal policy value. This estimator after $n$ iterations is defined as follows: 
\begin{equation}
\label{eq:DR est}
\widehat{V}_n = \frac{1}{n}\sum_{t=1}^n \left \langle \Msgd_{\hat{a}(X_t), t-1}, X_t \right \rangle + \frac{1}{n}\sum_{t=1}^n \frac{I\{ a_t = \hat{a}(X_t)\}}{1 - e_t}\left(y_t - \left \langle \Msgd_{\hat{a}(X_t), t-1}, X_t \right\rangle \right),
\end{equation}
where
\begin{equation}
    \label{eq:a_hat}
     \hat{a}(X_t) = I\{\langle \Msgd_{1, t-1} - \Msgd_{0,t-1}, X_t \rangle > 0\}, 
\end{equation}
and $e_t := 1 - \bP(a_t =\hat{a}(X_t)|\mathcal{F}_{t-1},X_t)$.
In the formation of this optimal policy value estimator, $\hat{a}(X_t)$ represents the estimated optimal action at time $t$, and $e_t$ represents the probability for exploration. To elaborate, if $\hat{a}(X_t) = 1$, the exploration probability becomes $e_t = \bP(a_t=0|\mathcal{F}_{t-1}, X_t) = 1- \pi_t$. Similar to the debiasing process used in parameter inference described in \eqref{eq: Munbs 1}, we also employ inverse probability weighting to correct distributional bias in this scenario. However, there is a key distinction: in parameter inference, the weighting factor is derived from the probability of taking each possible action, while here it suffices to use only the exploitation probability for the inverse weighting. This distinction arises because bias correction in parameter inference leverages samples gathered from each action individually. In the case of the optimal policy value estimator, however, we exclusively use samples collected from the estimated optimal action, regardless of whether it is action $1$ or $0$, to formulate this bias reduction. This forms the key reason that we allow a relaxed exploration probability in this section. 

In Equation \eqref{eq:DR est}, we can view the first term as a direct estimator for the optimal policy value. However, relying on this direct estimate exclusively can lead to potential failure when $\Msgd_{i,t}$ does not offer an accurate estimate of $M_i$. In the context of our study, where $\Msgd_{i,t}$ is inherently biased, the latter term of \eqref{eq:DR est} serves as a corrective mechanism, functioning in a manner analogous to how we formulated $\Munbs_{i,t}$ in Section \ref{sec: Inference}. For optimal policy value inference, samples contributed to the estimation should be selectively obtained from the exploitation part, which explains the reason that our estimator presented in \eqref{eq:DR est} only takes the samples generated by the estimated optimal action.

\subsection{Asymptotic Normality}
We start the discussion on the asymptotic normality of the optimal policy value estimator \eqref{eq:DR est} by introducing the following assumptions. 
\begin{assumption}
    \label{assum:Optimal-Gap}
    For $\alpha$ in the learning rate specified in Theorem \ref{thm: sgd consistent} and $\beta$ specified in Assumption \ref{assum: decay rate} such that $\alpha-\beta>\frac12$, as $n, d_1, d_2 \rightarrow \infty$,
    \begin{equation*}
        \max\Big\{ \sqrt{\frac{dr\log^2d}{n^{\alpha-\beta}}}, ~~ \frac{\sigma_i\|M_1 -M_0\|_{\mathrm{F}}^{-1} dr\log^2d}{n^{\alpha-\beta-\frac12}}\Big\} \rightarrow 0.
    \end{equation*}
    In addition, there exist constants $\gamma$, $\gamma_d > 0$ such that $n = o(d^\gamma)$ and $d_1/d_2+d_2/d_1 \le \gamma_d$.
\end{assumption}
 Assumption \ref{assum:Optimal-Gap} consists of two components: the first part ensures that $\Msgd_i$ serves as a consistent estimator of $M_i$, and the second condition ensures that the gap between $M_1$ and $M_0$ is sufficiently large compared to the noise, making the optimal action distinguishable. With these considerations, we are now prepared to discuss the asymptotic normality of $\sqrt{n}( \widehat{V}_n - V^*)$. 
 \begin{theorem}
\label{thm:value inference}
Under the conditions of Theorem \ref{thm: sgd consistent} and Assumption \ref{assum:Optimal-Gap}, if we denote $e^*_t(X) = \bP(a_t \ne a^*(X_t)|\mathcal{F}_{t-1},X_t = X)$ with $e^*_t(X) \xrightarrow{p} e^*_{\infty}(X)$ for any $X$. Then as $n,d_1,d_2 \rightarrow \infty$, we have
    \begin{equation*}
       \frac{ \widehat{V}_n - V^* }{S_V/\sqrt{n}} \xrightarrow{d} \mathcal{N}\left( 0, 1\right),
    \end{equation*}
    where
    \begin{equation*}
        S_V^2 = \int \frac{a^*(X)\sigma_1^2 + (1-a^*(X))\sigma_0^2}{1 - e^*_{\infty}(X)}dP_X + \mathrm{Var}_X\left[ \langle M_{a^*(X)}, X\rangle\right].
    \end{equation*}
\end{theorem}
Theorem \ref{thm:value inference} establishes the asymptotic normality of our proposed optimal policy value estimator. This asymptotic variance consists of two distinct components. The first term in $S_V^2$ serves as the weighted average variance of the noise, conditional on the optimal action for a given context. On the other hand, the second term in $S_V^2$ captures the variance associated with the context. If the estimated optimal action $\hat{a}(X_t)$ converges to the true optimal action $a^*(X_t)$, then the weight assigned to the first component of $S_V^2$ is determined by the limiting probability associated with exploitation. Note that the asymptotic probability of exploration $e^*_{\infty}(X)$ is allowed to be zero in this scenario, which marks the fundamental difference from the parameter inference in Theorem \ref{thm1}. 

\subsection{Optimal Policy Value Inference}

With the asymptotic normality introduced in Theorem \ref{thm:value inference}, we next construct a valid confidence interval for the optimal policy value. We first propose the empirical estimator for $S^2_V$ in a fully online fashion without requiring any storage for $d_1 \times d_2$ context matrix $X_t$. Define the online estimator as 
\begin{align*}
    \widehat{S}^2_V = & \frac{1}{n} \sum_{t = 1}^n \frac{\hat{\sigma}^2_{1,t} I\big \{\big \langle \Msgd_{1,t-1} - \Msgd_{0,t-1}, X_t\big \rangle > 0\big \} + \hat{\sigma}^2_{0,t} I\big \{\big \langle \Msgd_{1,t-1} - \Msgd_{0,t-1}, X_t\big \rangle \le 0 \big \}}{1 - e_t} \\
    & + \frac{1}{n}\sum_{t=1}^n \big\langle \Msgd_{\hat{a}(X_t), t-1}, X_t \big\rangle^2 - \Big(\frac{1}{n}\sum_{t=1}^n \big\langle \Msgd_{\hat{a}(X_t), t-1}, X_t \big\rangle \Big)^2,\numberthis 
    \label{eq:emp SV}
\end{align*}
where for $i = 0,1$,
\begin{equation}
\label{eq:emp sigma}
    \hat{\sigma}^2_{i,t} = \frac{1}{t}\sum_{s=1}^t \frac{I\{a_s = i \}}{i\pi_s + (1-i)(1 - \pi_s)} \left(y_s - \left\langle \Msgd_{i,s-1},X_s\right \rangle\right)^2.
\end{equation}
It is important to note that the running summation in \eqref{eq:emp SV} and \eqref{eq:emp sigma} can be sequentially updated. Theorem \ref{thm:consistent} below shows that $\widehat{S}^2_V$ is a consistent estimator for $S_V^2$, and thus the asymptotic normality is also guaranteed with the estimated variance.
\begin{theorem}
\label{thm:consistent}
Under the conditions of Theorem \ref{thm:value inference}, we have $\widehat{S}^2_V$ is a consistent estimator of $S^2_V$, i.e., $\widehat{S}^2_V \xrightarrow{p} S^2_V$. Furthermore, as $n, d_1, d_2 \rightarrow\infty$, we have
 \begin{equation*}
        \frac{ \widehat{V}_n - V^* }{\widehat{S}_V/\sqrt{n}} \xrightarrow{d} \mathcal{N}\left( 0, 1\right).
    \end{equation*}
\end{theorem}
In light of Theorem \ref{thm:consistent}, constructing a confidence interval for the optimal policy value $V^*$ becomes feasible. This opens the door to hypothesis testing to evaluate the performance of the currently available actions in achieving a desired level of outcome, even under the optimal policy. This addresses inferential questions posed in Equation \eqref{eq: hype test 3}. Unlike the parameter inference discussed in Section \ref{sec: Inference}, which necessitates computing the SVD for a $d_1 \times d_2$ matrix at the end of the online sequence for low-rank projection, the value inference approach introduced in this section sidesteps the computational overhead associated with SVD calculations. Finally, we summarize the optimal policy value inference procedure in Algorithm \ref{alg: optimal value inference p}.

\begin{algorithm}[t]
    \caption{Online Inference of Optimal Policy Value $V^*$}
    \label{alg: optimal value inference p}
        \begin{algorithmic}[1]
        \State \textbf{Input}: $\Minit_1$, $\Minit_0$, $\Usgd_{i,0}$, $\Vsgd_{i,0}$,  $r$.
        
        \State  \textbf{Initialization}: $\Msgd_{i,0} \leftarrow \Minit_i$, for $i = 0,1$.

        \State \For{$t\gets 1 $ to $n$}{

            Observe a contextual matrix $X_t$.

            Obtain $\pi_t = \bP(a_t = 1|\mathcal{F}_{t-1}, X_t)$ according to the decision-making policy. 

            Update $\hat{a}(X_t)$ by equation \eqref{eq:a_hat}, and calculate $e_t \leftarrow 1- \bP(a_t = \hat{a}(X_t)|\mathcal{F}_{t-1},X_t)$.

            Decide the action $a_t$ by $Ber(\pi_t)$.

            $\Usgd_{i,t}$, $\Vsgd_{i,t} \leftarrow$  Algorithm \ref{alg:sgd update practice} ($\Usgd_{i,t-1}$, $\Vsgd_{i,t-1}$, $X_t$, $y_t$, $a_t$, $\pi_t$)

            $\Msgd_{i,t} \leftarrow \Usgd_{i,t}\Vsgd_{i,t}^\top$.

            Get the estimator value $\widehat{V}_t$ by equation \eqref{eq:DR est}.

            Update the variance estimator $\widehat{S}^2_V$ by equation \eqref{eq:emp SV}.
        }
        \State Obtain the two-sided confidence interval with critical value $z$: $(\widehat{V}_n - z\widehat{S}_V/\sqrt{n},~~ \widehat{V}_n + z\widehat{S}_V/\sqrt{n} )$.
        \end{algorithmic}
\end{algorithm}

\section{Simulation Studies}
\label{sec: experiment}
In this section, we present extensive numerical studies to evaluate the performance of our online inference procedure. In the presented synthetic simulations, we consider a Gaussian noise $\xi_t|a_t = i \sim N(0, \sigma_i^2)$ with the noise level $\sigma_i = 0.1$ for both $i=0,1$. We generate the true low-rank matrices $M_1$ and $M_0$ with rank $r=3$, and dimensions $d = d_1 =d_2= 50$. The singular vectors, $U_i, V_i \in \mathbb{R}^{d\times r}$, are generated from the singular space of random Gaussian matrices. We set top-$r$ singular values of $M_i$ to be $1$, i.e., $\lambda_1(M_i) = \lambda_2(M_i) = \lambda_3(M_i) = 1$. For the simulation study of the parameter inference, we adopt $\varepsilon$-greedy policy with $\varepsilon=0.1$. The additional simulation results for optimal value inference with $\varepsilon \rightarrow 0$ are illustrated in Section B of the supplementary material. We set the learning rate $\eta_t = 0.1(\max\{t,t^\star\})^{-0.99}$ with $t^\star=300$. Finally, the initialization $\Minit_i$ is obtained from a nuclear-norm penalized estimation \citep{negahban2011estimation} with pre-collected offline data.

\begin{table}
\centering
\arrayrulecolor{black}
\caption{Coverage Probability, Average Confidence Interval Length and corresponding standard deviation for the scenario $T = T_1$ and $T = T_2$ based on $5000$ independent trails. }
\begin{tabular}{c|cc|c|c} 
\hline
\multicolumn{1}{l}{~} & \multicolumn{1}{l}{~}     & \multicolumn{1}{l|}{~} & Coverage Probability       & Average CI Length  \\ 
\hline
\multirow{6}{*}{$T_1$}   & \multirow{2}{*}{$n = 1000$} & $i=0$                    & $0.909$                      & $0.018$              \\
                      &                           & $i=1$                    & $0.913$                      & $0.010$               \\ 
\cline{2-5}
                      & \multirow{2}{*}{$n = 2000$} & $i=0$                    & $0.923$                      & 0.013              \\
                      &                           & $i=1 $                   & $0.925$                      & 0.008              \\ 
\cline{2-5}
                      & \multirow{2}{*}{$n = 3000$} & $i=0$                    & $0.929$                      & $0.011$              \\
                      &                           & $i=1$                    & $0.936$                      & $0.006$              \\ 
\hline
\multirow{6}{*}{$T_2$}   & \multirow{2}{*}{$n = 1000$} & $i=0 $                   & \multicolumn{1}{c|}{$0.906$} & $0.065$              \\
                      &                           & $i=1$                    & \multicolumn{1}{c|}{$0.908$} & $0.042$              \\ 
\cline{2-5}
                      & \multirow{2}{*}{$n = 2000$} & $i=0$                    & \multicolumn{1}{c|}{$0.924$} & $0.048$              \\
                      &                           & $i=1$                    & \multicolumn{1}{c|}{$0.923$} & $0.031$              \\ 
\cline{2-5}
                      & \multirow{2}{*}{$n = 3000$} & $i=0$                    & \multicolumn{1}{c|}{$0.931$} & $0.039 $             \\
                      &                           & $i=1 $                   & \multicolumn{1}{c|}{$0.930$}  & $0.026$              \\
\arrayrulecolor{black}\cline{1-1}\arrayrulecolor{black}\cline{2-5}
\end{tabular}
\label{tab: T1 and T2}
\arrayrulecolor{black}
\end{table}

\begin{table}
\centering
\arrayrulecolor{black}
\caption{Coverage Probability, Average Confidence Interval Length for $r = 3,5,7$ for $T = T_1$ and $n=3000$ based on $5000$ independent trails.}
\begin{tabular}{cc|c|c}
\hline
\multicolumn{1}{l}{~} & \multicolumn{1}{l|}{~} & Coverage Probability & Average CI Length  \\ 
\hline
\multirow{2}{*}{$r=3$}  & $i=0 $                   & $0.929$                & $0.011 $             \\
                      & $i=1 $                   & $0.936 $               & $0.006$              \\ 
\arrayrulecolor{black}\cline{1-1}\arrayrulecolor{black}\cline{2-4}
\multirow{2}{*}{$r=5$}  & $i=0 $                   & $0.917 $               & $0.015 $             \\
                      & $i=1 $                   & $0.921 $               & $0.014$              \\ 
\arrayrulecolor{black}\cline{1-1}\arrayrulecolor{black}\cline{2-4}
\multirow{2}{*}{$r=7$}  & $i=0 $                   & $0.913 $               & $0.021$              \\
                      & $i=1 $                   & $0.906$                & $0.021 $         \\
\hline
\end{tabular}
\label{tab: diff r}
\arrayrulecolor{black}
\end{table}

\begin{figure}
     \centering
     \begin{subfigure}{0.33\linewidth}
         \includegraphics[width=\linewidth]{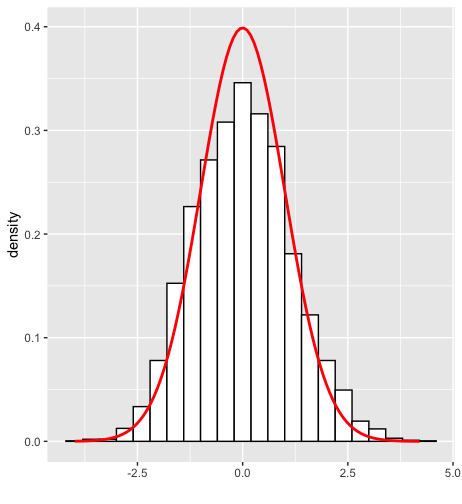}
         \caption{$n = 1000$, $r=3$}
     \end{subfigure}
     \hspace{4em}
     \begin{subfigure}{0.33\linewidth}
         \includegraphics[width=\linewidth]{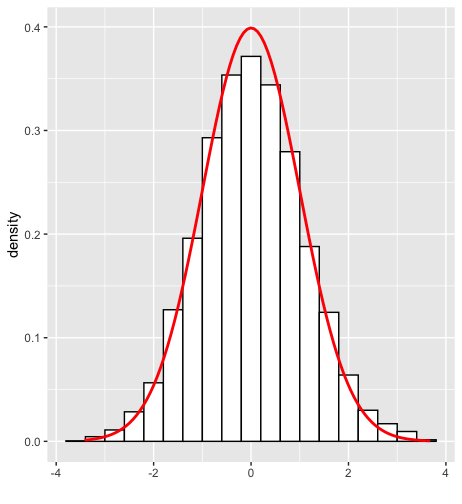}
         \caption{$n = 3000$, $r=3$}
     \end{subfigure}
     \caption{Empirical distribution of $\sqrt{n}(\widehat{m}^{(1)}_T - m^{(1)}_T)/\hat{\sigma}_1\hat{S}_1$ based on $5000$ independent trails for $T = e_1e_1^\top$. The red curve refers to the density of standard normal. }\label{fig:1}
\end{figure}

\begin{figure}
     \centering
     \begin{subfigure}{0.32\linewidth}
         \includegraphics[width=\linewidth]{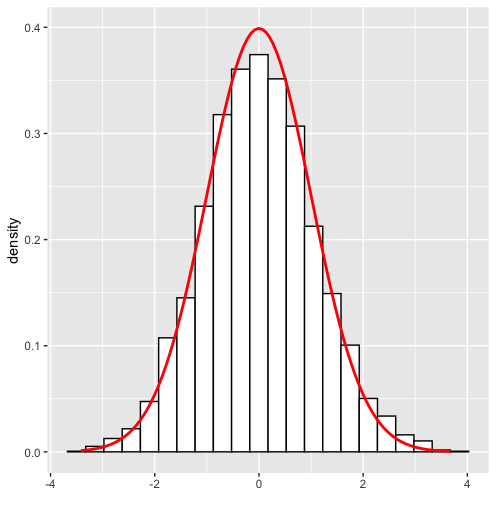}
         \caption{$n = 3000$, $r=3$}
     \end{subfigure}
     \hfill
     \begin{subfigure}{0.32\linewidth}
         \includegraphics[width=\linewidth]{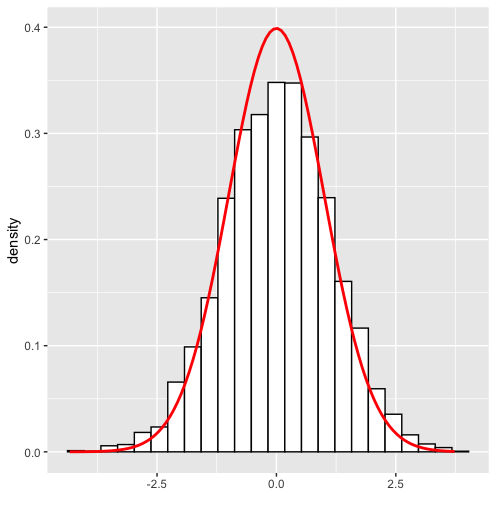}
         \caption{$n = 3000$, $r=5$}
     \end{subfigure}
     \hfill
     \begin{subfigure}{0.32\linewidth}
         \includegraphics[width=\linewidth]{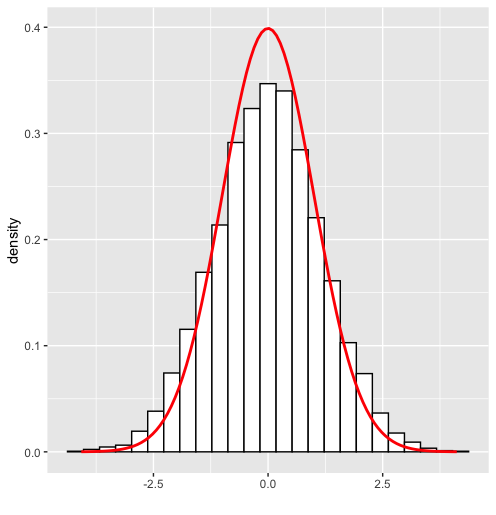}
         \caption{$n = 3000$, $r=7$}
     \end{subfigure}
     \caption{Empirical distribution of $\sqrt{n}(\widehat{m}^{(1)}_T - m^{(1)}_T)/\hat{\sigma}_1\hat{S}_1$ based on $5000$ independent trails for ranks $r=3,~5,~7$ and $T = e_1e_1^\top$.}\label{fig:2}
\end{figure}

We first validate the asymptotic normality of $\widehat{m}_T^{(i)}$ with $T = e_1e_1^\top$ by plotting the histogram of $\sqrt{n}(\widehat{m}^{(i)}_T - m^{(i)}_T)/\hat{\sigma}_i\hat{S}_i$ from $5000$ independent trails with $n = 1000$ and $3000$. We present the histogram of $\sqrt{n}(\widehat{m}^{(i)}_T - m^{(i)}_T)/\hat{\sigma}_i\hat{S}_i$ for $i=1$ in Figure \ref{fig:1}. The result for $i=0$ is similar and hence is omitted. As shown in Figure \ref{fig:1}, as $n$ increases, the empirical distribution of $\sqrt{n}(\widehat{m}^{(i)}_T - m^{(i)}_T)/\hat{\sigma}_i\hat{S}_i$ gets closer to the standard normal distribution.

In Table \ref{tab: T1 and T2}, we present the coverage probability and average confidence interval length in two scenarios with $T=T_1 = e_1e_1^\top$ and $T=T_2 = e_{1}e^\top_1 + 2e_2e^\top_{2} - 3 e_{3}e^\top_{3}$. The coverage probability is calculated as the ratio of the $5000$ independent trails that fall into $\big ( \widehat{m}_T^{(i)} - 1.96 \hat{\sigma}_i \hat{S}_i,  \widehat{m}_T^{(i)} + 1.96 \hat{\sigma}_i \hat{S}_i \big)$, which is the $95\%$ confidence interval constructed by the standard deviation estimation. The interval length is calculated as $2 \times 1.96 \hat{\sigma}_i \hat{S}_i$. We present the result as $n = 1000$, $2000$, and $3000$. As shown in Table \ref{tab: T1 and T2}, for both $T_1$ and $T_2$, as $n$ grows, the coverage probability is closer to $0.95$, and the confidence interval length decreases. In addition, when we increase the $\|T\|_{\mathrm{F}}$, i.e., from $\|T_1\|_{\mathrm{F}}$ to $\|T_2\|_{\mathrm{F}}$, the true $S_i$ gets larger which causes the average length of confidence interval increases. 

In Table \ref{tab: diff r}, we compare the converge probability and the average confidence interval lengths across different true ranks $r$. As the rank $r$ increases, the coverage probability shrinks, and the confidence interval length increases. We also compare the histograms for $r =3,5,7$ in Figure \ref{fig:2}, and the normal approximation gets slightly worse as the true rank increases.

\bibliographystyle{chicago}
\bibliography{ref}

\newpage
\appendix

\section{Optimal Policy Value Inference with Unknown Exploration Probability}
\label{app opt val}
In the main paper, we consider the case that the probability of action selection is known in the decision-making policy. In this section, we further relax this requirement and discuss the optimal policy value inference procedure when such probability can not be explicitly obtained, meaning it is necessary to estimate exploration probability empirically. When the probability for choosing each action is not explicitly known, the condition on $\min\{\pi_t, 1-\pi_t\}$ is impractical. Instead, we impose a clipping rate on sample realizations to ensure that each action receives an adequate sample size for estimation, stated in Assumption \ref{assum:clipping}.
\begin{assumption}
         \label{assum:clipping}
         There exist constants $p_0>0$ and $0\le \beta <1$ such that for $t>1$, 
         \begin{equation*}
             \min\left\{\sum_{s=1}^t I\{a_s = 0\}, ~~ \sum_{s=1}^t I\{a_s = 1\}\right\} > p_0t^{1-\beta}.
         \end{equation*}
\end{assumption}
The above assumption ensures that neither action should gather fewer than $p_0t^{1-\beta}$ samples up to time $t$. 
This condition can be satisfied with a ``force the exploration'' step in Algorithm \ref{alg: optimal value inference np}. Assumption \ref{assum:clipping} extends Assumption \ref{assum: decay rate} from the known exploration probability case to the case of unknown exploration probability. It reflects the commonly assumed clipping rate condition in literature \citep{deshpande2018accurate,zhang2020inference,shen2021doubly,shi2023statistical}. 

Since $\pi_t$ cannot be explicitly expressed in this scenario, we introduce a modification to our low-rank estimation method originally proposed in Section \ref{sec: low rank estimation}. Revisiting the naive SGD update outlined in \eqref{eq: naive update}, the update is applied to either $\Msgd_1$ or $\Msgd_0$ based on the action taken. Without altering the objective function given by \eqref{eq: pop loss}, we modify the update rule by merely counting the number of updates for each low-rank estimator, which still aligns with the goal of optimizing $F(\Usgd_i, \Vsgd_i)$ for each $i$. Specifically, we employ indices $s_1$ and $s_0$ to monitor the number of updates made for estimating $M_1$ and $M_0$, respectively. Formally, we set $s_i = \sum_{\tau=1}^{t-1} I\{a_\tau =i\}$, denoting the number of updates applied to $\Usgd_i$ and $\Vsgd_i$ prior to the $t$-th iteration. Taking into account the re-normalization trick discussed in Section \ref{sec: explain stochastic gradient}, we replace the representation of the stochastic gradient in \eqref{eq: practice gradient} accordingly by
\begin{equation}
    \label{eq:g modify}
    \tilde{g}\left( \Usgd_{i,s_i}, \Vsgd_{i,s_i}; \{X_t, y_t \}\right) = \left(\begin{array}{l}
       (\langle \Usgd_{i,s_i}\Vsgd_{i,s_i}^\top , X_{t}\rangle - y_t)X_{t}\Vsgd_{i,s_i}R_{\Vsgd}D_{\Vsgd}^{-\frac{1}{2}}Q_{\Vsgd}Q_{\Usgd}^\top D_{\Usgd}^{\frac{1}{2}}R_{\Usgd}^\top \\
        (\langle \Usgd_{i,s_i}\Vsgd_{i,s_i}^\top , X_{t}\rangle-y_t)X_{t}\Usgd_{i,s_i}R_{\Usgd}D_{\Usgd}^{-\frac{1}{2}}Q_{\Usgd}Q_{\Vsgd}^\top D_{\Vsgd}^{\frac{1}{2}}R_{\Vsgd}^\top
        \end{array}\right).
\end{equation}
With the gradient formally defined in \eqref{eq:g modify}, the one-step update for online estimation with  $\pi_t$ unknown is described in Algorithm \ref{alg:sgd update modify}. For the $\Msgd_{i,t}$ generated by Algorithm \ref{alg:sgd update modify} at each time $t$, the subsequent corollary outlines the convergence behavior of the low-rank estimator.
\begin{corollary}
    \label{cor:sgd consistent}
   Given the conditions in Theorem \ref{thm: sgd consistent} and Assumption \ref{assum:clipping}, we define the learning rate $\eta_{s_i} = c \cdot (\max\{s_i, s^\star \})^{-\alpha}$, where $s^\star = \left ( \gamma^2dr \log^2d\right )^{1/\alpha}$. Then, with probability at least $1 - \frac{4n}{d^{\gamma}}$, we have for any $1 < t \leq n$,
    \begin{equation*}
        \left\| \Msgd_{i,t} - M_i \right \|_{\mathrm{F}} \le C_1 \gamma \sigma_i \sqrt{\frac{dr\log^2 d}{t^{\alpha-\beta}}},
    \end{equation*}
    for some positive constant $C_1$.
\end{corollary}

 \begin{algorithm}[t]
     \caption{One-Step SGD Update at time $t$ with Unknown $\pi_t$}
        \label{alg:sgd update modify}
        \begin{algorithmic}[1]
        \State \textbf{Input}: $\Usgd_{i,s_i}$, $\Vsgd_{i,s_i}$, $s_i$ for $i = 0,1$, $X_t$, $y_t$, $a_t$ \vspace{2mm}
        
         \State \hspace{4mm} $R_{\Usgd}D_{\Usgd}R^\top_{\Usgd} \leftarrow$ SVD $\left (\Usgd_{a_t,s_{a_t}}^\top \Usgd_{a_t,s_{a_t}}\right)$ , $R_{\Vsgd}D_{\Vsgd}R^\top_{\Vsgd} \leftarrow$ SVD $\left (\Vsgd_{a_t,s_{a_t}}^\top \Vsgd_{a_t,s_{a_t}}\right)$. \vspace{2mm}

         \State \hspace{4mm} $Q_{\Usgd}DQ_{\Vsgd} \leftarrow$ SVD$\left (D_{\Usgd}^{\frac{1}{2}}R^\top_{\Usgd}R_{\Vsgd}D_{\Vsgd}^{\frac{1}{2}}\right)$. \vspace{2mm}

         \State \If{$a_t =1$}{
        $\left(\begin{array}{l}
        \Usgd_{1,s_1+1}\\
        \Vsgd_{1,s_1+1}
        \end{array}\right) = \left(\begin{array}{l}
        \Usgd_{1,s_1}\\
        \Vsgd_{1,s_1}
        \end{array}\right) - \eta_{s_1}\tilde{g}\left( \Usgd_{1,s_1}, \Vsgd_{1,s_1}; \{X_t, y_t \}\right)$\; \vspace{2mm}

        $s_1 = s_1 + 1$\; }
        \State \Else{
        $\left(\begin{array}{l}
        \Usgd_{0,s_0+1}\\
        \Vsgd_{0,s_0+1}
        \end{array}\right) = \left(\begin{array}{l}
        \Usgd_{0,s_0}\\
        \Vsgd_{0,s_0}
        \end{array}\right) - \eta_{s_0}\tilde{g}\left( \Usgd_{0,s_0}, \Vsgd_{0,s_0}; \{X_t, y_t \}\right)$\; \vspace{2mm}

        $s_0 = s_0 + 1$\; }\vspace{2mm}
        
        \State \textbf{Output}: $\Usgd_{i,s_i}$, $\Vsgd_{i,s_i}$, $\Msgd_{i,t} \leftarrow \Usgd_{i,s_i}\Vsgd_{i,s_i}^\top$, $s_i$ for both $i=0,1$.
        \end{algorithmic}
    \end{algorithm}

Remind that the exploration probability at each time $t$ is $e_t= 1 - \bP(a_t =\hat{a}(X_t)|\mathcal{F}_{t-1},X_t)$ where $\hat{a}(X_t) = I\{\langle \Msgd_{1, t-1} - \Msgd_{0,t-1}, X_t \rangle > 0\}$. In cases where $e_t$ is not known, it can be estimated empirically using historical data, denoted as $\hat{e}_t$. Utilizing this estimation, we then formulate the optimal policy value estimator for inference purposes as follows:
\begin{equation}
\label{eq:DR est np}
\widehat{V}_n = \frac{1}{n}\sum_{t=1}^n\left \langle \Msgd_{\hat{a}(X_t), t-1}, X_t \right \rangle + \frac{1}{n}\sum_{t=1}^n \frac{I\{ a_t = \hat{a}_t(X_t)\}}{1 - \hat{e}_t}\left(y_t - \left \langle \Msgd_{\hat{a}(X_t), t-1}, X_t \right\rangle \right) .
\end{equation}
To control its estimation error of $\hat e_t$, we require an additional assumption. 
\begin{assumption}
    \label{assum:DR Assumption}
    For $i = 0,1$, 
    \begin{equation*}
        \bE_X \left\vert(\hat{e}_t - e_t) \left \langle M_i - \Msgd_{i,t-1},X_t\right \rangle \right\vert = o_p(\sigma_i t^{-\frac{1}{2}}).
    \end{equation*}
\end{assumption}
Assumption \ref{assum:DR Assumption}, often referred to as the double robust property, is frequently invoked in the causal inference literature when the weighting probability is not directly observable \citep{bang2005doubly,luedtke2016statistical,shen2021doubly}. This assumption ensures convergence of the product of the estimated probability for exploration and the estimated reward function at a certain rate, which is crucial for establishing the asymptotic distribution of $\widehat{V}_n$. Additionally, this assumption offers a protection against imprecise estimation by ensuring that the accuracy of either one of the two estimators is sufficient for reliable results. Building on this, we further explore the asymptotic normality of $\sqrt{n}(\widehat{V}_n - V^*)$, which is central to conducting hypothesis tests for the estimated optimal value.

\begin{theorem}
\label{thm:value inference np}
    Under conditions of Theorem \ref{thm:value inference}, Corollary \ref{cor:sgd consistent}, and Assumption \ref{assum:DR Assumption}, we have as $n, d_1,d_2 \rightarrow \infty$, 
    \begin{equation*}
         \frac{\widehat{V}_n - V^*}{S_V/\sqrt{n}}  \xrightarrow{d} \mathcal{N}\left( 0, 1\right),
    \end{equation*}
    for $S^2_V$ defined in Theorem \ref{thm:value inference}.
\end{theorem}
We then define the estimator for $S_V^2$ with estimation $\hat{e}_t$ as
\begin{align*}
\label{eq:emp SV np}
    \widehat{S}^2_V = & \frac{1}{n} \sum_{t = 1}^n \frac{\hat{\sigma}^2_{1,t} I\left \{\left \langle \Msgd_{1,t-1} - \Msgd_{0,t-1}, X_t\right \rangle > 0\right \} + \hat{\sigma}^2_{0,t} I\left \{\left \langle \Msgd_{1,t-1} - \Msgd_{0,t-1}, X_t\right \rangle \le 0 \right \}}{1 - \hat{e}_t}\numberthis \\
    & + \frac{1}{n}\sum_{t=1}^n \left\langle \Msgd_{\hat{a}(X_t), t-1}, X_t \right\rangle^2 - \left(\frac{1}{n}\sum_{t=1}^n \left\langle \Msgd_{\hat{a}(X_t), t-1}, X_t \right\rangle \right)^2
\end{align*}
where for $i = 0,1$,
\begin{equation}
\label{eq:emp sigma np}
    \hat{\sigma}^2_{i,t} = \frac{ \sum_{s=1}^t I\{a_s = i\}\left(y_s - \left\langle \Msgd_{i,s-1},X_s\right \rangle\right)^2}{\sum_{s=1}^t I\{a_s = i\}}.
\end{equation}
In contrast to \eqref{eq:emp SV}, we substitute $e_t$ with $\hat{e}_t$ when calculating $\widehat{S}_V^2$. Given that $\pi_t$ is unknown in this scenario, the estimation of the noise level $\sigma^2_i$ relies on averaging the sample realizations, setting it apart from \eqref{eq:emp sigma}. The asymptotic normality is formalized in the following theorem. 

\begin{theorem}
\label{thm:consistent np}
Under the conditions of Theorem \ref{thm:value inference np}, we have $\widehat{S}^2_V$ is a consistent estimator of $S^2_V$, i.e., $\widehat{S}^2_V \xrightarrow{p} S^2_V$. Furthermore, we have
 \begin{equation*}
        \frac{ \widehat{V}_n - V^* }{\widehat{S}_V/\sqrt{n}} \xrightarrow{d} \mathcal{N}\left( 0, 1\right).
    \end{equation*}
\end{theorem}
The algorithm for constructing the confidence intervals based on $\widehat{S}_V$ is outlined in Algorithm \ref{alg: optimal value inference np}. The proof of Theorems \ref{thm:value inference np} and \ref{thm:consistent np} are with minor modifications to the proof of Theorems \ref{thm:value inference}, and \ref{thm:consistent} and are therefore relegated.

\begin{algorithm}[t]
    \caption{Online Inference of $V^*$ with Unknown Probability}
        \label{alg: optimal value inference np}
        \begin{algorithmic}[1]
        \State \textbf{Input}: $\Minit_1$, $\Minit_0$, $\Usgd_{i,0}$, $\Vsgd_{i,0}$, $r$.
        
        \State  \textbf{Initialization}: $\Msgd_{i,0} \leftarrow \Minit_i$, $s_i=1$ for $i = 0,1$.

        \State \For{$t\gets 1 $ to $n$}{

            Observe a contextual matrix $X_t$.

            Update $\hat{a}(X_t)$ by equation \eqref{eq:a_hat}.

            Calculate $a_t$ according to the current decision-making policy. 

            \textbf{if} $\sum_{\tau=1}^t I\{a_\tau = 1-a_t\} < p_0t^{1-\beta}$, \textbf{then} \vspace{1mm}
            
            \hspace{0.5cm} Take $1-a_t$ and observe reward $y_t$. ~~$//$~ Force the exploration.

            \textbf{else} 

            \hspace{0.5cm} Take $a_t$ and observe reward $y_t$. \vspace{1mm}

            $\Msgd_{i,t} \leftarrow$ Algorithm \ref{alg:sgd update modify} ($\Usgd_{i,s_i}$, $\Vsgd_{i,s_i}$, $s_i$, $X_t$, $y_t$, $a_t$)

            Calculate the estimator value $\widehat{V}_t$ by equation \eqref{eq:DR est np}.

            Update the variance estimator $\widehat{S}^2_V$ by equation \eqref{eq:emp SV np}.
        }
        \State Obtain the two-sided confidence interval with critical value $z$: $(\widehat{V}_n - z\widehat{S}_V/\sqrt{n}, ~~\widehat{V}_n + z\widehat{S}_V/\sqrt{n} )$.
        \end{algorithmic}
\end{algorithm}

\section{Additional Numerical Studies}\label{app:additional experiments}

In this section, we first present additional numerical studies for parameter inference to complement Section \ref{sec: experiment} in the main text. Then we provide additional experiments to demonstrate that our inference for the optimal policy value is valid in practice. Finally, we present the case where our optimal policy value inference can be done in a specific case of unknown exploration probability. 

\subsection{Comparison with Exploration-only Approach in Parameter Inference}\label{sec: simulation_exploration-only}

\begin{figure}
     \centering
     \begin{subfigure}{0.32\linewidth}
         \includegraphics[width=\linewidth,height = 5cm]{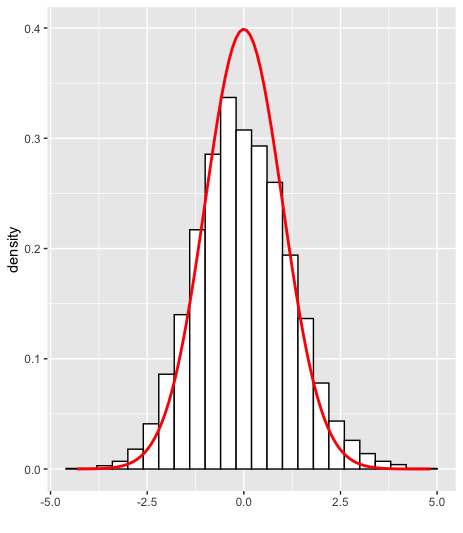}
         \caption{exploration-only: $n = 1000$}
     \end{subfigure}
     \hfill
     \begin{subfigure}{0.32\linewidth}
         \includegraphics[width=\linewidth,height = 5cm]{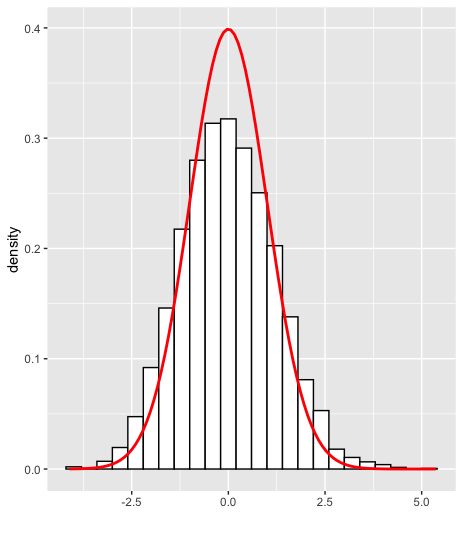}
         \caption{exploration-only: $n = 2000$}
     \end{subfigure}
     \hfill
     \begin{subfigure}{0.32\linewidth}
         \includegraphics[width=\linewidth, height = 5cm]{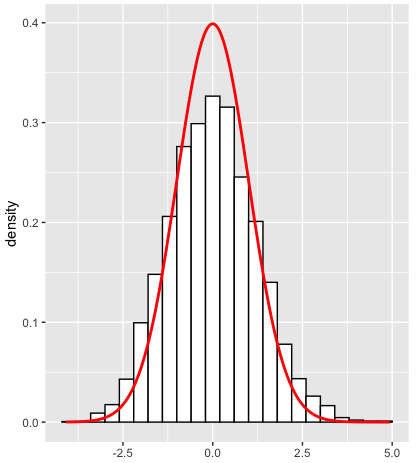}
         \caption{exploration-only: $n = 3000$}
     \end{subfigure}

     \vspace{10pt}
    \begin{subfigure}{0.32\linewidth}
         \includegraphics[width=\linewidth,height = 5cm]{n=1000_i=1_T1.png}
         \caption{our method: $n = 1000$}
     \end{subfigure}
     \hfill
     \begin{subfigure}{0.32\linewidth}
         \includegraphics[width=\linewidth,height = 5cm]{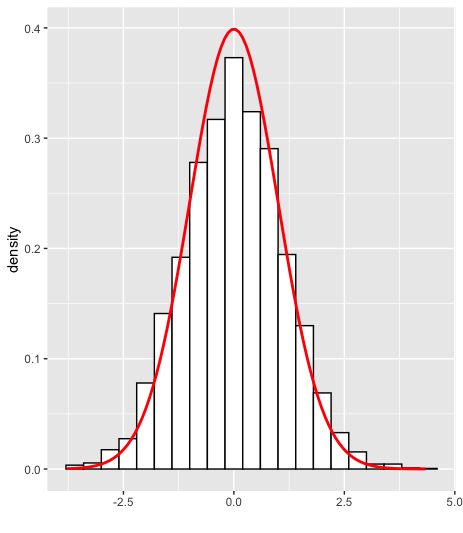}
         \caption{our method: $n = 2000$}
     \end{subfigure}
     \hfill
     \begin{subfigure}{0.32\linewidth}
         \includegraphics[width=\linewidth, height = 5cm]{T1_i=1_n=3000_r=3.png}
         \caption{our method: $n = 3000$}
     \end{subfigure}
     
     \caption{Empirical distribution of $\sqrt{n}(\widehat{m}^{(1)}_T - m^{(1)}_T)/\hat{\sigma}_1\hat{S}_1$ based on $5000$ independent trails for the comparison between exploration-only method and our method in Section \ref{sec: simulation_exploration-only}.}\label{fig:compare exploration only}
\end{figure}

In this section, we compare our inference method with a natural benchmark method where parameter inference relies on exploration-only samples collected under the online decision-making policy. Our goal is to show that the quality of inference results is compromised when relying solely on samples obtained through exploration, contrasting with our approach, which utilizes all samples, both from exploration and exploitation phases. Similar to the settings in Section \ref{sec: experiment}, the experimenter's decision-making policy is defined by an $\varepsilon$-greedy approach with $\varepsilon = 0.1$. We maintain the same simulation parameters as detailed in Section \ref{sec: experiment} and consider sample sizes of $n=1000$, $n=2000$, and $n=3000$ and matrices ($T = T_1$ and $T = T_2$), where $T_1 = e_1e_1^\top$, and $T_2 = e_1e_1^\top + 2e_2e_2^\top - 3e_3e_3^\top$. Table \ref{tab: compare explore CIlen} displays comparisons of estimation mean squared error (MSE) of $\widehat{m}^{(1)}_T$ and the confidence interval length for $m^{(1)}_T$ between two approaches: the exploration only method, which relies solely on exploration samples, and our method. The table indicates that the exploration-only method exhibits a greater estimation error compared to our approach and is also notably less efficient in the inference task. 
In addition, Figure \ref{fig:compare exploration only} provides a histogram illustration of this comparison, which shows that integrating exploitation samples in our method significantly enhances inference performance. 
\begin{table}
\centering
\captionsetup{width=.9\linewidth} 
\caption{Comparisons of estimation MSE for $\widehat{m}^{(1)}_T$ and the confidence interval length between our method 
 and exploration-only method in Section \ref{sec: simulation_exploration-only}.}
\renewcommand{\arraystretch}{1.4}
\setlength{\tabcolsep}{10pt}
\begin{tabular}{c|c|l|l|l|l}
\hline
& & \multicolumn{2}{c|}{Estimation MSE ($\times 10^{-4}$)} & \multicolumn{2}{c}{CI Length} \\ \cline{3-6} 
& & Our Method & Exploration-Only & Our Method & Exploration-Only \\ \hline
\multirow{3}{*}{$T_1$} & $n=1000$ & $0.096$ & $0.191 $ & $0.010$ & $0.014$ \\ \cline{2-6} 
& $n=2000$ & $0.046$ & $0.095$ & $0.008$ & $0.010$ \\ \cline{2-6} 
& $n=3000$ & $0.029$ & $0.062$ & $0.006$ & $0.008$ \\ \hline
\multirow{3}{*}{$T_2$} & $n=1000$ & $1.653$ & $3.517$ & 0.042 & 0.057 \\ \cline{2-6} 
& $n=2000$ & $0.786$ & $1.735$ & $0.031$ & $0.040$ \\ \cline{2-6} 
& $n=3000$ & $0.510$ & $1.149$ & $0.026$ & $0.033$ \\ \hline
\end{tabular}
\label{tab: compare explore CIlen}
\end{table}

\subsection{Optimal Policy Value Inference with Decaying Exploration Probability}\label{sec:sim_optimal} 
\begin{figure}
     \centering
     \begin{subfigure}[b]{0.32\linewidth}
     \subcaptionbox{Empirical histogram. Coverage probability: 0.945.\label{fig:sub1}}{%
         \includegraphics[width=\linewidth, height=5cm]{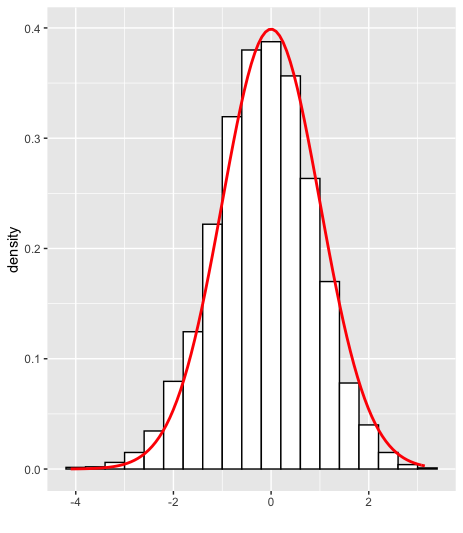}}
        \end{subfigure}
     \hfill
     \begin{subfigure}[b]{0.32\linewidth}
     \subcaptionbox{Estimation error for $\sigma_1$ from $n=100$ to $n=5000$.\label{fig:sub2}}{%
         \includegraphics[width=\linewidth, height=5cm]{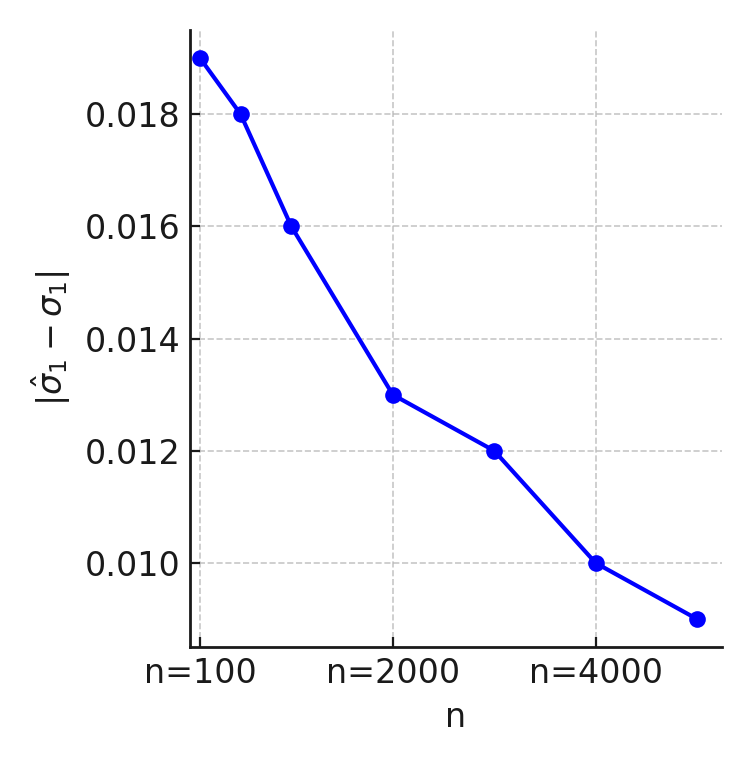}}
        \end{subfigure}
     \hfill
     \begin{subfigure}[b]{0.32\linewidth}
     \subcaptionbox{Estimation error for the variance of $\langle M_{a^*(X)}, X\rangle$ from $n=100$ to $n=5000$.\label{fig:sub3}}{
         \includegraphics[width=\linewidth, height=5cm]{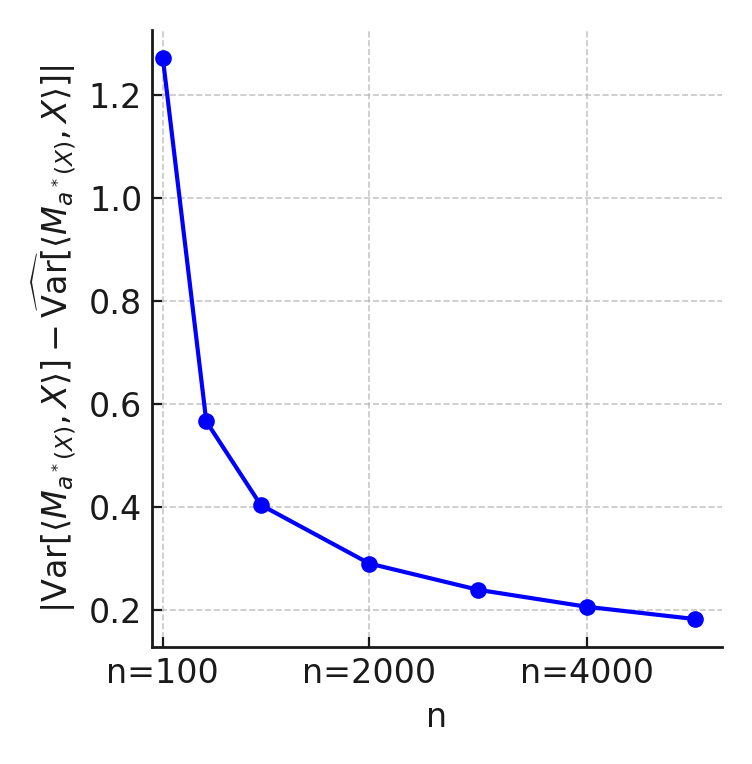}}
        \end{subfigure}
     \caption{Estimation and inference results for the optimal policy value in Section \ref{sec:sim_optimal}.}
     \label{fig:value inference}
\end{figure}

In this section, we assess the performance of optimal value inference. As discussed in Section \ref{sec:optimal policy value}, unlike parameter inference, optimal value inference relaxes the constant lower bound condition on the exploration probability, enabling it to decay over time. To illustrate this, we revisit the $\varepsilon$-greedy decision-making policy explored in Section \ref{sec: experiment}. However, in contrast to the setup where $\varepsilon = 0.1$ remained constant in Section \ref{sec: experiment}, we introduce a decaying exploration probability $\varepsilon_t=0.05t^{-0.1}$ for the new simulations. We set $\|M_1\|=15$ and $\|M_0\|=1$. As depicted in Figure \ref{fig:sub1}, when $n=1000$, the coverage probability has already reached $0.945$. Additionally, Figures \ref{fig:sub2} and \ref{fig:sub3} demonstrate the convergence of the estimation for $\sigma_i$ and $\mathrm{Var}[\langle M_{a^*(X)}, X\rangle]$, respectively, where the convergence behavior of the estimation error is assessed across sample sizes ranging from $n=100$ to $n = 5000$.

\subsection{Optimal Policy Value Inference with Approximate Thompson Sampling}
\label{sec: approx TS}

In our numerical investigation, in addition to utilizing the $\varepsilon$-greedy policy to illustrate the inference results, we expand the inference for optimal policy value by incorporating approximate Thompson sampling as our decision-making policy. While Thompson sampling has demonstrated efficacy in various simple online decision-making contexts \citep{agrawal2013thompson,russo2018tutorial}, its application encounters challenges in deriving the posterior distribution in low-rank matrix scenarios due to the non-convex nature of the parameter space, hindering the feasibility of obtaining a closed-form posterior. Therefore, we employ ensemble sampling, an efficient approximate Thompson Sampling technique, for sequential decision-making \citep{lu2017ensemble,lu2021low, zhou2024stochastic}.

Instead of sampling from the true posterior (which might be computationally infeasible or unknown), ensemble sampling maintains an ensemble of models. Each model in the ensemble represents a possible set of parameters about the true underlying process. When making decisions, the algorithm randomly selects a model from the ensemble and uses its parameters to determine the action. For each model, we update its parameter by deriving the Maximum A Posteriori (MAP) estimate, which serves as the most probable parameter fitting the current observations for each model. One can also view this MAP estimate as a reflection of the exploitation as this suggests actions that are optimal according to the most probable parameter given its experience. On the other hand, the exploration is also considered in this method since each model might have different beliefs about the best action, selecting between them introduces variability and thus exploration. As a consequence, the number of models in the ensemble directly impacts the degree of exploration. With a larger ensemble size, there's a higher chance of having diverse models representing different sets of parameters that characterize the true environment.

We detail our method as follows: Let $K$ represent the total number of models to be combined through ensembling. Each model begins with a Gaussian prior over its parameters. Initially, each row of the parameters $\Usgd^{(m)}_{i,0}$ and $\Vsgd^{(m)}_{i,0}$ are sampled from a Gaussian prior distribution for all $m \in [K]$, i.e.,
\begin{align*}
     [\Usgd_{i,0}^{(m)}]_j \sim \mathcal{N}\left(\boldsymbol{\mu}_{j,u}, \sigma_u^2 I\right), \quad [\Vsgd_{i,0}^{(m)}]_k \sim \mathcal{N}\left(\boldsymbol{\mu}_{k,v}, \sigma_v^2 I\right), ~~ j\in[d_1],~ k\in [d_2],
\end{align*}
where $[\Usgd]_j$ denotes the $j$-th row of matrix $\Usgd$. At each step $t$, we randomly select one model, denoted by $\tilde{m}$, from the $K$ available models. The decision-making is made after observing $X_t$ and the resulting action is based on the parameters of the chosen model. The parameters $(\Usgd^{(m)}_{1},\Vsgd^{(m)}_{1})$ or $(\Usgd^{(m)}_{0},\Vsgd^{(m)}_{0})$ are then updated for all $m \in [K]$ using a closed-form MAP estimate that incorporates all accumulated data for the selected action $i$. In line with the principles of ensemble sampling, the observed reward $y_t$ according to \eqref{eq: model} is perturbed by a random noise $w_t^{(m)} \sim \mathcal{N}(0, \tilde{\sigma}^2)$, to obtain $\tilde{y}_t^{(m)} = y_t + w_t^{(m)}$ for each model. The parameters $(\Usgd^{(m)}_{1},\Vsgd^{(m)}_{1})$ or $(\Usgd^{(m)}_{0},\Vsgd^{(m)}_{0})$ are then updated for all $m \in [K]$ using a closed-form MAP estimate that incorporates all accumulated data for the selected action $i$. In particular, $\Usgd_{i,t}^{(m)}$ and $\Vsgd_{i,t}^{(m)}$ can be obtained by solving 
\begin{align}
\label{eq:MAP obj}
    (\Usgd_t^{(m)}, \Vsgd_t^{(m)})  &=  \argmin_{U,V} \frac{1}{2\sigma^2}\sum_{s = 1}^{t-1} \left(\tilde{y}^{(m)}_s - \langle X_s, UV^\top \rangle\right)^2 \notag \\
    + & \frac{1}{2\sigma_u^2} \sum_{j=1}^{d_1} \left \|[U]_j - \boldsymbol{\mu}_{j,u} \right\|^2 + \frac{1}{2\sigma_v^2} \sum_{j=1}^{d_2} \left\|[V]_j - \boldsymbol{\mu}_{j,v} \right\|^2  + \log f(X_1 \ldots X_{t-1}). 
\end{align}
Note that we use perturbed reward $\tilde{y}^{(m)}_s$ instead of $y_t$ to obtain the MAP estimate to further diversify the point estimates to form the approximated posterior. In practice, we can solve \eqref{eq:MAP obj} using Alternative Least Square (ALS). Our estimation procedure can be seen as an extension of ensemble sampling techniques for contextual bandits \citep{lu2017ensemble, lu2021low} and low-rank bandits \citep{zhou2024stochastic} to the low-rank matrix contextual bandit setting. Based on these estimators, we are ready to present the procedure for conducting optimal policy value inference with ensemble sampling in Algorithm \ref{alg: TS opt value inference}.

\begin{algorithm}[t]
    \caption{Optimal Policy Value Inference via Ensemble Sampling}
    \label{alg: TS opt value inference}
    \begin{algorithmic}[1]
        \State \textbf{Input}: $r$, $\{\boldsymbol{\mu}_{i,u}^{(1)}\}_{j \in [d_1]}$, $\{\boldsymbol{\mu}_{i,v}^{(1)}\}_{i \in [d_1]}$, $\{\boldsymbol{\mu}_{j,u}^{(0)}\}_{j \in [d_2]}$, $\{\boldsymbol{\mu}_{j,v}^{(0)}\}_{j \in [d_2]}$, $\sigma_u^2$, $\sigma_v^2$, $K$, and  $\tilde{\sigma}^2$
        \State \textbf{Initialization}: $[\Usgd_{1,0}^{(m)}]_i \sim \mathcal{N}(\boldsymbol{\mu}_{i,u}^{(1)}, \sigma_u^2 I_{r\times r})$, $[\Vsgd_{1,0}^{(m)}]_j \sim \mathcal{N}(\boldsymbol{\mu}_{j,v}^{(1)}, \sigma_v^2 I_{r\times r})$, 
        
        $[\Usgd_{0,0}^{(m)}]_i \sim \mathcal{N}(\boldsymbol{\mu}_{i,u}^{(0)}, \sigma_u^2 I_{r\times r})$, $[\Vsgd_{0,0}^{(m)}]_j \sim \mathcal{N}(\boldsymbol{\mu}_{j,v}^{(0)}, \sigma_v^2 I_{r\times r})$ for all $m \in [K]$, $i \in [d_1]$, $j \in [d_2]$.
        
        \State \For{$t = 1 $ to $n$}{
             Sample $\tilde{m} \sim$ Unif $\{1, 2, \ldots,K\}$.
             
             Observe context $X_t$.
             
             Calculate $a_t = I\left\{\left\langle \Usgd_{1,t-1}^{(\tilde{m})} \Vsgd_{1,t-1}^{(\tilde{m})\top} - \Usgd_{0,t-1}^{(\tilde{m})} \Vsgd_{0,t-1}^{(\tilde{m})\top}, X_t \right\rangle > 0\right\}$.
             
             Observe $y_t$ according to \eqref{eq: model}.
            
             Calculate $\eta_{1,t} = \frac{1}{K} \sum_{m=1}^{K} \left\langle \Usgd_{1,t-1}^{(m)} \Vsgd_{1,t-1}^{(m)\top},X_t \right\rangle$ ; $\eta_{0,t} = \frac{1}{K} \sum_{m=1}^{K} \left\langle \Usgd_{0,t-1}^{(m)} \Vsgd_{0,t-1}^{(m)\top},X_t \right\rangle$.
            
             Calculate $\hat{a}(X_t) = I\left\{\eta_{1,t} > \eta_{0,t} \right\}$.
            
             Calculate $\hat{e}_t = \frac{1}{K} \sum_{m=1}^{K} I\left\{\left \langle \Usgd_{\hat{a}(X_t),t-1}^{(m)} \Vsgd_{\hat{a}(X_t),t-1}^{(m)\top},X_t\right\rangle < \left \langle \Usgd_{1-\hat{a}(X_t),t-1}^{(m)} \Vsgd_{1-\hat{a}(X_t),t-1}^{(m)\top},X_t\right\rangle\right\}$.
            
             Get the values for $\hat{V}_t$, and $\hat{S}_V$ according to \eqref{eq:DR est np} and \eqref{eq:emp SV np}, respectively.
            
             \For{$m = 1 $ to $K$}{
                 Sample $w_t^{(m)} \sim \mathcal{N}(0, \tilde{\sigma}^2)$.
                 
                 Calculate $\tilde{y}_t^{(m)} = y_t + w_t^{(m)}$.
                 
                 Update $(\Usgd^{(m)}_{a_t,t}, \Vsgd^{(m)}_{a_t,t})$ by solving \eqref{eq:MAP obj}.

                 Set the parameters for the un-selected action: $(\Usgd^{(m)}_{1-a_t,t}, \Vsgd^{(m)}_{1-a_t,t}) \leftarrow (\Usgd^{(m)}_{1-a_t,t-1}, \Vsgd^{(m)}_{1-a_t,t-1}).$
            }
        }
        \State Obtain the two-sided confidence interval with critical value $z$: $(\widehat{V}_n - z\widehat{S}_V/\sqrt{n}, \widehat{V}_n + z\widehat{S}_V/\sqrt{n} )$.
    \end{algorithmic}
\end{algorithm}

In Algorithm \ref{alg: TS opt value inference}, the input $\boldsymbol{\mu}_{i,u}^{(1)}$ denote the mean of the prior distribution for the $i$-th row of $\Usgd_{1}$. Consequently, during initialization, each row of $\Usgd$ and $\Vsgd$ across all models is sampled from a normal distribution. The input $\sigma_u$ and $\sigma_v$ denote the perturbation to the prior sample and result in the covariance matrices of the prior distribution are defined by $\sigma_u^2 I_{r\times r}$ and $\sigma_v^2 I_{r\times r}$. In addition, the input $K$ represents the number of models, and $\tilde{\sigma}$ specifies the perturbation noise level applied on $y_t$ for each model. Notably, the action at time $t$ is determined by the parameter estimation of a model selected at random. In this decision-making framework, every model refines its parameter estimation through the MAP estimation, leveraging the most likely parameters given the data observed by each model, which can be viewed as exploiting the current data collected by each model. Subsequently, the updated parameters across all models constitute an empirical distribution, from which the estimated optimal action $\hat{a}(X_t)$ is determined based on the empirical mean of the estimated rewards across models. The exploration probability $\hat{e}_t$ is then calculated as the fraction of models for which the suboptimal action, $1-\hat{a}(X_t)$, yields a higher estimated reward. Following the determination of $\hat{a}(X_t)$ and $\hat{e}_t$, the algorithm proceeds to update each model's parameters using the perturbed observed reward corresponding to each model. We can see that the perturbation noise level impacts the degree of exploration: higher perturbation noise leads to more diversified models, resulting in the algorithm incorporating greater exploration. It is worth to note that when $K=1$, and $\tilde{\sigma} = 0$, the decision-making policy reflects pure exploitation.

\begin{figure}
     \centering
     \begin{subfigure}[b]{0.4\linewidth}
     \subcaptionbox{$n=300$\label{fig:ts1}}{
         \includegraphics[width=\linewidth, height=5cm]{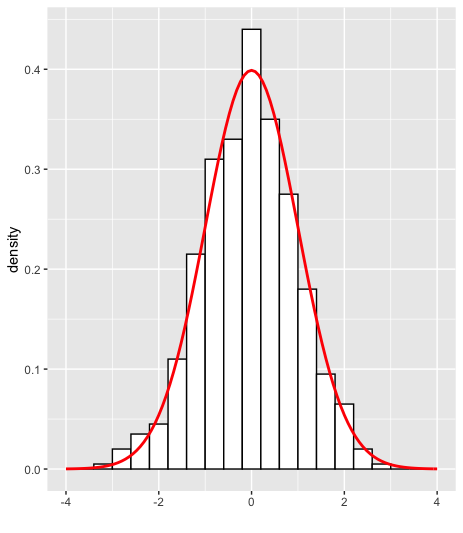}}
        \end{subfigure}
     \hfill
     \begin{subfigure}[b]{0.4\linewidth}
     \subcaptionbox{$n=700$\label{fig:ts2}}{
         \includegraphics[width=\linewidth, height=5cm]{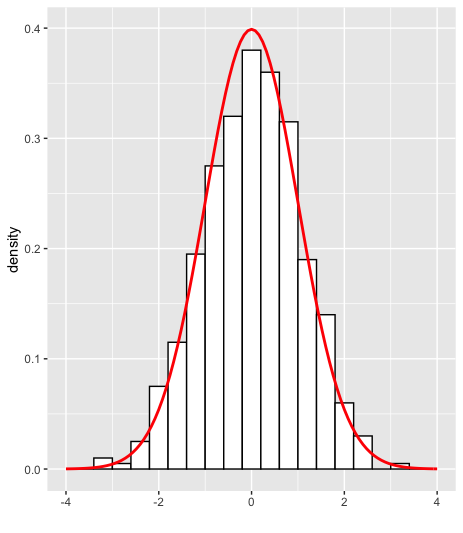}}
        \end{subfigure}
     \hfill
     \caption{Histogram of $\sqrt{n}(\widehat{V}_{n} - V^*)/\widehat{S}_V$ with varying sample sizes for the approximated Thompson Sampling in Section \ref{sec: approx TS}.}\label{fig:TS value inference}
\end{figure}

We next use a simulation to demonstrate the inference procedure provided in Algorithm \ref{alg: TS opt value inference}. The setting is the same as Section \ref{sec: experiment}, except that $d=20$ and $r=1$. In Algorithm \ref{alg: TS opt value inference}, we set $\sigma_u = \sigma_v = \sigma_1 = \sigma_0 = 0.1$, and the perturbation noise level $\tilde{\sigma} = 0.05$. Finally, we choose the number of models to be $K=10$, and the results are reported based on $500$ independent trails. Figure \ref{fig:TS value inference} illustrates the histogram of $\sqrt{n}(\widehat{V}_{n} - V^*)/\widehat{S}_V$ for both $n=300$ and $n=700$. Even when the sample size $n$ is as small as $300$, we can see our procedure still shows a reasonably good normal approximation. Moreover, Figure \ref{fig:TS value inference} shows that when we increase the sample size from $300$ to $700$, the proposed method achieves a better inference result.

\section{Discussion on Different Distributions for \texorpdfstring{$X$}{X}.}\label{sec:completion setting}

In this section, we discuss generalization of the Gaussian assumption in Assumption \ref{assum: noise} (ii) to a scenario where $X$ is sampled from a different distribution. Specifically, we consider the case where the matrix $X$ is uniformly sampled from the set $\{e_{j_1}e_{j_2}^\top:$ $j_1 \in [d_1], j_2 \in [d_2]\}$, where $e_{j_1} \in \mathbb{R}^{d_1}$ and $e_{j_2} \in \mathbb{R}^{d_2}$ are the canonical basis vectors. This corresponds to the low-rank matrix completion setting with uniformly missing entries. Under this distribution of $X$, at each time $t$, the reward is a noisy observation of the entries of $M_i$. The goal is to recover the matrix $M_i$ and conduct valid statistical inference on its entries. This problem is particularly relevant in the context of online recommendation systems \citep{koren2009bellkor,jin2016provable,jain2022online}, where the matrix represents user-item ratings, with each entry indicating how a user rates a product.

Even when $X$ has only one active entry at each time step, we can still apply SGD for sequential estimatio, with a slight modification to the updating rule presented in \eqref{eq: practice update}. Recall that for $X_t = e_{j_1}e_{j_2}^\top$, where $j_1$ and $j_2$ are independently sampled uniformly from $\{1,2,\dots, d\}$. The probability of $X_t(j_1,j_2)=1$ is $(d_1d_2)^{-1}$, and thus our updating rule is
\begin{equation*}
\left(\begin{array}{l}
\Usgd_{i,t}\\
\Vsgd_{i,t}
\end{array}\right) = \left(\begin{array}{l}
\Usgd_{i,t-1}\\
\Vsgd_{i,t-1}
\end{array}\right) - \eta_t d_1d_2 g(\Usgd_{i,t-1}, \Vsgd_{i,t-1};{X_t, y_t,a_t, \pi_t}),
\end{equation*}
ensuring that the new gradient remains an unbiased estimator of $\nabla F(\Usgd_{i,t-1}, \Vsgd_{i,t-1})$. Additionally, our online debiasing procedure requires a similar adjustment. Given the distribution of $X_t$, we have:
\begin{equation*}
\widetilde{M}_{1,t} = \Msgd_{1,t-1} + d_1d_2 \frac{I\{a_t = 1\}}{\pi_t} (y_t - \langle \Msgd_{1,t-1}, X_t \rangle )X_t.
\end{equation*}
Following Section \ref{sec: online de-bias}, $\Munbs_{i,n}$ is calculated as the running average of all $\widetilde{M}_{i,t}$ after $n$ iterations. Specifically, for $i=1$, we have:
\begin{equation*}
\Munbs_{1,n} = \frac{d_1d_2}{n}\sum_{t=1}^n\Msgd_{1,t-1} + \frac{d_1d_2}{n} \sum_{t=1}^n \frac{I\{a_t = 1\}}{\pi_t} (y_t - \langle \Msgd_{1,t-1}, X_t \rangle )X_t.
\end{equation*}
It is expected that $\Munbs_{1,n}$ remains an unbiased estimator of $M_1$ given the distribution of $X_t$. The inference procedure then follows as described in Section \ref{sec: Inference}. We leave a comprehensive investigation of this online matrix completion setting as future work. 

\section{Proof of Main Theorems}\label{appA}

In this proof section, we set $i=1$. Since the analysis is identical for $i=0$, we drop the index $i$ for notational simplicity. For the theoretical proofs in both Sections \ref{appA} and \ref{appB}, we define a convex function $\psi_{p} : \mathbb{R}^{+} \rightarrow \mathbb{R}^{+}$ for $p \in (0,\infty)$. When $p\in [1, \infty)$, we define the function $\psi_p(u) = \exp(u^p) - 1$ for $u \ge 0$. When $p \in (0,1)$, we define $\psi_p(u) = \exp(u^p) - 1$ for $u>u_0$, and $\psi_p(u)$ is linear for $ 0\le u \le u_0$ to preserve the convexity of $\psi_p$ \citep[Theorem 6.21]{ledoux1991probability}. In addition, the corresponding Orlicz norm is defined as
\begin{equation*}
    \| Y\|_{\psi_{p}} = \inf \{\upsilon \in (0,\infty): \mathbb{E}[\psi_p(\vert Y \vert/\upsilon)] \le 1 \}.
\end{equation*}
For example, $ \| \cdot \|_{\psi_{1}}$ and $ \| \cdot \|_{\psi_{2}}$  denote the sub-exponential and sub-Gaussian norms. 

\subsection{A generalized version of Theorem \ref{thm: sgd consistent}}
We first provide a generalized version of Theorem \ref{thm: sgd consistent} under relaxed initial condition, as stated in the following. 
\begin{theorem}
    \label{thm: sgd consistent_gen}
    Define the learning rate $\eta_{t} = C_\eta \cdot (\max\{t, t^\star \})^{-\alpha}$ where 
    \begin{equation}\label{eqeqSNR}
    t^\star\ge c^\star
    \left\{(\gamma^2 dr\sigma_i^2 (\log d)^2/\lambda_r^2)^{\frac{1}{\alpha-\beta}},\left(\frac{1}{2-\alpha+\beta}\ln\Bigl(\dfrac{\gamma\,d r\,(\log d)^2}{1+\lambda_r^2/\sigma_i^2}\Bigr)\right)^{\frac{1}{1-\alpha}}\right\}
    \end{equation}
    with $\alpha\in(\beta,1)$ and some constants $C,c^\star>0$.  Assume initialization $ \|\Minit_i - M_i \|_{\mathrm{F}} \le C_0 \lambda_r~$ for some constant $C_0\in(0,1/20)$. Under Assumptions \ref{assum: noise},  \ref{assum: decay rate}, with probability at least $1 - \frac{4n}{d^{\gamma}}$, 
    \begin{equation*}
        \left\| \Msgd_{i,t} - M_i \right \|^2_{\mathrm{F}} \le  \|\Minit_i - M_i \|_{\mathrm{F}}^2\prod_{\tau=1}^t (1-\frac{\eta_\tau\lambda_r}{2})+  C_1 \gamma^2 dr\sigma_i^2 (\log d)^2 t^{\beta}\eta_t,
    \end{equation*}
    for some positive constant $C_1$.
    \end{theorem}
It is strightforward to verify that Theorem \ref{thm: sgd consistent} is a direct corollary of Theorem \ref{thm: sgd consistent_gen}.

\subsection{Proof of Theorem \ref{thm: sgd consistent_gen}}
\label{sec: proof of sgd consistent}
Based on the updating rule presented in the Algorithm \ref{alg:sgd update practice} we note that $ \tilde{\Usgd}_t \tilde{\Vsgd}_t^\top = \Usgd_t \Vsgd_t^\top$, and thus we have 
\begin{equation*}
    \Usgd_{t}\Vsgd_{t}^\top = \Usgd_{t-1}\Vsgd_{t-1}^\top - \Delta_{\Usgd \Vsgd,t},
\end{equation*}
where
\begin{align*}
	\Delta_{\Usgd \Vsgd,t} = & \hspace{2mm}\frac{I\{a_t=1 \}}{\pi_t}\eta_t(\langle \Usgd_{t-1} \Vsgd_{t-1}^\top-M,X_t\rangle - \xi_t)\left(X_t\tilde{\Vsgd}_{t-1}\tilde{\Vsgd}_{t-1}^\top +\tilde{\Usgd}_{t-1}\tilde{\Usgd}_{t-1}^\top X_t\right)\\
	& - \frac{I\{a_t=1 \}}{\pi_t^2}\eta_t^2(\langle \Usgd_{t-1}\Vsgd_{t-1}^\top-M,X_t\rangle - \xi_t)^2 X_t \Vsgd_{t-1}\Usgd_{t-1}^\top X_t,
\end{align*}
and thus we can write
\begin{equation}
\label{eq: expand}
\|\Usgd_t\Vsgd_t^\top - M\|^2_{\mathrm{F}} = \|\Usgd_{t-1}\Vsgd^\top_{t-1} - M\|^2_{\mathrm{F}} +R_t,
\end{equation}
where
\begin{equation}
\label{eq: RtHt}
R_t=-2\langle \Delta_{\Usgd \Vsgd,t}, \Usgd_{t-1}\Vsgd_{t-1}^\top - M \rangle+\| \Delta_{\Usgd \Vsgd,t}\|_\mathrm{F}^2.
\end{equation}
We define the event $E_t$ as 
\begin{equation}
\label{eq: event Et}
    E_t = \left \{\forall \tau \le t:  \|\Usgd_{\tau}\Vsgd_{\tau}^\top-M \|^2_{\mathrm{F}} \le \|\Usgd_{0}\Vsgd_{0}^\top-M \|^2_{\mathrm{F}}\prod_{\tau=1}^t (1-\frac{\eta_\tau\lambda_r}{2})+ C_1 \gamma^2 dr\sigma^2 (\log d)^2 \tau^{\beta} \eta_\tau  \right \},
\end{equation}
for some positive constant $C_1$. By definition $\mathbb{P}(E_0)=1$. Meanwhile, define a region
\begin{equation*}
    D = \left \{(\Usgd,\Vsgd) \Big| \|\Usgd\Vsgd^\top - M \|_{\mathrm{F}}^2 \le \left( \frac{\lambda_r}{10}\right)^2\right\}.
\end{equation*}
It is easy to see that under $E_t$, \eqref{eqeqSNR} and the initial condition, we have $(\Usgd_\tau,\Vsgd_\tau)\in D$ for all $\tau\leq t$. We next restate Lemmas C.3 and C.4 in \cite{jin2016provable} below.
\begin{lemma}
\label{lm: region D}
For $(\Usgd,\Vsgd) \in D$, and for $\Usgd = W_{\Usgd}D^{\frac{1}{2}}$, $\Vsgd = W_{\Vsgd}D^{\frac{1}{2}}$, where $\Usgd\Vsgd^\top=W_\Usgd D W_\Vsgd^\top$, then we have
\begin{equation*}
    \| (\Usgd\Vsgd^\top -M)\Vsgd\|_{\mathrm{F}}^2 + \| (\Usgd\Vsgd^\top -M)^\top\Usgd\|_{\mathrm{F}}^2 \ge \frac{\lambda_r}{2} \| \Usgd\Vsgd^\top -M\|_{\mathrm{F}}^2,
\end{equation*}
and $\|\Vsgd\| = \|\Usgd\| \le \sqrt{2\|M\|}$, $\|\Usgd\Vsgd^\top\| = \|\Vsgd\Usgd^\top\|$, $\|\Vsgd\Vsgd^\top\| \le 2\|M\|$, $\|\Usgd\Usgd^\top\| \le 2\|M\|$.
\end{lemma}

Let $\tilde\Delta_t$ denote $\Usgd_{t}\Vsgd^\top_{t} - M$, and $\delta_t=\|\tilde\Delta_t\|_{\mathrm{F}}$. For any $t$, under $E_{t-1}$, we have
\begin{align*}
\delta_t^2=&\delta_{t-1}^2 -2\langle \mathbb{E}[\Delta_{\Usgd \Vsgd,t}|\mathcal{F}_{t-1}], \tilde\Delta_{t-1} \rangle +\mathbb{E}\left[ \| \Delta_{\Usgd \Vsgd,t}\|_\mathrm{F}^2 |\mathcal{F}_{t-1}\right]\\
&-2\langle \Delta_{\Usgd \Vsgd,t}-\mathbb{E}[\Delta_{\Usgd \Vsgd,t}|\mathcal{F}_{t-1}], \tilde\Delta_{t-1}\rangle +\left( \| \Delta_{\Usgd \Vsgd,t}\|_\mathrm{F}^2 -\mathbb{E}\left[ \| \Delta_{\Usgd \Vsgd,t}\|_\mathrm{F}^2 |\mathcal{F}_{t-1}\right]\right).
\end{align*}

We first note that under $E_{t-1}$,
\begin{align*}
\label{ieq: cross term}
    & 2\langle \mathbb{E}[\Delta_{\Usgd \Vsgd,t}|\mathcal{F}_{t-1}], \tilde\Delta_{t-1}\rangle \\
    =& 2\eta_t \Big \langle \tilde\Delta_{t-1}\tilde{\Vsgd}_{t-1}\tilde{\Vsgd}_{t-1}^\top+ \tilde{\Usgd}_{t-1}\tilde{\Usgd}_{t-1}^\top\tilde\Delta_{t-1},\tilde\Delta_{t-1}\Big \rangle  \\
&-2\eta_t^2 \mathbb{E} \Big[\frac{I\{a_t = 1\}}{\pi_t^2}\Big ( \langle \tilde\Delta_{t-1}, X_t \rangle^2 +\xi_t^2\Big )\left \langle X_t\Vsgd_{t-1}\Usgd_{t-1}^\top X_t, \tilde\Delta_{t-1}\right \rangle \Big | \mathcal{F}_{t-1}\Big] \\ \ge & 2\eta_t \|\tilde\Delta_{t-1}\tilde{\Vsgd}_{t-1} \|_{\mathrm{F}}^2 + 2\eta_t\|\tilde\Delta_{t-1}^\top\tilde{\Usgd}_{t-1} \|_{\mathrm{F}}^2- \frac{6\eta_t^2}{\pi_t}\sqrt{r} \|\Vsgd_{t-1}\Usgd_{t-1}^\top \| \delta_{t-1} \left(\delta_{t-1}^2+ \sigma^2\right ).
\end{align*}
By Lemma \ref{lm: region D}, we have 
\begin{equation}
\label{ieq: cross term}
    2\eta_t \|\tilde\Delta_{t-1}\tilde{\Vsgd}_{t-1} \|_{\mathrm{F}}^2 + 2\eta_t\|\tilde\Delta_{t-1}^\top\tilde{\Usgd}_{t-1} \|_{\mathrm{F}}^2 
\ge {\eta_t}\lambda_r \|\tilde\Delta_{t-1}\|^2_{\mathrm{F}}.
\end{equation}
Meanwhile, we have 
\begin{align}
\label{eq:high order Delta}
   & \mathbb{E}\left[ \| \Delta_{\Usgd \Vsgd,t}\|_{\mathrm{F}}^2 |\mathcal{F}_{t-1}\right] 
   \le  C_0 \frac{\eta_t^2 }{\pi_t}dr \left (\delta_{t-1}^2 + \sigma^2 \right) + C_0\frac{\eta_t^4 }{\pi_t^3} d^2r^2 \left (\delta_{t-1}^4 + \sigma^4 \right).
\end{align}
for an absolute constant $C_0$. 
Therefore, 
\begin{align*}
\|\tilde\Delta_t\|^2_{\mathrm{F}}\leq&\|\tilde\Delta_{t-1}\|^2_{\mathrm{F}}-\eta_t\lambda_r\|\tilde\Delta_{t-1}\|^2_{\mathrm{F}}\\
&+C_0\left(\frac{\eta_t^2}{\pi_t}\sqrt{r}  \delta_{t-1} + \frac{\eta_t^2 }{\pi_t}dr + \frac{\eta_t^4 }{\pi_t^3} d^2r^2 \delta_{t-1}^2\right)\delta_{t-1}^2\\
&+C_0\left(\frac{\eta_t^2}{\pi_t}\sqrt{r} \delta_{t-1} \sigma^2+\frac{\eta_t^2 }{\pi_t}dr \sigma^2  + \frac{\eta_t^4 }{\pi_t^3} d^2r^2 \sigma^4\right)\\
&-2\langle \Delta_{\Usgd \Vsgd,t}-\mathbb{E}[\Delta_{\Usgd \Vsgd,t}|\mathcal{F}_{t-1}], \tilde\Delta_{t-1}\rangle +\left( \| \Delta_{\Usgd \Vsgd,t}\|_\mathrm{F}^2 -\mathbb{E}\left[ \| \Delta_{\Usgd \Vsgd,t}\|_\mathrm{F}^2 |\mathcal{F}_{t-1}\right]\right).
\end{align*}
By the definition of $\eta_t$, we can set $C_\eta$ small enough, such that for any $t$,
\begin{align*}
& C_0\left(\frac{t^\beta\eta_t}{\pmin}\sqrt{r}  \delta_{t-1} + \frac{t^\beta\eta_t }{\pmin}dr + \frac{t^{3\beta}\eta_t^3 }{\pmin^3} d^2r^2 \delta_{t-1}^2\right)\le\lambda_r.
\end{align*}
Then we can write
\[
\|\tilde\Delta_t\|^2_{\mathrm{F}}\leq (1-\frac{\eta_t\lambda_r}{2})\|\tilde\Delta_{t-1}\|^2_{\mathrm{F}}+Q_t+\overline R_t,
\]
where 
\begin{align*}
Q_t&=C_0\big(t^\beta \eta_t^2\sqrt{r}  \delta_{t-1} \sigma^2+t^\beta \eta_t^2dr \sigma^2 + t^{3\beta} \eta_t^4 d^2r^2 \sigma^4\big); \\
\overline R_t&=-2\langle \Delta_{\Usgd \Vsgd,t}-\mathbb{E}[\Delta_{\Usgd \Vsgd,t}|\mathcal{F}_{t-1}], \tilde\Delta_{t-1}\rangle+\| \Delta_{\Usgd \Vsgd,t}\|_\mathrm{F}^2 -\mathbb{E}\left[ \| \Delta_{\Usgd \Vsgd,t}\|_\mathrm{F}^2 |\mathcal{F}_{t-1}\right].
\end{align*}
Therefore, by telescoping we have
\[
\delta_t^2\leq\delta_0^2\prod_{\tau=1}^t \left(1-\frac{\eta_\tau\lambda_r}{2}\right)+\sum_{\tau=1}^{t} (Q_\tau+ \overline R_\tau)\prod_{s=\tau+1}^t \left(1-\frac{\eta_s\lambda_r}{2}\right).
\]
We next prove that under $E_{t-1}$, the following with probability at least $1-4d^{-\gamma}$,
\begin{align*}
    &\left \vert \sum_{\tau=1}^{t} (Q_\tau + \overline R_\tau) \prod_{s=\tau+1}^t \left (1-\frac{\eta_s\lambda_r}{2}\right) \right \vert \le  C_1 \gamma^2 dr\sigma^2 (\log d)^2 t^{\beta}\eta_t.
\end{align*}
By Assumption \ref{assum: noise} and Lemma \ref{lm: region D}, conditional on $\mathcal{F}_{t-1}$, we have 
\begin{align*}
&\|\langle \tilde\Delta_{t-1},X_t\rangle|\mathcal{F}_{t-1}\|_{\psi_2}\leq \delta_{t-1},\qquad \|\xi_t|\mathcal{F}_{t-1}\|_{\psi_2}\leq \sigma,\\
&\|\langle\tilde{\Usgd}_{t-1}\tilde{\Usgd}_{t-1}^\top X_t,\tilde\Delta_{t-1}\rangle|\mathcal{F}_{t-1}\|_{\psi_2}\leq 2\delta_{t-1},\\
&
\|\langle X_t \Vsgd_{t-1}\Usgd_{t-1}^\top X_t,\tilde\Delta_{t-1}\rangle|\mathcal{F}_{t-1}\|_{\psi_1}\leq 4r\delta_{t-1},\\
&\big\|\|\tilde{\Usgd}_{t-1}\tilde{\Usgd}_{t-1}^\top X_t\|_{\mathrm{F}}^2-\mathbb{E}[\|\tilde{\Usgd}_{t-1}\tilde{\Usgd}_{t-1}^\top X_t\|_{\mathrm{F}}^2|\mathcal{F}_{t-1}]\big|\mathcal{F}_{t-1}\big\|_{\psi_1}\leq 4dr,\\
&\big\|\|X_t \Vsgd_{t-1}\Usgd_{t-1}^\top X_t\|_{\mathrm{F}}^2-\mathbb{E}[\|X_t \Vsgd_{t-1}\Usgd_{t-1}^\top X_t\|_{\mathrm{F}}^2|\mathcal{F}_{t-1}]\big|\mathcal{F}_{t-1}\big\|_{\psi_{\frac12}}\leq 4d^2r.
\end{align*}
Define 
\begin{align*}
    R_t^{(1)}=&-\frac{2 \cdot I\{a_t=1 \}}{\pi_t}\eta_t(\langle \tilde\Delta_{t-1},X_t\rangle - \xi_t)\langle X_t\tilde{\Vsgd}_{t-1}\tilde{\Vsgd}_{t-1}^\top,\tilde\Delta_{t-1}\rangle\\
    &-\frac{2 \cdot I\{a_t=1 \}}{\pi_t} \eta_t(\langle \tilde\Delta_{t-1},X_t\rangle - \xi_t)\langle\tilde{\Usgd}_{t-1}\tilde{\Usgd}_{t-1}^\top X_t,\tilde\Delta_{t-1}\rangle; \\
    R_t^{(2)}=& \frac{2 \cdot I\{a_t=1 \}}{\pi_t^2}\eta_t^2(\langle \tilde\Delta_{t-1},X_t\rangle - \xi_t)^2 \langle X_t \Vsgd_{t-1}\Usgd_{t-1}^\top X_t,\tilde\Delta_{t-1}\rangle\\
    &+\frac{ I\{a_t=1 \}}{\pi_t^2}{\eta_t^2}(\langle \tilde\Delta_{t-1},X_t\rangle - \xi_t)^2\| X_t\tilde{\Vsgd}_{t-1}\tilde{\Vsgd}_{t-1}^\top\|_{\mathrm{F}}^2\\
    &+\frac{ I\{a_t=1 \}}{\pi_t^2} \eta_t^2(\langle \tilde\Delta_{t-1},X_t\rangle - \xi_t)^2\|\tilde{\Usgd}_{t-1}\tilde{\Usgd}_{t-1}^\top X_t\|_{\mathrm{F}}^2; \\
    R_t^{(3)}=& \frac{ I\{a_t=1 \}}{\pi_t^4}\eta_t^4(\langle \tilde\Delta_{t-1},X_t\rangle - \xi_t)^4 \|X_t \Vsgd_{t-1}\Usgd_{t-1}^\top X_t\|_{\mathrm{F}}^2.
\end{align*}
Define $\overline R_t^{(k)}=R_t^{(k)}-\mathbb{E}[R_t^{(k)}|\mathcal{F}_{t-1}]$ for $k=1,2,3$. Note 
\begin{align*}
&\|\overline R_t^{(1)}|\mathcal{F}_{t-1}\|_{\psi_{1}}\leq C \Psi_t^{(1)},\quad \Psi_t^{(1)}=t^{\beta}\eta_t \delta_{t-1}(\delta_{t-1} + \sigma );\\
&\|\overline R_t^{(2)}|\mathcal{F}_{t-1}\|_{\psi_{\frac12}}\leq C \Psi_t^{(2)},\quad \Psi_t^{(2)}=t^{2\beta}\eta_t^2  (\delta_{t-1}+d)r(\delta_{t-1}^2 + \sigma^2);\\
&\|\overline R_t^{(3)}|\mathcal{F}_{t-1}\|_{\psi_{\frac14}}\leq C\Psi_t^{(3)},\quad \Psi_t^{(3)}= t^{4\beta}\eta_t^4  d^2r(\delta_{t-1}+ \sigma)^4.
\end{align*}

According to Assumption \ref{assum: decay rate}, $\mathcal{P}(a_t=1|\mathcal{F}_{t-1})=\pi_t \geq t^{-\beta}\pmin$, where $\pmin$ is a constant, 
\begin{align*}
&\mathbf{Var}(\overline R_t^{(k)}|\mathcal{F}_{t-1})\leq C\mathcal{V}_t^{(k)},
\end{align*}
where
\begin{align*}
\mathcal{V}_t^{(1)}&=t^{\beta}\eta_t^2\delta_{t-1}^2(\delta_{t-1} + \sigma )^2\\
\mathcal{V}_t^{(2)}&=t^{3\beta}\eta_t^4  (\delta_{t-1}+d)^2r^2(\delta_{t-1} + \sigma )^4;\\
\mathcal{V}_t^{(3)}&=t^{7\beta}\eta_t^8  d^4r^2(\delta_{t-1} + \sigma )^8
\end{align*}
Define $
\zeta_t=\prod_{s=\tau+1}^t \left (1-\frac{\eta_s\lambda_r}{2}\right)$. 
By a martingale concentration inequality, with probability $1-4d^{-\gamma}$,
\begin{equation}\label{eq:Rbound}
\left|\sum_{\tau =1}^t \zeta_\tau \overline  R_\tau ^{(k)}\right|\leq C\sqrt{\tilde{\mathcal{V}}_t\gamma\log d}+C(\gamma\log d)^{2^{k-1}}\widetilde\Psi_t,\quad k=1,2,3,
\end{equation}
where
\[
\widetilde Q_t=\sum_{\tau=1}^t\zeta_\tau Q_\tau,\quad \widetilde{\mathcal{V}}_t^{(k)}=\sum_{\tau=1}^t\zeta_\tau^2\mathcal{V}_{\tau}^{(k)},\quad \widetilde\Psi_t^{(k)}=\max_{1\leq \tau\leq t}(\zeta_\tau\Psi_\tau^{(k)}).
\]
We then introduce two lemmas for the computation of $\widetilde Q_t$, $\widetilde{\mathcal{V}}_t^{(k)}$, $\widetilde \Psi_t^{(k)}$. 

\begin{lemma}
\label{lm: sum of prod}
For $0<\rho<\alpha h$ and $h\ge1$, under the assumptions in Theorem \ref{thm: sgd consistent},
\begin{equation*}
    \sum_{\tau=1}^{t}\tau^{\rho}\eta_{\tau}^h  \prod_{s=\tau+1}^t \left (1-\frac{\eta_s\lambda_r}{2}\right) \le \tilde{C}_1t^{\rho}\eta_t^{h-1}.
\end{equation*}
\end{lemma}

\begin{lemma}
\label{lm: max term}
For $0<\rho<\alpha h$ and $h\ge1$, under the assumptions in Theorem \ref{thm: sgd consistent},
\begin{equation*}
    \max_{1\leq \tau \leq t}\left (\tau^\rho \eta_{\tau}^h \prod_{s=\tau+1}^t \left (1-\frac{\eta_s\lambda_r}{2}\right)\right) \le \tilde{C}_2t^\rho\eta_{t}^h.
\end{equation*}
\end{lemma}

When $t\leq t^\star$, we have $\eta_t=\eta_{t^\star}$, and $\delta_{t-1}\leq C(\lambda_r+\sigma)$. Therefore 
By Lemma \ref{lm: sum of prod}, we have 
\[
\widetilde Q_t\leq C\left(t^\beta \eta_t\sqrt{r} (\lambda_r+\sigma) \sigma^2+t^\beta \eta_t dr \sigma^2 + t^{3\beta} \eta_t^3 d^2r^2 \sigma^4\right).
\]
and
\[
\widetilde{\mathcal{V}}_t^{(1)}+\widetilde{\mathcal{V}}_t^{(2)}+\widetilde{\mathcal{V}}_t^{(3)}\leq C\left(t^{\beta}\eta_t(\lambda_r + \sigma )^4+t^{3\beta}\eta_t^3  r^2(\lambda_r + \sigma +d)^2(\lambda_r + \sigma )^4+t^{7\beta}\eta_t^7  d^4r^2(\lambda_r+ \sigma )^8\right).
\]
By Lemma \ref{lm: max term}, we have
\[
\widetilde\Psi_t^{(1)}+\widetilde\Psi_t^{(2)}+\widetilde\Psi_t^{(3)}\leq C\left(t^{\beta}\eta_t(\lambda_r+ \sigma )^4+t^{3\beta}\eta_t^3  (\delta_{t-1}+d)^2r^2(\lambda_r + \sigma +d)^2(\lambda_r + \sigma )^4+t^{7\beta}\eta_t^7  d^4r^2(\lambda_r + \sigma )^8\right).
\]
Therefore, $\delta_t\leq C\lambda_r$ for $1\leq t\leq t^\star$. 

For $t>t^\star$, by \eqref{eqeqSNR}, we have, under $E_{t-1}$,
\[
\delta_{t-1}^2\leq C\min\left\{\gamma^2 dr\sigma^2 (\log d)^2 (t-1)^{\beta} \eta_{t-1},\sigma^2\right\}. 
\]
Again by Lemma \ref{lm: sum of prod}, we have 
\[
\widetilde Q_t\leq C\left(t^\beta \eta_t\sqrt{r} \delta_{t-1} \sigma^2+t^\beta \eta_t dr \sigma^2 + t^{3\beta} \eta_t^3 d^2r^2 \sigma^4\right).
\]
and
\[
\widetilde{\mathcal{V}}_t^{(1)}+\widetilde{\mathcal{V}}_t^{(2)}+\widetilde{\mathcal{V}}_t^{(3)}\leq t^{\beta}\eta_t\delta_{t-1}^2(\delta_{t-1} + \sigma )^2+t^{3\beta}\eta_t^3  (\delta_{t-1}+d)^2r^2(\delta_{t-1} + \sigma )^4+t^{7\beta}\eta_t^7  d^4r^2(\delta_{t-1} + \sigma )^8.
\]
By Lemma \ref{lm: max term}, we have
\[
\widetilde\Psi_t^{(1)}+\widetilde\Psi_t^{(2)}+\widetilde\Psi_t^{(3)}\leq t^{\beta}\eta_t\delta_{t-1}^2(\delta_{t-1} + \sigma )^2+t^{3\beta}\eta_t^3  (\delta_{t-1}+d)^2r^2(\delta_{t-1} + \sigma )^4+t^{7\beta}\eta_t^7  d^4r^2(\delta_{t-1} + \sigma )^8.
\]
Then with \eqref{eq:Rbound}, we have
\[
\delta_t^2\leq\delta_0^2\prod_{\tau=1}^t \left(1-\frac{\eta_\tau\lambda_r}{2}\right)+C_1 \gamma^2 dr\sigma^2 (\log d)^2 t^{\beta}\eta_t,
\]
which finalizes the proof.

\subsection{Proof of Theorem \ref{thm1}}
\label{sec: proof of thm1}
Define $\Delta_{t-1} = \Msgd_{t-1} - M$ and $\widehat{Z} = \widehat{Z}_1 + \widehat{Z}_2$ where 
\begin{equation*}
        \Munbs_n = M + \underbrace{\frac{1}{n} \displaystyle \sum^n_{t=1} I\{a_t = 1\} \xi_t \bX_t / \pi_t}_{\widehat{Z}_1} +  \underbrace{\frac{1}{n} \displaystyle \sum^n_{t=1} \left(\frac{I\{a_t = 1\} \langle \Delta_{t-1}, \bX_t \rangle \bX_t}{\pi_t} - \Delta_{t-1}\right)}_{\widehat{Z}_2}.
    \end{equation*}
    We can decompose the term $\widehat{m}_T - m_T$ as
\begin{equation}
\label{eq: small m decompose}
    \widehat{m}_T - m_T =  \underbrace{\left \langle \Uhat\Uhat^\top \widehat{Z} \Vhat\Vhat^\top, T \right\rangle}_{\text{negligible term}} + \underbrace{\left \langle \Uhat\Uhat^\top M \Vhat\Vhat^\top - M, T \right \rangle}_{\text{main term}},
\end{equation}
where $\Uhat$ and $\Vhat$ denote the left and right top-$r$ singular vectors of $\Munbs_n$. We use $\pmin$ to denote a constant lower bound of $\pi_t$ for all $t$, as $\beta = 0$. 

First we define the following $(d_1 +d_2) \times 2r$ matrices
    \begin{equation}
    \label{eq: define theta}
    \mathbf{\Theta} = \begin{pmatrix}
    \bU & 0\\
    0  & \bV
    \end{pmatrix}, \hspace{3mm }\widehat{\mathbf{\Theta}} = \begin{pmatrix}
    \Uhat & 0\\
    0  & \Vhat
    \end{pmatrix},
    \end{equation}
    where $U$ and $V$ are the top-$r$ singular vectors for $M$, and also define the $(d_1 +d_2) \times (d_1 +d_2)$ matrices
    \begin{equation}
    \label{eq: define AET}
    A = \begin{pmatrix}
    0 & M\\
    M^\top  & 0
    \end{pmatrix}, \hspace{3mm} \widehat{E} = \begin{pmatrix}
    0 & \widehat{Z}\\
    \widehat{Z}^\top  & 0
    \end{pmatrix}, \hspace{3mm} \text{and} \hspace{3mm}
    \tilde{T} = \begin{pmatrix}
    0 & T\\
    0  & 0
    \end{pmatrix}.
    \end{equation}
    We next apply the decomposition in \cite{xia2021normal} to our analysis. 
    Define $
    \mathfrak{B}^{-s}$ for $s\geq 1$ as 
    \begin{equation*}
    \mathfrak{B}^{-s}=\begin{cases}
         \begin{pmatrix}
    0 & U\Lambda^{-s}V^\top\\
    V\Lambda^{-s}U^\top  & 0
    \end{pmatrix}, \hspace{3mm} \text{if $s$ is odd},\\
    \begin{pmatrix}
    U\Lambda^{-s}U^\top & 0\\
    0 & V\Lambda^{-s}V^\top 
    \end{pmatrix}, \hspace{3mm} \text{if $s$ is even}\\
    \end{cases}
    \end{equation*}
    and 
    \begin{equation*}
         \mathfrak{B}^{0} = \mathfrak{B}^{\perp} =  \begin{pmatrix}
    U_\perp U^\top_\perp & 0 \\
    0 & V_\perp V_\perp^\top
    \end{pmatrix}.
    \end{equation*}
    We next state a necessary lemma before we can apply the decomposition in \cite{xia2021normal}.
    \begin{lemma}
    \label{lm:Opnorm}
    For any fixed unit vector $u,v\in \mathbb{S}^{d-1}$, under the assumptions of Theorem \ref{thm1}, as $n,d\rightarrow\infty$, we have $\|\widehat{Z}\| =O_p(\sigma \sqrt{d/n})$, and $u^\top \widehat Z v=O_p(\sigma/\sqrt{n})$.
    \end{lemma}
    By Assumption \ref{assum: SNR condition}, we have $\lambda_r \ge 2 \| \widehat{Z} \|$, and we can apply Theorem 1 in \cite{xia2021normal} that 
    \begin{equation*}
    \widehat{\mathbf{\Theta}}\widehat{\mathbf{\Theta}}^\top - \mathbf{\Theta}\mathbf{\Theta}^\top = \mathcal{S}_{A,1}(\widehat{E}) + \sum_{k\ge 2}^{\infty} \mathcal{S}_{A,k} (\widehat{E}),
    \end{equation*}
    \begin{equation}
    \label{eq: define S}
    \mathcal{S}_{A,k}(\widehat{E}) = \sum_{\mathbf{s} = s_1 + \ldots + s_{k+1} = k} (-1)^{1+\tau(\mathbf{s})}\cdot  \mathfrak{B}^{-s_1} \widehat{E} \mathfrak{B}^{-s_2} \widehat{E} \cdot \cdot \cdot \widehat{E} \mathfrak{B}^{-s_{k+1}},
    \end{equation}
    where $s_1, s_2,\ldots ,s_{k+1} \geq 0$ and $\tau(\mathbf{s}) = \sum_{j}^{k+1} I\{s_j >0\}$. 
    Given the definition of $\mathbf{\Theta}$, $\widehat{\mathbf{\Theta}}$ and $A$,  we have rewrite the main term as
    \begin{equation*}
      \left \langle \Uhat\Uhat^\top M \Vhat\Vhat^\top - M, T \right \rangle = \left \langle \widehat{\mathbf{\Theta}}\widehat{\mathbf{\Theta}}^\top A \widehat{\mathbf{\Theta}}\widehat{\mathbf{\Theta}}^\top - \mathbf{\Theta}\mathbf{\Theta}^\top A \mathbf{\Theta}\mathbf{\Theta}^\top, \tilde{T}\right \rangle.
    \end{equation*}
    By rearranging the terms of the above equation and then combining the decomposition of $\widehat{m}_T - m_T$ as in equation \eqref{eq: small m decompose}, the following decomposition 
     \begin{align*}
        \widehat{m}_T - m_T =& \left \langle \Uhat\Uhat^\top \widehat{Z} \Vhat\Vhat^\top, T \right \rangle  \numberthis \label{eq:18}\\ 
        & + \left \langle \mathcal{S}_{A,1}(\widehat{E}) A \mathbf{\Theta}\mathbf{\Theta}^\top + \mathbf{\Theta}\mathbf{\Theta}^\top A \mathcal{S}_{A,1}(\widehat{E}), \Tilde{T} \right \rangle \numberthis \label{eq:19}\\
        & +  \left \langle  \sum_{k\ge 2}^{\infty} \mathcal{S}_{A,k}A \mathbf{\Theta}\mathbf{\Theta}^\top + \mathbf{\Theta}\mathbf{\Theta}^\top A \sum_{k\ge 2}^{\infty} \mathcal{S}_{A,k}, \Tilde{T} \right \rangle \numberthis \label{eq:20}\\
        & + \left \langle (\widehat{\mathbf{\Theta}}\widehat{\mathbf{\Theta}}^\top - \mathbf{\Theta}\mathbf{\Theta}^\top) A (\widehat{\mathbf{\Theta}}\widehat{\mathbf{\Theta}}^\top - \mathbf{\Theta}\mathbf{\Theta}^\top), \Tilde{T} \right \rangle.  \numberthis \label{eq:21}
    \end{align*}
    With this decomposition, we will show that the equation \eqref{eq:19} is asymptotic normal, and the terms in \eqref{eq:18}, \eqref{eq:20}, and \eqref{eq:21} are negligible.  By the definition of the polynomial $\mathcal{S}_{A,k}(\widehat{E})$, 
    \begin{align*}
    \label{eq: re-write clt}
         & \left \langle \mathcal{S}_{A,1}(\widehat{E}) A \mathbf{\Theta}\mathbf{\Theta}^\top + \mathbf{\Theta}\mathbf{\Theta}^\top A \mathcal{S}_{A,1}(\widehat{E}), \Tilde{T} \right \rangle= \Big \langle \Uorg\Uorg^\top \widehat{Z}VV^\top, T\Big \rangle + \Big \langle UU^\top \widehat{Z} V_{\bot} V^\top_{\bot}, T \Big \rangle.
    \end{align*}
    The next lemma shows the asymptotic normality of \eqref{eq:19}.
    \begin{lemma}
    \label{lm:main term}
        Under the Assumptions of Theorem \ref{thm1} , as $n, d_1, d_2 \rightarrow \infty$, we have 
        \begin{equation*}
            \frac{\Big \langle \Uorg\Uorg^\top \widehat{Z}VV^\top, T\Big \rangle + \Big \langle UU^\top \widehat{Z} V_{\bot} V^\top_{\bot}, T \Big \rangle}{\sigma S/\sqrt{n}} \xrightarrow{d} \mathcal{N} ( 0, 1),
        \end{equation*}
        where
        \begin{equation*}
         S^2 = \int \frac{\Big \langle U_{\bot}U_{\bot}^\top \bX VV^\top+  U_1U_1^\top \bX V_{\bot} V^\top_{\bot}, T \Big \rangle ^2}{\pi_{\infty}(X)} dP_X,
    \end{equation*}
    \end{lemma}
    The following lemmas provide bounds on the negligible terms. 
    \begin{lemma}
    \label{lm: outter nelig}
    Under the assumptions of Theorem \ref{thm1}, as $n,d_1,d_2\rightarrow\infty$, 
    \begin{align*}
      \left \langle \Uhat\Uhat^\top \widehat{Z} \Vhat\Vhat^\top, T \right \rangle=& O_p\left( \frac{\sigma^2}{\lambda_r}( \|TV\|_{\mathrm{F}} + \|U^\top T \|_{\mathrm{F}})\frac{d\sqrt{r}\log d}{n}\right).
    \end{align*}
    \end{lemma}
    
    \begin{lemma}
    \label{lm: minor 1}
    Under the assumptions of Theorem \ref{thm1}, as $n,d_1,d_2\rightarrow\infty$, 
    \begin{align*}
        &\left \vert \left \langle  \sum_{k\ge 2}^{\infty} \mathcal{S}_{A,k}A \mathbf{\Theta}\mathbf{\Theta}^\top + \mathbf{\Theta}\mathbf{\Theta}^\top A \sum_{k\ge 2}^{\infty} \mathcal{S}_{A,k}, \Tilde{T} \right \rangle \right \vert =O_p\left(\frac{\sigma^2}{\lambda_r^2} ( \|U^\top T\|_{\mathrm{F}} +\|TV\|_{\mathrm{F}} ) \frac{d\sqrt{r}}{n}\right).
    \end{align*}
    \end{lemma}

    \begin{lemma}
    \label{lm : minor 2}
    Under the assumptions of Theorem \ref{thm1}, as $n,d_1,d_2\rightarrow\infty$, 
    \begin{align*}
        &\left \vert \Big \langle (\widehat{\mathbf{\Theta}}\widehat{\mathbf{\Theta}}^\top - \mathbf{\Theta}\mathbf{\Theta}^\top) A (\widehat{\mathbf{\Theta}}\widehat{\mathbf{\Theta}}^\top - \mathbf{\Theta}\mathbf{\Theta}^\top), \Tilde{T} \Big \rangle \right \vert=O_p\left(\frac{\sigma^2}{\lambda_r} ( \|U^\top T\|_{\mathrm{F}} +\|TV\|_{\mathrm{F}} ) \frac{d}{n}\right).
    \end{align*}
    \end{lemma}
    Recall that $\pi_{\infty} = \lim_{t\rightarrow \infty}\pi_t$ is lower bounded by $p_0>0$. The lower bound for the $S^2$ is 
    \begin{align*}
        S^2 \ge\frac{1}{p_0} \Big( \| \bV^\top T^\top \bU_{\bot}\|^2_{\mathrm{F}} + \| \bU^\top T \bV_{\bot} \|^2_{\mathrm{F}} \Big ).
    \end{align*}
    By Assumption \ref{assum: inco_assum}, we have $
    \|U^\top T V\|_{\mathrm{F}}^2 \leq\frac{r}{d_1}\|U^\top T\|_{\mathrm{F}}^2$. 
    Since $
        \|U^\top T V_{\bot}\|_{\mathrm{F}}^2 = \|U^\top T \|_{\mathrm{F}}^2 - \|U^\top T V\|_{\mathrm{F}}^2$, we have
    \begin{equation}
        \label{eq: ratio lim}
        \frac{\| \bV^\top T^\top \bU_{\bot}\|^2_{\mathrm{F}} + \| \bU^\top T \bV_{\bot} \|^2_{\mathrm{F}}}{\| TV\|^2_{\mathrm{F}} + \| U^\top T\|^2_{\mathrm{F}}}  = 1 - \frac{\| \bV^\top T^\top \bU\|^2_{\mathrm{F}} + \| \bU^\top T \bV \|^2_{\mathrm{F}}}{\| TV\|^2_{\mathrm{F}} + \| U^\top T\|^2_{\mathrm{F}}}  \rightarrow 1,
    \end{equation}
    as $d_1,d_2 \rightarrow \infty$.
    Combining \eqref{eq: ratio lim} and Lemmas \ref{lm:main term}--\ref{lm : minor 2}, we conclude the proof.

\subsection{Proof of Theorem \ref{thm2}}
\label{sec: proof of theorem 2}
We separately prove that $\hat{S}^2$ and $\hat{\sigma}^2$ converge in probability.

\begin{enumerate}
    \item \textbf{Consistency of $\hat{S}^2$}
\end{enumerate}

We first realize that we can write
\begin{align*}
    \hat{S}^2 =& \frac{1}{n}\sum_{t=1}^n \frac{I\{ a_t=1\}}{\pi_t^2} \Big\langle \Usp_{t-1,\bot}\Usp^\top_{t-1,\bot} X_t \Vsp_{t-1}\Vsp_{t-1}^\top,T \Big\rangle^2 \\
    +& \frac{1}{n}\sum_{t=1}^n \frac{I\{ a_t=1\}}{\pi_t^2} \Big\langle \Usp_{t-1}\Usp_{t-1}^\top X_t \Vsp_{t-1,\bot}\Vsp^\top_{t-1,\bot} ,T\Big\rangle^2 \\
    + & \frac{2}{n}\sum_{t=1}^n \frac{I\{ a_t=1\}}{\pi_t^2} \Big\langle \Usp_{t-1,\bot}\Usp^\top_{t-1,\bot} \bX_t \Vsp_{t-1}\Vsp_{t-1}^\top,T \Big\rangle \Big\langle \Usp_{t-1}\Usp_{t-1}^\top X_t \Vsp_{t-1,\bot}\Vsp^\top_{t-1,\bot} ,T\Big\rangle,
\end{align*}
Define 
\begin{align*}
    \tilde{S}^2 = &\frac{1}{n}\sum_{t=1}^n \frac{I\{ a_t=1\}}{\pi_t^2} \Big\langle U_{\bot}U^\top_{\bot} \bX_t VV^\top,T \Big\rangle^2 + \frac{1}{n}\sum_{t=1}^n \frac{I\{a_t=1\}}{\pi_t^2} \Big\langle UU^\top X_t V_{\bot}V^\top_{\bot} ,T\Big\rangle^2 \\
    + & \frac{2}{n}\sum_{t=1}^n \frac{I\{ a_t=1\}}{\pi_t^2} \Big\langle U_{\bot}U^\top_{\bot} X_t VV^\top,T \Big\rangle \Big\langle UU^\top X_t V_{\bot}V^\top_{\bot} ,T\Big\rangle.
\end{align*}
Since $\pi_{\infty} = \bP(a_t=1|X_t, \mathcal{F}_{t-1})$ is bounded away from zero, we can achieve $\tilde{S}^2 / S^2\rightarrow 1$ immediately by a martingale LLN, see for example, Theorem 2.19 from \cite{hall2014martingale}. We next show $ (\hat{S}^2 - \tilde{S}^2 )/S^2 \xrightarrow{p} 0$. According to Theorem \ref{thm: sgd consistent} and \cite{wedin1972perturbation}, we have with probability at least $1 - \frac{4n}{d^\gamma}$,
\begin{equation*}
   \max\left\{ \| \Usp_t\Usp_t^\top - UU^\top \|_{\mathrm{F}},\| \Vsp_t\Vsp_t^\top - VV^\top \|_{\mathrm{F}}\right\}  \le C_1\frac{\sigma}{\lambda_r} \sqrt{\frac{dr\log^2 d}{t^{\alpha}}}.
\end{equation*}
Since $X_t$ is Gaussian and independent of $(\Usp_{t-1}, \Vsp_{t-1})$, 
\begin{align*}
    &\mathbb{E}_X\left |\Big\langle \Usp_{t-1,\bot}\Usp^\top_{t-1,\bot} X_t \Vsp_{t-1}\Vsp_{t-1}^\top,T \Big\rangle^2 - \Big\langle U_{\bot}U^\top_{\bot} \bX_t VV^\top,T \Big\rangle^2 \right |\\
    \le & C_1^2 \|T\|^2_{\mathrm{F}} \frac{\sigma^2}{\lambda_r^2} \frac{dr\log^2 d}{t^{\alpha}} + C_1\|T\|_{\mathrm{F}}(\|U^\top T\|_{\mathrm{F}}+\|TV\|_{\mathrm{F}}) \frac{\sigma}{\lambda_r}\sqrt{\frac{dr\log^2 d}{t^{\alpha}}},
\end{align*}
Therefore as $n=o(d^\gamma)$, 
\begin{align*}
    & \frac{1}{n}\frac{4}{\pmin^2} \sum_{t = 1}^n \left |\Big\langle \Usp_{t-1,\bot}\Usp^\top_{t-1,\bot} X_t \Vsp_{t-1}\Vsp{t-1}^\top,T \Big\rangle^2 - \Big\langle U_{\bot}U^\top_{\bot} \bX_t VV_t^\top,T \Big\rangle^2\right |\\
    = & O_p\left(\|T\|^2_{\mathrm{F}} \frac{\sigma^2}{\lambda_r^2}\frac{1}{n} \sum_{t = 1}^n\frac{dr\log^2 d}{t^{\alpha}} + \|T\|_{\mathrm{F}}(\|U^\top T\|_{\mathrm{F}}+\|TV\|_{\mathrm{F}})  \frac{\sigma}{\lambda_r}\frac{1}{n} \sum_{t = 1}^n\sqrt{\frac{dr\log^2 d}{t^{\alpha}}}\right) \\
    = & O_p\left(\|T\|^2_{\mathrm{F}} \frac{\sigma^2}{\lambda_r^2}\frac{dr\log^2 d}{n^{\alpha}} + \|T\|_{\mathrm{F}}(\|U^\top T\|_{\mathrm{F}}+\|TV\|_{\mathrm{F}})  \frac{\sigma}{\lambda_r}\sqrt{\frac{dr\log^2 d}{n^{\alpha}}}\right).
\end{align*}
The bounds on the other two terms in $\hat{S}^2 - \tilde{S}^2 $ share the same argument and are therefore omitted. By Assumptions \ref{assum: null space} and \ref{assum: SNR condition}, we have
\begin{align*}
    & \frac{\|T\|_{\mathrm{F}}\left(\|U^\top_{\bot}TV\|_{\mathrm{F}} + \|U^\top TV_{\bot}\|_{\mathrm{F}}\right)}{S^2}\frac{\sigma}{\lambda_r} \sqrt{\frac{dr\log^2 d}{n^{\alpha}}}
    \le C\frac{\sigma}{\lambda_r} \sqrt{\frac{d^2\log^2 d}{n^{\alpha}}} \rightarrow 0,
\end{align*}
\begin{align*}
    &\frac{\|T\|^2_{\mathrm{F}}}{S^2} \frac{\sigma^2}{\lambda_r^2} \frac{dr\log^2 d}{n^{\alpha}}
    \le  \frac{\|T\|^2_{\mathrm{F}}}{\|TV\|^2_{\mathrm{F}} +\|U^\top V\|^2_{\mathrm{F}}}\frac{\sigma^2}{\lambda_r^2} \frac{dr\log^2 d}{n^{\alpha}} \le \frac{\sigma^2}{\lambda_r^2} \frac{d^2\log^2 d}{n^{\alpha}} \rightarrow 0,
\end{align*}
as $n, d_1, d_2 \rightarrow \infty$. Therefore $
    \hat{S}^2 /S^2\xrightarrow{p} \tilde{S}^2/S^2$.\vspace{1mm}

 \begin{enumerate}
\setcounter{enumi}{1}
    \item \textbf{Consistency of $\hat{\sigma}^2$}
\end{enumerate}

We have 

\begin{align*}
    \hat{\sigma}^2 = & \frac{1}{n}\sum_{t=1}^n \frac{I\{a_t = 1 \}}{\pi_t}\left(y_t - \langle \Msgd_{t-1} ,\bX_t \rangle \right)^2 \\
    =& \underbrace{\frac{1}{n}\sum_{t=1}^n \frac{I\{a_t = 1 \}}{\pi_t} \langle M -  \Msgd_{t-1} ,\bX_t \rangle^2}_{\uppercase\expandafter{\romannumeral1}} +  \underbrace{\frac{2}{n} \sum_{t=1}^n \frac{I\{a_t = 1 \}}{\pi_t} \langle M -  \Msgd_{t-1} ,\bX_t \rangle \xi_t}_{\uppercase\expandafter{\romannumeral2}}\\
    &+ \underbrace{\frac{1}{n}\sum_{t=1}^n \frac{I\{a_t = 1 \}}{\pi_t}\xi_t^2}_{\uppercase\expandafter{\romannumeral3}}.
\end{align*}
Note that $\pmin$ is a nonzero constant. By Theorem \ref{thm: sgd consistent} and Assumption \ref{assum: SNR condition}, we have
\begin{align*}
   \uppercase\expandafter{\romannumeral2} = O_{p}\left(\frac1n\sum_{t=1}^n\frac{\sigma^2 dr \log^2d}{x^\alpha} \right)=O_{p}\left( \frac{\sigma^2  dr \log^2d}{n^\alpha}\right)=o_p(1),\\
   \uppercase\expandafter{\romannumeral1} = O_{p}\left(\frac1n\sum_{t=1}^n\sigma^2\sqrt\frac{ dr \log^2d}{x^\alpha} \right)=O_{p}\left( \sigma^2\sqrt\frac{  dr \log^2d}{n^\alpha}\right)=o_p(1).
\end{align*}
Combine the above results, we conclude the proof of the consistency of $\hat{\sigma}^2$, and consequently
\begin{equation*}
    \frac{\widehat{m}_T - m_T}{\hat{\sigma} \hat{S} /\sqrt{n}} \xrightarrow{d} \mathcal{N}(0,1).
\end{equation*}

\subsection{Proof of Theorem \ref{thm:value inference}}
\label{sec:proof opt value}
Define $\Df := \|M_1 -M_0\|_{\mathrm{F}}$. Without loss of generality, we assume $\sigma_1 \ge \sigma_0$ throughout the proof. We first state twos lemmas used in the proof.

\begin{lemma}
\label{Lemma 1}
    Under the conditions of Theorem \ref{thm: sgd consistent}, we have, for some constant $C_1$,
    \begin{equation*}
        \bP(\hat{a}(X_t) \ne a^*(X_t)|\mathcal{F}_{t-1}) \le C_1 \frac{\sum_{i=0}^1\| \Msgd_{i,t-1} - M_i\|_{\mathrm{F}}}{\Df}.
    \end{equation*}
\end{lemma}

\begin{lemma}
\label{lemma 2}
    Under the conditions of Theorem \ref{thm: sgd consistent} and Theorem \ref{thm:value inference}, we have
    $$
    \frac{1}{\sqrt{n}} \sum_{t=1}^n \left \vert \left\langle  M_{\hat{a}(X_t)} - M_{a^*(X_t)}, X_t\right\rangle \right\vert = o_p(\sigma_1),\quad\text{as }n, d \rightarrow \infty.
    $$
\end{lemma}

With the above lemmas, we are ready to prove Theorem \ref{thm:value inference}. First of all, recall that we have our mean optimal outcome estimator as 
\begin{equation*}
    \widehat{V}_n = \frac{1}{n}\sum_{t=1}^n \frac{I\{ a_t = \hat{a}(X_t)\}}{1 - \hat{e}_t}\left(y_t - \left \langle \Msgd_{\hat{a}(X_t), t-1}, X_t \right\rangle \right) + \left \langle \Msgd_{\hat{a}(X_t), t-1}, X_t \right \rangle.
\end{equation*}
We first define
\begin{equation*}
    \widetilde{V}_n = \frac{1}{n}\sum_{t=1}^n  \frac{I\{ a_t = \hat{a}(X_t)\}}{1 - e_t} \left( y_t -  \left \langle M_{\hat{a}(X_t)}, X_t\right \rangle\right) + \left \langle M_{\hat{a}(X_t)}, X_t\right \rangle,
\end{equation*}
and 
\begin{equation*}
    \bar{V}_n = \frac{1}{n}\sum_{t=1}^n \frac{I\{ a_t = a^*(X_t)\}}{\bP_t(a_t = a^*(X_t))} \left(y_t - \left \langle M_{a^*(X_t)}, X_t \right\rangle \right) + \left \langle M_{a^*(X_t)}, X_t \right\rangle,
\end{equation*}
where $\bP_t(\cdot)=\bP(\cdot|\mathcal{F}_{t-1},X_t)$. 
Then we have
\begin{equation*}
     \frac{\sqrt{n}(\widehat{V}_n - V^*)}{S_V} = \frac{\sqrt{n}(\widehat{V}_n - \widetilde{V}_n )}{S_V} +  \frac{\sqrt{n}(\widetilde{V}_n - \bar{V}_n)}{S_V} + \frac{\sqrt{n}(\bar{V}_n - V^*)}{S_V},
\end{equation*}
and we will show that $\sqrt{n}(\bar{V}_n - V^*)/S_V$ is asymptotically normal and its variance dominates those of the negligible terms. 

Step 1: Showing $\sqrt{n}( \widehat{V}_n -  \widetilde{V}_n)/S_V \xrightarrow{p} 0$ as $n, d \rightarrow \infty$.

\begin{align*}
    \sqrt{n}\left( \widehat{V}_n -  \widetilde{V}_n \right)
    = 
    & \frac{1}{\sqrt{n}} \sum_{t=1}^n \frac{I\{a_t = \hat{a}(X_t)\}}{1 - e_t} \left \langle M_{\hat{a}(X_t)} - \Msgd_{\hat{a}(X_t), t-1}, X_t\right\rangle - \left \langle M_{\hat{a}(X_t)} - \Msgd_{\hat{a}(X_t), t-1}, X_t\right\rangle \\
    = & \frac{1}{\sqrt{n}} \sum_{t=1}^n \left( \frac{I\{a_t = \hat{a}(X_t)\}}{1 - e_t} - 1\right)\left \langle M_{\hat{a}(X_t)} - \Msgd_{\hat{a}(X_t), t-1}, X_t\right\rangle. 
\end{align*}
We notice that 
\begin{align*}
    & \bE \left [\left( \frac{I\{a_t = \hat{a}(X_t)\}}{1 - e_t} - 1\right)\left \langle M_{\hat{a}(X_t)} - \Msgd_{\hat{a}(X_t), t-1}, X_t\right\rangle \Big | \mathcal{F}_{t-1}\right] \\
    = & \bE \left [\bE\left[ \frac{I\{a_t = \hat{a}(X_t)\}}{1 - e_t} - 1 \Big | \mathcal{F}_{t-1}, X_t\right]\left \langle M_{\hat{a}(X_t)} - \Msgd_{\hat{a}(X_t), t-1}, X_t\right\rangle \Big | \mathcal{F}_{t-1}\right] = 0,
\end{align*}
where the last equality is due to the fact that $1 - e_t = \bP(a_t = \hat{a}(X_t)|\mathcal{F}_{t-1},X_t)$. Next, since $1-e_t$ is lower bounded by a positive constant and $X_t$ is Gaussian, we have
\begin{align*}
    & \bE \left [\left( \frac{I\{a_t = \hat{a}(X_t)\}}{1 - e_t} - 1\right)^2\left \langle M_{\hat{a}(X_t)} - \Msgd_{\hat{a}(X_t), t-1}, X_t\right\rangle^2 \Big | \mathcal{F}_{t-1}\right]\\
    = & \bE \left [\left( \frac{I\{a_t = \hat{a}(X_t)\}}{1 - e_t} - 1\right)^2\left \langle M_{1} - \Msgd_{1, t-1}, X_t\right\rangle^2 I\{\hat{a}(X_t) = 1\}\Big | \mathcal{F}_{t-1}\right] \\
    & + \bE \left [\left( \frac{I\{a_t = \hat{a}(X_t)\}}{1 - e_t} - 1\right)^2\left \langle M_{0} - \Msgd_{0, t-1}, X_t\right\rangle^2 I\{\hat{a}(X_t) = 0\}\Big | \mathcal{F}_{t-1}\right]\\
    \le & C_1 \left (\left \| M_{1} - \Msgd_{1, t-1}\right\|_{\mathrm{F}}^2 + \left \| M_{0} - \Msgd_{0, t-1}\right\|_{\mathrm{F}}^2 \right),
\end{align*}
for some positive $C_1$. Then by Assumption \ref{assum:Optimal-Gap}, we have
\begin{align*}
\label{eq:first part}
    & \frac{1}{\sqrt{n}} \sum_{t=1}^n \left( \frac{I\{a_t = \hat{a}(X_t)\}}{1 - e_t} - 1\right)\left \langle M_{\hat{a}(X_t)} - \Msgd_{\hat{a}(X_t), t-1}, X_t\right\rangle= o_p(\sigma_1).
\end{align*}
To see that, we use $\left\| M_{1} - \Msgd_{1, t-1}\right\|_{\mathrm{F}} \le C_1\gamma\sigma_1 \sqrt{\frac{dr\log^2 d}{t^{\alpha-\beta}}}$ by Theorem \ref{thm: sgd consistent}, and therefore
\begin{equation*}
    \frac{1}{n}\sum_{t=1}^n \left (\left \| M_{1} - \Msgd_{1, t-1}\right\|_{\mathrm{F}}^2 + \left \| M_{0} - \Msgd_{0, t-1}\right\|_{\mathrm{F}}^2 \right) \leq C_1\gamma^2\sigma_1^2 \frac{dr\log^2d}{n^{\alpha - \beta}},
\end{equation*}
with high probability. Meanwhile, we notice that
\begin{equation}
\label{eq:lower bound SV}
    S_V \ge \sqrt{\sigma_0^2 + \mathrm{Var}[\langle M_{a^*(X)}, X\rangle]}.
\end{equation}
We can conclude that $\sqrt{n}(\widehat{V}_n - \widetilde{V}_n)/S_V = o_p(1).$

Step 2: Showing $\sqrt{n}(\widetilde{V}_n - \bar{V}_n )/S_V \xrightarrow{p} 0$ as $n,d \rightarrow \infty$.
\begin{align*}
    &\sqrt{n}\left(\widetilde{V}_n - \bar{V}_n \right)\\
    = & \frac{1}{\sqrt{n}} \sum_{t=1}^n \frac{I\{a_t = \hat{a}(X_t)\}}{1 - e_t} \left( y_t - \left \langle M_{\hat{a}(X_t)}, X_t\right \rangle\right) + \left \langle M_{\hat{a}(X_t)}, X_t\right \rangle \\
    & -\frac{1}{\sqrt{n}} \sum_{t=1}^n \frac{I\{a_t = a^*(X_t)\}}{\bP_t(a_t = a^*(X_t))}\left( y_t - \left \langle M_{a^*(X_t)}, X_t\right \rangle\right) + \left \langle M_{a^*(X_t)}, X_t\right \rangle \\
    = & \frac{1}{\sqrt{n}} \sum_{t=1}^n \frac{I\{a_t = \hat{a}(X_t)\}}{1 - e_t} \left( y_t - \left \langle M_{a^*(X_t)}, X_t\right \rangle\right) + \frac{1}{\sqrt{n}} \sum_{t=1}^n \frac{I\{a_t = \hat{a}(X_t)\}}{1 - e_t}  \left \langle M_{a^*(X_t)} -M_{\hat{a}(X_t)} , X_t\right \rangle \\
    & - \frac{1}{\sqrt{n}} \sum_{t=1}^n \frac{I\{a_t = a^*(X_t)\}}{\bP_t(a_t = a^*(X_t))}\left( y_t - \left \langle M_{a^*(X_t)}, X_t\right \rangle\right) - \frac{1}{\sqrt{n}} \sum_{t=1}^n \left \langle M_{a^*(X_t)} -M_{\hat{a}(X_t)} , X_t\right \rangle \\
    =& I +II,
\end{align*}
where 
\begin{align*}
    I = \frac{1}{\sqrt{n}} \sum_{t=1}^n \left( \frac{I\{a_t = \hat{a}(X_t)\}}{1 - e_t} - \frac{I\{a_t = a^*(X_t)\}}{\bP_t(a_t = a^*(X_t))} \right) \left( y_t - \left \langle M_{a^*(X_t)}, X_t\right \rangle\right),
\end{align*}
and
\begin{align*}
    II = \frac{1}{\sqrt{n}} \sum_{t=1}^n \left( \frac{I\{a_t = \hat{a}(X_t)\}}{1 - e_t} - 1\right)\left \langle M_{a^*(X_t)} -M_{\hat{a}(X_t)} , X_t\right \rangle.
\end{align*}
Then we realize that
\begin{align*}
    I = &  \frac{1}{\sqrt{n}} \sum_{t=1}^n \left( \frac{I\{a_t = \hat{a}(X_t)\}}{1 - e_t} - \frac{I\{a_t = a^*(X_t)\}}{\bP_t(a_t = a^*(X_t))} \right) \left( y_t - \left \langle M_{a^*(X_t)}, X_t\right \rangle\right)I\{\hat{a}(X_t) = a^*(X_t)\}  \\
    & + \frac{1}{\sqrt{n}} \sum_{t=1}^n \left( \frac{I\{a_t = \hat{a}(X_t)\}}{1 - e_t} - \frac{I\{a_t = a^*(X_t)\}}{\bP_t(a_t = a^*(X_t))} \right) \left( y_t - \left \langle M_{a^*(X_t)}, X_t\right \rangle\right)I\{\hat{a}(X_t) \ne a^*(X_t)\},
\end{align*}
while the first term is zero due to the fact that $\hat{a}(X_t) = a^*(X_t)$ implies $1 - e_t = \bP_t(a_t = \hat{a}(X_t)) = \bP_t(a_t = a^*(X_t))$. For the second term,
\begin{align*}
    & \frac{1}{\sqrt{n}} \sum_{t=1}^n \left( \frac{I\{a_t = \hat{a}(X_t)\}}{1 - e_t} - \frac{I\{a_t = a^*(X_t)\}}{\bP_t(a_t = a^*(X_t))} \right) \left( y_t - \left \langle M_{a^*(X_t)}, X_t\right \rangle\right)I\{\hat{a}(X_t) \ne a^*(X_t)\} \\
    = & \underbrace{\frac{1}{\sqrt{n}} \sum_{t=1}^n \left( \frac{I\{a_t = \hat{a}(X_t)\}}{1 - e_t} - \frac{I\{a_t = a^*(X_t)\}}{\bP_t(a_t = a^*(X_t))} \right) \left( y_t - \left \langle M_{a^*(X_t)}, X_t\right \rangle\right)I\{\hat{a}(X_t) \ne a^*(X_t)\} I\{a_t = \hat{a}(X_t) \}}_{\text{i}} \\
    & + \underbrace{\frac{1}{\sqrt{n}} \sum_{t=1}^n \left( \frac{I\{a_t = \hat{a}(X_t)\}}{1 - e_t} - \frac{I\{a_t = a^*(X_t)\}}{\bP_t(a_t = a^*(X_t))} \right) \left( y_t - \left \langle M_{a^*(X_t)}, X_t\right \rangle\right)I\{\hat{a}(X_t) \ne a^*(X_t)\}I\{a_t = a^*(X_t)\}}_{\text{ii}}.
\end{align*}
Then we have
\begin{align*}
    \text{i} =& \frac{1}{\sqrt{n}}\sum_{t=1}^n \frac{I\{a_t = \hat{a}(X_t)\}}{1 - e_t}\xi_tI\{\hat{a}(X_t)\ne a^*(X_t) \}I\{a_t = \hat{a}(X_t) \}\\
    & + \frac{1}{\sqrt{n}}\sum_{t=1}^n \frac{I\{a_t = \hat{a}(X_t)\}}{1 - e_t}\left \langle M_{\hat{a}(X_t)} -  M_{a^*(X_t)}, X_t \right\rangle I\{\hat{a}(X_t)\ne a^*(X_t) \}I\{a_t = \hat{a}(X_t) \}.
\end{align*}
We note that $\bE[\xi_t|\mathcal{F}_{t-1},X_t]=0$, $\bE[\xi_t^2|\mathcal{F}_{t-1},X_t]\le \sigma_1^2$, and $1-e_t$ is lower bounded. By the result of Lemma \ref{Lemma 1}, Theorem \ref{thm: sgd consistent}, and Assumption \ref{assum:Optimal-Gap}, we conclude that
\begin{equation}
\label{eq:part 2 noise(i)}
     \frac{1}{\sqrt{n}}\sum_{t=1}^n \frac{I\{a_t = \hat{a}(X_t)\}}{1 - e_t}\xi_tI\{\hat{a}(X_t)\ne a^*(X_t) \} =o_p(\sigma_1).
\end{equation}
 On the other hand, by Lemma \ref{lemma 2}, we have that 
\begin{equation*}
    \frac{1}{\sqrt{n}}\sum_{t=1}^n \left \vert\left \langle M_{\hat{a}(X_t)} -  M_{a^*(X_t)}, X_t \right\rangle \right \vert = o_p(\sigma_1).
\end{equation*}
Combined with \eqref{eq:part 2 noise(i)}, we have $\text{i} = o_p(\sigma_1)$. 

The term $\text{ii}$ can be bounded with a similar proof, as
\begin{align*}
    \vert \text{ii} \vert = \left \vert \frac{1}{\sqrt{n}} \sum_{t=1}^n \frac{I\{a_t = a^*(X_t)\}}{\bP_t(a_t = a^*(X_t))}  \xi_t I\{\hat{a}(X_t) \ne a^*(X_t)\} \right \vert= o_p(1).
\end{align*} Next, we recall that
\begin{align*}
     II = & \frac{1}{\sqrt{n}} \sum_{t=1}^n \left( \frac{I\{a_t = \hat{a}(X_t)\}}{1 - e_t} - 1\right)\left \langle M_{a^*(X_t)} -M_{\hat{a}(X_t)} , X_t\right \rangle \\
     & \le c \frac{1}{\sqrt{n}}\sum_{t=1}^n \left \vert\left \langle M_{\hat{a}(X_t)} -  M_{a^*(X_t)}, X_t \right\rangle \right \vert.
\end{align*}
Then by Lemma \ref{lemma 2} again, we have $II = o_p(\sigma_1)$, and by \eqref{eq:lower bound SV} we have
\begin{align*}
    \sqrt{n}\left( \widetilde{V}_n - \bar{V}_n\right)\Big/S_V = (I + II)/S_V = o_p(1).
\end{align*}

Step 3: The asymptotic normality of $\sqrt{n} \left(\bar{V}_n - V^* \right)/S_V$.

We have
\begin{align*}
    \bar{V}_n - V^* =& \frac{1}{n}\sum_{t=1}^n \frac{I\{ a_t = a^*(X_t)\}}{\bP_t(a_t = a^*(X_t))} \left(y_t - \left \langle M_{a^*(X_t)}, X_t \right\rangle \right) + \left \langle M_{a^*(X_t)}, X_t \right\rangle - \bE\left[\left\langle M_{a^*(X)}, X \right\rangle \right] \\
    =& \frac{1}{n}\sum_{t=1}^n \underbrace{\frac{I\{ a_t = a^*(X_t)\}}{\bP_t(a_t = a^*(X_t))} \xi_t}_{W_t} + \frac{1}{n}\sum_{t=1}^n \underbrace{\left \langle M_{a^*(X_t)}, X_t \right\rangle - \bE\left[\left \langle M_{a^*(X)}, X \right\rangle \right]}_{H_t}.
\end{align*}
Note that we have
\begin{equation}
\label{eq:H converge}
    \frac{1}{n}\sum_{t=1}^n \bE \left[ H_t^2 | \mathcal{F}_{t-1}\right] = \mathrm{Var}\left[ \left \langle M_{a^*(X)}, X \right\rangle\right].
\end{equation}
Meanwhile,
\begin{align*}
    \frac{1}{n}\sum_{t=1}^n \bE \left[ W_t^2 | \mathcal{F}_{t-1}\right] &= \frac{1}{n}\sum_{t=1}^n \bE \left[ \frac{I\{ a_t = a^*(X_t)\}}{\bP_t(a_t = a^*(X_t))^2} \xi_t^2\Big| \mathcal{F}_{t-1}\right] \\
    & = \frac{1}{n}\sum_{t=1}^n \bE \left[ \bE \left[\frac{I\{ a_t = a^*(X_t)\}}{\bP_t(a_t = a^*(X_t))^2} \xi_t^2 \Big | X_t\right]\Big | \mathcal{F}_{t-1}\right] \\
    & = \frac{1}{n} \sum_{t=1}^n\bE \left[ \frac{\sigma^2_{a^*(X_t)}}{\bP_t(a_t = a^*(X_t))} \right].
\end{align*}
Since  $\bP_t(a_t = a^*(X_t))$ is lower bounded, we have
\begin{equation*}
    \frac{1}{n} \sum_{t=1}^n\bE \left[ \frac{\sigma^2_{a^*(X_t)}}{\bP_t(a_t = a^*(X_t))} \right] \rightarrow \int \frac{\sigma_1^2I\{\left \langle M_{1} - M_0, X \right\rangle>0 \}+\sigma_0^2I\{\left \langle M_1 - M_0, X \right\rangle\le 0\}}{1 - e^*_{\infty}(X)} dP_X.
\end{equation*}
Finally, the proof of theorem \ref{thm:value inference} is concluded by combining Step 1, 2 and 3.

\subsection{Proof of Theorem \ref{thm:consistent}}
\label{sec:opt value consistency proof p}
~

Step 1: Proof the consistency for $\hat{\sigma}_{i,t}$.
Therefore, the consistency of $\hat{\sigma}_{i,t}$ shares exactly the same argument as the proof of $\hat{\sigma}_{i,t}$ in section \ref{sec: proof of theorem 2}. The only difference is that we apply the Assumption \ref{assum:Optimal-Gap} in this case to ensure that $dr/n^{\alpha-\beta} = o(1)$. We therefore skip the proof for the consistency of  $\hat{\sigma}_{i,t}$.

Step 2: The consistency of the first term in \eqref{eq:emp SV}.
We refer to this term as term I and show
\begin{equation*}
    I \xrightarrow{p} \int \frac{a^*(X)\sigma_1^2 + (1-a^*(X))\sigma_0^2}{1 - e_{\infty}}dP_X.
\end{equation*}
We first realize that we can re-write $I$ as
\begin{align*}
    I = & \frac{1}{n}\sum_{t=1}^n \frac{(\hat{\sigma}_1^2 - \sigma_1^2)I\{\widehat{\Delta}_{X_t} > 0\} + (\hat{\sigma}_0^2 - \sigma_0^2)I\{\widehat{\Delta}_{X_t} \le 0\}}{1 - e_t} \\
    & + \frac{1}{n}\sum_{t=1}^n \frac{\sigma^2_1\left( I\{\widehat{\Delta}_{X_t} > 0\} - I\{\Delta_{X_t} > 0\}\right) + \sigma^2_0\left( I\{\widehat{\Delta}_{X_t} \le 0\} - I\{\Delta_{X_t} \le 0\}\right)}{1 - \hat{e}_t} \\
    & + \frac{1}{n}\sum_{t=1}^n \frac{\sigma_1^2 I\{\Delta_{X_t} > 0\} + \sigma_0^2 I\{\Delta_{X_t} \le 0\}}{1 - e_t}.
\end{align*}
First of all, by \text{Step 1} and that $1-e_t$ is lower bounded, we immediately have 
\begin{equation*}
     \frac{1}{n}\sum_{t=1}^n \frac{(\hat{\sigma}_1^2 - \sigma_1^2)I\{\widehat{\Delta}_{X_t} > 0\} + (\hat{\sigma}_0^2 - \sigma_0^2)I\{\widehat{\Delta}_{X_t} \le 0\}}{1 - e_t} =o_p(\sigma_1^2).
\end{equation*}Meanwhile for the second term in I,
\begin{align*}
    & \frac{1}{n}\sum_{t=1}^n \frac{\sigma_1^2\left( I\{\widehat{\Delta}_{X_t} > 0\} - I\{\Delta_{X_t} > 0\}\right) + \sigma_0^2\left( I\{\widehat{\Delta}_{X_t} \le 0\} - I\{\Delta_{X_t} \le 0\}\right)}{1 - \hat{e}_t} \\
    \le & \frac{C\sigma_1^2}{n}\sum_{t=1}^n \left |I\{\widehat{\Delta}_{X_t} > 0\} - I\{\Delta_{X_t} > 0\}  \right| + \frac{C\sigma_0^2}{n}\sum_{t=1}^n \left | I\{\widehat{\Delta}_{X_t} \le 0\} - I\{\Delta_{X_t} \le 0\} \right|.
\end{align*}
Next we use the following lemma,
\begin{lemma}
\label{lemma 3}
    Under the conditions of Theorem \ref{thm: sgd consistent} and Theorem \ref{thm:value inference}, we have
    $$
    \frac{1}{n}\sum_{t=1}^n \left| I\{\widehat{\Delta}_{X_t} > 0\} - I\{ \Delta_{X_t} >0\}\right| \overset{p}{\rightarrow}0,
    \quad\text{as }n, d \rightarrow \infty.$$
\end{lemma}
It remains to show that 
\begin{equation*}
    \frac{1}{n}\sum_{t=1}^n \frac{\sigma_1^2 I\{\Delta_{X_t} > 0\} + \sigma_0^2 I\{\Delta_{X_t} \le 0\}}{1 - e_t} \xrightarrow{p} \int \frac{a^*(X)\sigma_1^2 + (1-a^*(X))\sigma_0^2}{1 - e_{\infty}}dP_X.
\end{equation*}
First recall that $e_\infty = \lim_{t\rightarrow \infty} \bP(a_t \ne a^*(X_t))$, and then we notice that
\begin{align*}
    & \left |\frac{1}{n}\sum_{t=1}^n \frac{\sigma_1^2 I\{\Delta_{X_t} > 0\} + \sigma_0^2 I\{\Delta_{X_t} \le 0\}}{1 - \bP_t(a_t \ne \hat{a}(X_t))} - \frac{1}{n}\sum_{t=1}^n \frac{\sigma_1^2 I\{\Delta_{X_t} > 0\} + \sigma_0^2 I\{\Delta_{X_t} \le 0\}}{1 -\bP_t(a_t \ne a^*(X_t))} \right| \\
    \le & \frac{1}{n} \sum_{t=1}^n \left( \sigma_1^2 I\{\Delta_{X_t} > 0\} + \sigma_0^2 I\{\Delta_{X_t} \le 0\}\right) \left| \bP_t(a_t \ne \hat{a}(X_t))-\bP_t(a_t \ne a^*(X_t))\right|\\
    \le & \frac{C_1(\sigma_1^2 +\sigma_0^2)}{n}\sum_{t=1}^n \bP_t(\hat{a}(X_t)\ne a^*(X_t)|\mathcal{F}_{t-1})\\
    \le & \frac{(\sigma_1^2 +\sigma_0^2)}{n}\sum_{t=1}^n  \frac{\sum_{i=0}^1\|\Msgd_{i,t-1}-M_i\|_{\mathrm{F}}}{\Df}\\
    \le &  C_1 (\sigma_1^2 + \sigma_0^2)\frac{\sigma_1}{\Df}\sqrt{\frac{dr\log^2 d}{n^{\alpha - \beta}}}.
\end{align*}
In addition, by Theorem \ref{thm: sgd consistent} and Assumption \ref{assum:Optimal-Gap}, the above expression is $o_p(\sigma_1^2)$. 
Therefore, 
\begin{equation*}
    \frac{1}{n}\sum_{t=1}^n \frac{\sigma_1^2 I\{\Delta_{X_t} > 0\} + \sigma_0^2 I\{\Delta_{X_t} \le 0\}}{1 - e_t} \xrightarrow{p}   \frac{1}{n}\sum_{t=1}^n \frac{\sigma_1^2 I\{\Delta_{X_t} > 0\} + \sigma_0^2 I\{\Delta_{X_t} \le 0\}}{1 -\bP_t(a_t \ne a^*(X_t))}.
\end{equation*}
By martingale LLN,
\begin{align*}
    \frac{1}{n}\sum_{t=1}^n \frac{\sigma_1^2 I\{\Delta_{X_t} > 0\} + \sigma_0^2 I\{\Delta_{X_t} \le 0\}}{1 -\bP(a_t \ne a^*(X_t))} \xrightarrow{p} \int \frac{\sigma_1^2 I\{\Delta_{X} > 0\} + \sigma_0^2 I\{\Delta_{X} \le 0\}}{1 -\bP(a_t \ne a^*(X))} dP_X.
\end{align*}
Finally by the continuous mapping theorem, we arrive at
\begin{equation*}
    I \xrightarrow{p} \int \frac{a^*(X)\sigma_1^2 + (1-a^*(X))\sigma_0^2}{1 - e_{\infty}}dP_X.
\end{equation*}

Step 3 The consistency of the second term in \eqref{eq:emp SV}.
We refer to this term as term II and show $II \xrightarrow{p} \mathrm{Var}\left[\langle M_{a^*(X)},X \rangle \right]$. Specifically, we divide the whole argument into two parts. We first show that 
\begin{equation}
\label{eq:var cons first part}
    \frac{1}{n}\sum_{t=1}^n \left\langle \Msgd_{\hat{a}(X_t), t-1}, X_t \right\rangle^2 \xrightarrow{p} \bE \left[ \langle M_{a^*(X)},X \rangle^2\right], 
\end{equation}
and
\begin{equation}
\label{eq:var cons second part}
    \left(\frac{1}{n}\sum_{t=1}^n \left\langle \Msgd_{\hat{a}(X_t), t-1}, X_t \right\rangle \right)^2 \xrightarrow{p} \bE \left[ \langle M_{a^*(X)},X \rangle\right]^2.
\end{equation}
We break down the proof of \eqref{eq:var cons first part} into the following steps with order.
\begin{enumerate}
    \item Proof of $ \frac{1}{n}\sum_{t=1}^n \left\langle \Msgd_{\hat{a}(X_t), t-1}, X_t \right\rangle^2 \xrightarrow{p}  \frac{1}{n}\sum_{t=1}^n \left\langle M_{\hat{a}(X_t)}, X_t \right\rangle^2$.
   We notice that 
    \begin{align*}
        \frac{1}{n}\sum_{t=1}^n \left\langle \Msgd_{\hat{a}(X_t), t-1}, X_t \right\rangle^2
        =& \frac{1}{n}\sum_{t=1}^n \left\langle M_{\hat{a}(X_t), t-1}, X_t\right\rangle^2 \\
         & + \frac{1}{n}\sum_{t=1}^n \left\langle \Msgd_{\hat{a}(X_t), t-1}- M_{\hat{a}(X_t)},X_t \right\rangle^2 \numberthis \label{eq:one term} \\
         & + \frac{2}{n}\sum_{t=1}^n \left\langle \Msgd_{\hat{a}(X_t), t-1}- M_{\hat{a}(X_t)},X_t \right\rangle  \left\langle M_{\hat{a}(X_t)}, X_t\right\rangle, \numberthis \label{eq:other term}
    \end{align*}
    and we show that \eqref{eq:one term} and \eqref{eq:other term} are both $o_p(\sigma_1^2)$. Note that
    \begin{align*}
        & \frac{1}{n}\sum_{t=1}^n \left\langle \Msgd_{\hat{a}(X_t), t-1}- M_{\hat{a}(X_t)},X_t \right\rangle^2 \\
        = & \frac{1}{n}\sum_{t=1}^n I\{ \hat{a}(X_t) =1\} \left\langle \Msgd_{1, t-1}- M_{1},X_t \right\rangle^2 +  \frac{1}{n}\sum_{t=1}^n I\{ \hat{a}(X_t) =0\} \left\langle \Msgd_{0, t-1}- M_{0},X_t \right\rangle^2.
    \end{align*}
    By Theorem 2.19 in \cite{hall2014martingale}, we have
    \begin{equation*}
        \frac{1}{n}\sum_{t=1}^n \left\langle \Msgd_{1, t-1}- M_{1},X_t \right\rangle^2 \xrightarrow{p} \frac{1}{n}\sum_{t=1}^n \bE \left[ \left\langle \Msgd_{1, t-1}- M_{1},X_t \right\rangle^2 \Big| \mathcal{F}_{t-1}\right],
    \end{equation*}
    where 
    \begin{align*}
        \frac{1}{n}\sum_{t=1}^n \bE \left[ \left\langle \Msgd_{1, t-1}- M_{1},X_t \right\rangle^2 \Big| \mathcal{F}_{t-1}\right] \le \frac{C}{n} \sum_{t=1}^n \left \| \Msgd_{1, t-1}- M_{1}\right\|^2_{\mathrm{F}}.
    \end{align*}
    By Theorem \ref{thm: sgd consistent} and Assumption \ref{assum:Optimal-Gap}, we have
    \begin{equation*}
        \frac{1}{n} \sum_{t=1}^n \left \| \Msgd_{1, t-1}- M_{1}\right\|^2_{\mathrm{F}} = o_p(\sigma_1^2).
    \end{equation*}
    On the other hand, we applied a similar argument to \eqref{eq:other term}. By Theorem 2.19 in \cite{hall2014martingale}, we have 
    \begin{align*}
        &\frac{1}{n}\sum_{t=1}^n I\{a_t = \hat{a}(X_t) \}\left\langle \Msgd_{1, t-1}- M_{1},X_t \right\rangle  \left\langle M_{1}, X_t\right\rangle \\
        \xrightarrow{p} & \frac{1}{n}\sum_{t=1}^n \bE \left[ I\{a_t = \hat{a}(X_t) \}\left\langle \Msgd_{1, t-1}- M_{1},X_t \right\rangle  \left\langle M_{1}, X_t\right\rangle \Big | \mathcal{F}_{t-1}\right].
    \end{align*}
    Notice that 
    \begin{align*}
        & \bE \left[ I\{a_t = \hat{a}(X_t) \}\left\langle \Msgd_{1, t-1}- M_{1},X_t \right\rangle  \left\langle M_{1}, X_t\right\rangle \Big | \mathcal{F}_{t-1}\right] 
        \le  C' \sqrt{r}\left\| \Msgd_{1,t-1} - M_1\right\|_{\mathrm{F}} \|M_1\|.
    \end{align*}
    By Theorem \ref{thm: sgd consistent}, we have that 
    \begin{equation*}
        \frac{2}{n}\sum_{t=1}^n \bE \left[ I\{a_t = \hat{a}(X_t) \}\left\langle \Msgd_{1, t-1}- M_{1},X_t \right\rangle  \left\langle M_{1}, X_t\right\rangle \Big | \mathcal{F}_{t-1}\right]  = o_p(\sigma_1^2).
    \end{equation*} Combining above results we have shown that \eqref{eq:one term} and \eqref{eq:other term} are all of smaller order, thus
    \begin{equation*}
        \frac{1}{n}\sum_{t=1}^n \left\langle \Msgd_{\hat{a}(X_t), t-1}, X_t \right\rangle^2 \xrightarrow{p}  \frac{1}{n}\sum_{t=1}^n \left\langle M_{\hat{a}(X_t)}, X_t \right\rangle^2.
    \end{equation*}

    \item Proof of $\frac{1}{n}\sum_{t=1}^n \left\langle M_{\hat{a}(X_t)}, X_t \right\rangle^2 \xrightarrow{p} \frac{1}{n}\sum_{t=1}^n \left\langle M_{a^*(X_t)}, X_t \right\rangle^2$.
    Similarly, we notice that 
    \begin{align*}
        \frac{1}{n}\sum_{t=1}^n \left\langle M_{\hat{a}(X_t)}, X_t \right\rangle^2 
        = & \frac{1}{n}\sum_{t=1}^n \left\langle  M_{a^*(X_t)}, X_t \right\rangle^2 \\
        & + \frac{1}{n}\sum_{t=1}^n \left\langle M_{\hat{a}(X_t)} - M_{a^*(X_t)}, X_t \right\rangle^2 \numberthis \label{eq:one} \\
        & + \frac{2}{n}\sum_{t=1}^n \left\langle M_{\hat{a}(X_t)} - M_{a^*(X_t)}, X_t \right\rangle  \left\langle  M_{a^*(X_t)}, X_t \right\rangle. \numberthis \label{eq:two}
    \end{align*}
    We then need to show that both \eqref{eq:one} and \eqref{eq:two} are of $o_p(\sigma_1^2)$. By a similar arguments as in the proof of Lemma \ref{lemma 2}, we know that \eqref{eq:one} is $o_p(1)$. Meanwhile, we have
    \begin{align*}
        & \frac{1}{n}\left| \sum_{t=1}^n \left\langle M_{\hat{a}(X_t)} - M_{a^*(X_t)}, X_t \right\rangle  \left\langle  M_{a^*(X_t)}, X_t \right\rangle \right | \\
        \le & \sqrt{\frac{1}{n}\sum_{t=1}^n \left\langle M_{\hat{a}(X_t)} - M_{a^*(X_t)}, X_t \right\rangle^2 } \sqrt{\frac{1}{n}\sum_{t=1}^n \left\langle  M_{a^*(X_t)}, X_t \right\rangle^2},
    \end{align*} and
    \begin{equation*}
        \frac{1}{n}\sum_{t=1}^n \left\langle  M_{a^*(X_t)}, X_t \right\rangle^2 = \frac{1}{n}\sum_{t=1}^n\Big( I\{ a^*(X_t) = 1\}\left\langle  M_{1}, X_t \right\rangle^2 + I\{ a^*(X_t) = 0\}\left\langle  M_{0}, X_t \right\rangle^2\Big).
    \end{equation*}
    We also note that by the law of large numbers, there is
    \begin{equation*}
        \frac{1}{n} \sum_{t=1}^n \left\langle M_1, X_t\right\rangle^2 \xrightarrow{p} \bE_X \left[\left\langle M_1, X\right\rangle^2 \right].
    \end{equation*}
    Therefore, we can also see that \eqref{eq:two} is dominated by the order of $\frac{1}{n}\sum_{t=1}^n \left\langle M_{a^*(X_t)}, X_t \right\rangle^2$, and we thus finish the proof of 
    \begin{equation*}
        \frac{1}{n}\sum_{t=1}^n \left\langle M_{\hat{a}(X_t)}, X_t \right\rangle^2 \xrightarrow{p} \frac{1}{n}\sum_{t=1}^n \left\langle M_{a^*(X_t)}, X_t \right\rangle^2.
    \end{equation*}
    \item 
    Since $X_t$ are \emph{i.i.d.} distributed, by LLN,
    \begin{equation*}
        \frac{1}{n}\sum_{t=1}^n \left\langle M_{a^*(X_t)}, X_t \right\rangle^2 \xrightarrow{p} \bE [\langle M_{a^*(X)}, X\rangle^2].
    \end{equation*}
\end{enumerate}
Combining all the previous steps, we conclude the proof of \eqref{eq:var cons first part}. For \eqref{eq:var cons second part}, we first note that 
\begin{align*}
    & \frac{1}{n} \sum_{t=1}^n \left\langle \Msgd_{\hat{a}(X_t), t-1} - M_{\hat{a}(X_t)}, X_t \right\rangle \\
    = & \frac{1}{n} \sum_{t=1}^n I\{\hat{a}(X_t)=1\}\left\langle \Msgd_{1, t-1} - M_{1}, X_t \right\rangle + \frac{1}{n} \sum_{t=1}^n I\{\hat{a}(X_t)=0\}\left\langle \Msgd_{0, t-1} - M_{0}, X_t \right\rangle.
\end{align*}
We illustrate the bound for $a = 1$, while the analysis for $a=0$ is similar. Note that,
\begin{align*}
    & \left|\frac{1}{n} \sum_{t=1}^n I\{\hat{a}(X_t)=1\}\left\langle \Msgd_{1, t-1} - M_{1}, X_t \right\rangle\right|
    \le \frac{1}{n} \sum_{t=1}^n \left| \left\langle \Msgd_{1, t-1} - M_{1}, X_t \right\rangle \right|.
\end{align*}
Meanwhile
\begin{equation*}
     \frac{1}{n} \sum_{t=1}^n \left| \left\langle \Msgd_{1, t-1} - M_{1}, X_t \right\rangle \right| \xrightarrow{p} \frac{1}{n} \sum_{t=1}^n \bE\left[ \left| \left\langle \Msgd_{1, t-1} - M_{1}, X_t \right\rangle \right| \Big |\mathcal{F}_{t-1}\right],
\end{equation*}
and by Theorem \ref{thm: sgd consistent}, \begin{align*}
    \frac{1}{n} \sum_{t=1}^n \bE\left[ \left| \left\langle \Msgd_{1, t-1} - M_{1}, X_t \right\rangle \right| \Big |\mathcal{F}_{t-1}\right] \le \frac{C}{n} \sum_{t=1}^n \left\|  \Msgd_{1, t-1} - M_{1}\right\|_{\mathrm{F}} = o_p(\sigma_1).
\end{align*}
On the other hand, by similar arguments as in Lemma \ref{lemma 2}, we have 
\begin{equation*}
    \frac{1}{n}\sum_{t=1}^n \left\langle M_{\hat{a}(X_t), t-1} - M_{a^*(X_t)}, X_t\right\rangle = o_p(\sigma_1),
\end{equation*}
and thus by the independence of $X_t$ for all $t$, we have 
\begin{equation*}
    \frac{1}{n}\sum_{t=1}^n \left\langle M_{a^*(X_t)}, X_t\right\rangle \xrightarrow{p} \bE \left[\left\langle M_{a^*(X)}, X\right\rangle \right].
\end{equation*}
Therefore, combining all the relationships above, we have 
\begin{equation*}
    \frac{1}{n}\sum_{t=1}^n \left\langle \Msgd_{\hat{a}(X_t), t-1}, X_t \right\rangle \xrightarrow{p} \bE \left[ \langle M_{a^*(X)},X \rangle\right].
\end{equation*}
Finally, combining all the steps above, we conclude the proof of Theorem \ref{thm:consistent}.

\section{Supporting Technical Results}\label{appB}
    
    \subsection{Proof of Corollary \ref{co: the difference}}
    \label{sec: proof of corollary}
    We first realize that 
    \begin{equation*}
        \left (\widehat{m}^{(1)}_T - \widehat{m}^{(0)}_T \right) - \left (m^{(1)}_T - m^{(0)}_T \right) = \left (\widehat{m}^{(1)}_T - m^{(1)}_T \right) - \left ( \widehat{m}^{(0)}_T - m^{(0)}_T \right).
    \end{equation*}
    Recall the decomposition in the proof of Theorem \ref{thm1}, and we can apply the exact same decomposition for both $ (\widehat{m}^{(1)}_T - m^{(1)}_T )$ and $( \widehat{m}^{(0)}_T - m^{(0)}_T)$ as in Section \ref{sec: proof of thm1}. Therefore, the upper bound of all the negligible terms for both $i=0$ and $1$ follows the Lemmas \ref{lm: outter nelig}--\ref{lm : minor 2}. It remains to deal with the main term
    \begin{align*}
    &\sum_{i=1}^2 (-1)^{i+1} \left( \Big \langle \Uorg^{(i)} \Uorg^{(i)\top} \widehat{Z}^{(i)}_1V^{(i)}V^{(i)\top}, T\Big \rangle + \Big \langle U^{(i)}U^{(i)\top} \widehat{Z}^{(i)}_1 V^{(i)}_{\bot} V^{(i)\top}_{\bot}, T \Big \rangle \right)  \numberthis \label{eq: clt sum 1}\\ 
    + & \sum_{i=1}^2 (-1)^{i+1} \left(\left \langle \Uorg^{(i)}\Uorg^{(i)\top} \widehat{Z}^{(i)}_2V^{(i)}V^{(i)\top}, T\right \rangle + \left \langle U^{(i)}U^{(i)\top} \widehat{Z}^{(i)}_2 V^{(i)}_{\bot} V^{(i)\top}_{\bot}, T \right \rangle \right).\numberthis \label{eq: clt sum 2}
    \end{align*}
        
    First, we have
    \begin{align*}
    \widehat{Z}^{(1)}_1\widehat{Z}^{(0)\top}_1 = \frac{1}{n^2}\sum_{t_1 = 1}^n\sum_{t_2 = 1}^n \frac{I\{a_{t_1} = 1\}I\{a_{t_2} = 0\}}{p_{t_1}(1-p_{t_2})} \xi_{t_1}\xi_{t_2} X_{t_1}X_{t_2}^\top.
    \end{align*}
    When $t_1 = t_2$, we have $I\{a_{t_1} = 1\}I\{a_{t_2=0}\} = 0$. On the other hand, when $t_1 \ne t_2$,
   \begin{equation*}
       \mathbb{E}\left[\frac{1}{n^2}\sum_{t_1 = 1}^n\sum_{t_2 = 1}^n \frac{I\{a_{t_1} = 1\}I\{a_{t_2} = 0\}}{p_{t_1}(1-p_{t_2})} \xi_{t_1}\xi_{t_2} X_{t_1}X_{t_2}^\top \right] = 0.
   \end{equation*}
    It follows directly from the proof of Lemma \ref{lm:main term} that the asymptotic variance of \eqref{eq: clt sum 1} is given by $\sigma^2_1S_1^2 + \sigma^2_0S_0^2$. It remains to show that
    \begin{equation*}
        \frac{\sqrt{n}\sum_{i=1}^2 (-1)^{i+1} \left \langle \Uorg^{(i)}\Uorg^{(i)\top} \widehat{Z}^{(i)}_2V^{(i)}V^{(i)\top}+ U^{(i)}U^{(i)\top} \widehat{Z}^{(i)}_2 V^{(i)}_{\bot} V^{(i)\top}_{\bot}, T \right \rangle }{\sqrt{\sigma^2_1S_1^2 + \sigma^2_0S_0^2 }} \xrightarrow{p} 0,
    \end{equation*}
    which shares the same argument as the proof of Lemma \ref{lm:main term}, and we omit the details here.
    
    \subsection{Proof of Lemma \ref{lm: region D}}
    
    By Lemma C.4 in \cite{jin2016provable}, as long as $\left ( \Usgd, \Vsgd\right) \in D$ for $\Usgd$, $\Vsgd$ defined in Lemma \ref{lm: region D}, we have 
    \begin{equation}
    \label{eq: u v tilde op norm}
        \|\Usgd\| \le \sqrt{2\|M\|}, \hspace{3mm} \|\Usgd\| \le \sqrt{2\|M\|},
    \end{equation}
    and
    \begin{equation}
    \label{eq: min singular value}
        \lambda_{\min} \left(A^\top \Usgd \right) \ge \sqrt{\frac{\lambda_r}{2}}, \hspace{3mm} \lambda_{\min} \left(B^\top \Vsgd \right)\ge \sqrt{\frac{\lambda_r}{2}},
    \end{equation}
    where $A$ and $B$ are the top-$r$ singular vectors of $M$, and recall that $\lambda_r$ denotes the $r$-th singular vector of $M$. To prove the latter claim in Lemma \ref{lm: region D}, we see that because of equation \eqref{eq: u v tilde op norm}, we have
    \begin{equation*}
        \|\Usgd \Vsgd^\top\| = \|\Vsgd \Usgd^\top\| \le \|\Vsgd\| \|\Usgd\| \le 2\|M\|,
    \end{equation*}
    similarly, we have 
    \begin{equation*}
        \|\Vsgd\Vsgd^\top\| \le \|\Vsgd\| \|\Vsgd\| \le 2\|M\|, \|\Usgd\Usgd^\top\| \le \|\Usgd\| \|\Usgd\| \le 2\|M\|.
    \end{equation*}
    On the other hand, by equation \eqref{eq: min singular value}, and the proof of Lemma C.3 in \cite{jin2016provable}, we have
    \begin{equation*}
         \| (\Usgd\Vsgd^\top -M)\Vsgd\|_{\mathrm{F}}^2 + \| (\Usgd\Vsgd^\top -M)^\top\Usgd\|_{\mathrm{F}}^2 \ge \frac{\lambda_r}{2}.
    \end{equation*}

\subsection{Proof of Lemma \ref{lm: sum of prod}}
The proof follows a similar argument as Lemma F.4 in \cite{chen2022first}. 
We first note that 
\begin{equation*}
    \prod_{s = \tau+1}^t\left (1-\frac{\eta_s}{\kappa}\right) = \frac{\prod_{s = \tau}^t\left (1-\frac{\eta_s}{\kappa}\right)}{\left (1-\frac{\eta_\tau}{\kappa}\right)},
\end{equation*}
then we can see that for $\tau \ge \frac{c}{\kappa}\cdot 2^{\frac{1}{\alpha}}$, we have 
\begin{equation*}
    \frac{1}{\tau^\alpha} \le \frac{\kappa}{2c} \Leftrightarrow \eta_\tau \le \frac{\kappa}{2} \Leftrightarrow 1 - \frac{\eta_\tau}{\kappa} \ge \frac{1}{2} \Leftrightarrow \frac{1}{\left (1-\frac{\eta_\tau}{\kappa}\right)}\le 2.
\end{equation*}
Therefore, we have for $\tau \ge \frac{c}{\kappa}\cdot2^{\frac{1}{\alpha}}$,
\begin{equation*}
     \sum_{\tau=1}^{t}\eta_{\tau}^h \prod_{s=\tau+1}^t  \left (1-\frac{\eta_s}{\kappa}\right) \le 2\sum_{\tau=1}^{t} \eta_{\tau}^h \prod_{s=\tau}^t \left (1-\frac{\eta_s}{\kappa}\right).
\end{equation*}
We then note that for function $f(x) = (1 - cx^{-\alpha}/\kappa)^{x^\alpha \kappa/c}$ is monotonically increasing in $x$ and converges to $e^{-1}$. Therefore, we have
\begin{equation*}
    \sum_{\tau=1}^{t} \eta_{\tau}^h \prod_{s=\tau}^t \left (1-\frac{\eta_s}{\kappa}\right) \le \sum_{\tau = 1}^{t} \eta_\tau^h \exp \left (-\frac{1}{\kappa} \sum_{s =\tau}^t \eta_s\right),
\end{equation*}
then
\begin{align*}
\label{eq: first term in sum of prod}
    & \sum_{\tau = 1}^{t} \tau^\beta \eta_\tau^h \exp \left (-\frac{1}{\kappa} \sum_{s =\tau}^t \eta_s\right) \\
    = & \sum_{\tau = 1}^{t^\star} \tau^\beta\eta^h_{t^\star} \exp\left(-\frac{1}{\kappa} \sum_{s=\tau}^{t^\star} \eta_{t^\star} -\frac{c}{\kappa}\sum_{s = t^\star+1}^t s^{-\alpha} \right) + \sum_{\tau = t^\star + 1}^{t}\tau^\beta\eta_\tau^h\exp\left (-\frac{c}{\kappa}\sum_{s=\tau}^t s^{-\alpha}\right) \\
    \le & \sum_{\tau = 1}^{t^\star} \tau^\beta\eta^h_{t^\star} \exp\left(-\frac{1}{\kappa} \sum_{s=\tau}^{t^\star} \eta_{t^\star} -\frac{c}{\kappa}\int_{t^\star+1}^{t}x^{-\alpha}dx \right) + \sum_{\tau = t^\star + 1}^{t}\tau^\beta\eta_\tau^h\exp\left (-\frac{c}{\kappa}\sum_{s=\tau}^t s^{-\alpha}\right) \\
    \le  & \sum_{\tau = 1}^{t^\star} \tau^\beta\eta^h_{t^\star} \exp\left( -\frac{c}{\kappa (1-\alpha)}\left (t^{1-\alpha} - (t^{\star} + 1)^{ 1-\alpha }\right)\right) + \sum_{\tau = t^\star + 1}^{t}\tau^\beta\eta_\tau^h\exp\left (-\frac{c}{\kappa}\sum_{s=\tau}^t s^{-\alpha}\right) \\
    \le &(t^\star)^{1+\beta} \eta_{t^\star}^h \exp\left( -\frac{c}{\kappa (1-\alpha)}\left (t^{1-\alpha} - (t^\star+1)^{ 1-\alpha }\right)\right)+ \sum_{\tau = t^\star + 1}^{t}\tau^\beta\eta_\tau^h\exp\left (-\frac{c}{\kappa}\sum_{s=\tau}^t s^{-\alpha}\right). \numberthis
\end{align*}
Then we deal with the second term by realizing that 
\begin{align*}
\label{eq: second term in sum of prod}
     & \sum_{\tau = t^\star + 1}^{t}\tau^\beta\eta_\tau^h\exp\left (-\frac{c}{\kappa}\sum_{s=\tau}^t s^{-\alpha}\right) \\
     \le & c^h \sum_{\tau = t^\star + 1}^{t} \tau^{-h\alpha+\beta} \exp\left( -\frac{c}{\kappa}\int_{\tau}^t x^{-\alpha}dx\right) = c^h \sum_{\tau = t^\star + 1}^{t} \tau^{-h\alpha+\beta} \exp\left( -\frac{c}{\kappa}  \frac{t^{1-\alpha }- \tau^{1-\alpha}}{1-\alpha}\right) \\
     = & c^h\exp\left( -\frac{c}{\kappa} \frac{t^{1-\alpha}}{1-\alpha}\right) \sum_{\tau = t^\star+1}^{t}\tau^{-h\alpha+\beta}\exp\left( \frac{c}{\kappa}\frac{\tau^{1-\alpha}}{1-\alpha}\right) \\
     \le &  c^h\exp\left( -\frac{c}{\kappa} \frac{t^{1-\alpha}}{1-\alpha}\right) \int_{ t^\star+1}^{t}x^{-h\alpha+\beta}\exp\left( \frac{c}{\kappa}\frac{x^{1-\alpha}}{1-\alpha}\right) dx. \numberthis
\end{align*}
Note that for any $u \in [1,t]$, $\kappa>0$, and $\alpha <1$, using integration by parts we have
\begin{align*}
    &\int_{u}^{t}x^{-h\alpha + \beta}\exp\left( \frac{c}{\kappa}\frac{x^{1-\alpha}}{1-\alpha}\right) dx \\
    = & \frac{\kappa}{c} x^{-h\alpha + \alpha +\beta}\exp\left( \frac{c}{\kappa}\frac{x^{1-\alpha}}{1-\alpha}\right)\bigg |_{u}^t + \int_{u}^t \frac{\kappa (h-1)\alpha - \beta}{c} x^{-h\alpha +\beta + \alpha -1}\exp\left( \frac{c}{\kappa}\frac{x^{1-\alpha}}{1-\alpha}\right) dx\\
    \le & \frac{\kappa}{c} x^{-h\alpha+\alpha+\beta}\exp\left( \frac{c}{\kappa}\frac{x^{1-\alpha}}{1-\alpha}\right)\bigg |_{u}^t + u^{\alpha-1}\int_{u}^t \frac{\kappa (h-1)\alpha -\beta}{c} x^{-h\alpha +\beta}\exp\left( \frac{c}{\kappa}\frac{x^{1-\alpha}}{1-\alpha}\right)dx,
\end{align*}
therefore, using the fact that $u^{\alpha -1} \le 1$, we have
\begin{equation}
\label{eq: int fact}
    \int_{u}^{t}x^{-h\alpha + \beta}\exp\left( \frac{c}{\kappa}\frac{x^{1-\alpha}}{1-\alpha}\right) dx \le \frac{1}{c/\kappa - (h-1)\alpha - \beta/\kappa}x^{-(h-1)\alpha + \beta}\exp\left( \frac{c}{\kappa}\frac{x^{1-\alpha}}{1-\alpha}\right)\bigg |_{u}^t.
\end{equation}
Then together with equation \eqref{eq: second term in sum of prod} and equation \eqref{eq: int fact}, we have
\begin{align*}
     & \sum_{\tau = t^\star + 1}^{t}\tau^\beta \eta_\tau^h\exp\left (-\frac{c}{\kappa}\sum_{s=\tau}^t s^{-\alpha}\right)\\
    \le &  c^h\exp\left( -\frac{c}{\kappa} \frac{t^{1-\alpha}}{1-\alpha}\right) \cdot \frac{1}{c/\kappa -(h-1)\alpha  - \beta/\kappa}t^{-(h-1)\alpha +\beta}\exp\left( \frac{c}{\kappa}\frac{t^{1-\alpha}}{1-\alpha}\right)\\
    = & \frac{c^h}{c/\kappa - (h-1)\alpha - \beta/\kappa} t^{-(h-1)\alpha + \beta} \le c^{h-1}t^{-(h-1)\alpha + \beta},
\end{align*}
for large enough $c$ such that $c/\kappa >(h-1)\alpha + \beta/\kappa$. Finally, recall that $\kappa$ is a positive constant by assuming $M_i$ is well-conditioned matrix, then together with equation \eqref{eq: first term in sum of prod} we have 
\begin{align*}
    & \sum_{\tau = 1}^{t} \tau^\beta \eta_\tau^h \exp \left (-\frac{1}{\kappa} \sum_{s =\tau}^t \eta_s\right) \\
    \le & c^h(t^\star)^{1+\beta} \eta_{t^\star}^h \exp\left( -\frac{c}{\kappa (1-\alpha)}\left (t^{1-\alpha} - (t^\star+1)^{ 1-\alpha }\right)\right) + \sum_{\tau = t^\star + 1}^{t}\eta_\tau^h\exp\left (-\frac{c}{\kappa}\sum_{s=\tau}^t s^{-\alpha}\right) \\
    \le & c^h(t^\star)^{1+\beta} \eta_{t^\star}^h \exp\left( -\frac{c}{\kappa (1-\alpha)}\left (t^{1-\alpha} - (t^\star+1)^{ 1-\alpha }\right)\right) + c^{h-1}t^{-(h-1)\alpha +\beta}
    \leq \tilde{C}t^\beta\eta_t^{h-1},
\end{align*}
for an absolute constant $\tilde{C}$. We thus conclude the proof of Lemma \ref{lm: sum of prod}. 

\subsection{Proof of Lemma \ref{lm: max term}}

We first note that for any $x \in [0,1]$, we have $1-x \le \exp\left(-\frac{1}{2}x \right)$. This is because if we define $h(x) = 1 -x - \exp\left(-\frac{1}{2}x \right)$, then 
\begin{equation*}
    h'(x) = -1 + \frac{1}{2}\exp\left(-\frac{1}{2}x \right) < 0,
\end{equation*}
which indicates that $h(x)$ is decreasing function for $x \in [0,1]$. Then we have $h(x) \le h(0) = 0$, which implies  $1-x \le \exp\left(-\frac{1}{2}x \right)$. Therefore, we have
\begin{align*}
\label{eq:max lm eq}
    & \prod_{s=\tau+1}^t \left (1-\frac{\eta_s}{\kappa}\right)\tau^\beta\eta^h_{\tau}   \\
    \le & \prod_{s=\tau+1}^t \exp\left( -\frac{\eta_s\lambda_r}{2}\right)\tau^\beta\eta^h_{\tau} = \exp\left(-\frac{1}{2\kappa} \sum_{s = \tau+1}^t \eta_s\right) \tau^\beta \eta_\tau^h \\
    \le& c^h\exp\left(-\frac{c}{\kappa}\int_{\tau +1}^t x^{-\alpha} dx\right)\tau^{-h\alpha +\beta} = c^h \exp\left( -\frac{c}{2\kappa} \frac{t^{1-\alpha} - \left(\tau +1\right)^{1-\alpha}}{1-\alpha} \right)\tau^{-h\alpha+\beta}\\
    = & c^h \exp\left( -\frac{ct^{1-\alpha}}{2\kappa (1-\alpha)}\right)\exp\left( \frac{c(\tau+1)^{1-\alpha}}{2\kappa (1-\alpha)}\right) \tau^{-h\alpha+\beta}. \numberthis
\end{align*}
Then we define, for $x\ge 1$
\begin{equation*}
    f(x) = \exp\left( \frac{c(x+1)^{1-\alpha}}{2\kappa (1-\alpha)}\right) x^{-h\alpha+\beta},
\end{equation*}
then its derivative is given by
\begin{equation*}
    f'(x) = -(h\alpha-\beta) x^{-h\alpha +\beta -1} \exp\left( \frac{c(x+1)^{1-\alpha}}{2\kappa (1-\alpha)}\right) + x^{-h\alpha +\beta}  \exp\left( \frac{c(x+1)^{1-\alpha}}{2\kappa (1-\alpha)}\right)\frac{c(x+1)^{-\alpha}}{2\kappa}.
\end{equation*}
To prove the claim of Lemma \ref{lm: max term}, we only need to show that $f(\tau)$ is an increasing function, and the $\max$ can be reached at $\tau = t$. To see that, we only need to show
\begin{align*}
    & -(h\alpha-\beta) x^{-h\alpha +\beta -1} \exp\left( \frac{c(x+1)^{1-\alpha}}{2\kappa (1-\alpha)}\right) + x^{-h\alpha + \beta}  \exp\left( \frac{c(x+1)^{1-\alpha}}{2\kappa (1-\alpha)}\right)\frac{c(x+1)^{-\alpha}}{2\kappa} \ge 0 \\
    \Leftrightarrow & -(h\alpha-\beta) x^{-1} + \frac{c}{2\kappa}(x+1)^{-\alpha} \ge 0 \Leftrightarrow (x+1)^{-\alpha}x \ge \frac{2(h\alpha-\beta)\kappa}{c} \Leftrightarrow (x+1)^{1-\alpha} \ge \frac{4(h\alpha-\beta)\kappa}{c}. 
\end{align*}
Then we conclude that for $x \ge \left( \frac{4(h\alpha-
\beta)\kappa}{c}\right)^{\frac{1}{1-\alpha}}-1 $, $f(x)$ is an non-decreasing function. Therefore, $f(\tau) \le f(t)$ for any $t^\star \le\tau \le t$, and thus by recalling \eqref{eq:max lm eq}, we have
\begin{align*}
    \prod_{s=\tau+1}^t \left (1-\frac{\eta_s}{\kappa}\right)\tau^\beta\eta^2_{\tau} \le \exp\left( \frac{c\left((t+1)^{1-\alpha} - t^{1-\alpha} \right)}{2\kappa (1-\alpha)}\right) c^h t^{-h\alpha+\beta},
\end{align*}
where we can see that
\begin{equation*}
    (t+1)^{1-\alpha} - t^{1-\alpha} = (1-\alpha)\int_{t}^{t+1} x^{-\alpha} dx \le (1-\alpha)t^{-\alpha} \le (1-\alpha).
\end{equation*}
Then we have $
    \prod_{s=\tau+1}^t \left (1-\frac{\eta_s}{\kappa}\right)\tau^\beta\eta^h_{\tau} \le \exp \left( \frac{c}{2\kappa}\right)t^\beta\eta_t^h$, 
and thus conclude the proof. 

\subsection{Proof of Lemma \ref{lm:Opnorm}}
 \label{sec: proof of op norm lemma}
    By Assumption \ref{assum: noise}, we have $\mathbb{E}[\|X_t\|^2\leq d$ and  $\mathbb{E}[\xi_t^2|\mathcal{F}_{t-1}]\le\sigma^2$. Since $\pi_t$ is lower bounded by a constant, 
    \begin{equation*}
    \mathbb{E}[\|\widehat Z_1\|^2]=\frac{1}{n^2}\sum_{t=1}^n\frac{\sigma_1^2d}{p_0}\lesssim \frac{\sigma^2d}{n}.
    \end{equation*}
    Therefore by Markov inequality, we have $\|\widehat Z_1\|= O_p(\sigma\sqrt{d/n})$. For
    \begin{equation*}
        \widehat{Z}_2 = \frac{1}{n}\sum_{t=1}^n \left(\frac{I\{a_t = 1\} \langle \Delta_{t-1}, X_t \rangle X_t}{\pi_t} - \Delta_{t-1}\right),
    \end{equation*}
    By Assumption \ref{assum: noise}, we have $\mathbb{E}[\langle \Delta_{t-1}, X_t \rangle^2|\mathcal{F}_{t-1}]\leq \|\Delta_{t-1}\|_{\mathrm{F}}^4$ and $\mathbb{E}[\|X_t\|^2]\leq d^2$. Thus
    \begin{align*}
        \mathbb{E}\big[\|\widehat Z_2\|^2\big]\lesssim  \frac{d}{n^2\pmin} \sum_{t=1}^n \mathbb{E}\| \Delta_{t-1} \|^2_{\mathrm{F}}
    \end{align*}
    by Cauchy-Schwarz. Following the same argument in the proof of Theorem \ref{thm: sgd consistent}, 
    we have $\mathbb{E}\| \Delta_{t-1} \|^2_{\mathrm{F}}\leq \frac{dr(\log d)^2}{n^\alpha}$ and therefore we have the following bounds by Assumption \ref{assum: SNR condition},
    \[
        \|\widehat Z_2\|=O_p\left( \frac{\sigma d\sqrt{r}\log d}{\sqrt{n^{1+\alpha}}}\right),\quad \|\widehat Z\|\leq \|\widehat Z_1\|+\|\widehat Z_2\|=O_p\Big( \sigma\sqrt \frac dn\Big).
    \]
    Second, for fixed unit vectors \(u,v\), note that for each \(t\), $u^\top X_t v \sim \mathcal{N}(0,1)$, and is independent of \(\xi_t\) (and \(\mathcal{F}_{t-1}\)). Therefore, the \(t\)-th summand in $u^\top \widehat{Z}_1 v$ is mean zero and has conditional variance bounded by $\sigma^2/p_0$. Since the \(X_t\) and \(\xi_t\) are uncorrelated across \(t\), summing over \(n\) terms yields a variance of order \(\sigma^2/(n\,p_0^2)\). Hence, we have $u^\top \widehat{Z}_1v = O_p\!\Bigl(\frac{\sigma}{\sqrt{n}}\Bigr)$. 
    On the other hand, $\langle \Delta_{t-1}, X_t\rangle u^\top X_t v$ has conditional variance \(\|\Delta_{t-1}\|_{\mathrm{F}}^2\). 
Therefore,
\[
u^\top \widehat{Z}_2v = O_p\!\Bigl(\sigma\sqrt{\frac{d\,r\,(\log d)^2}{n^{1+\alpha}}}\Bigr).
\]

\subsection{Proof of Lemma \ref{lm:main term}}
\label{sec: proof of main term}
 We first divide the main term,
 \begin{equation*}
     \Big \langle \Uorg \Uorg^\top \widehat{Z}VV^\top, T\Big \rangle + \Big \langle UU^\top \widehat{Z} V_{\bot} V^\top_{\bot}, T \Big \rangle
 \end{equation*}
into two parts as follows,
\begin{equation}
\label{eq: main clt}
    \Big \langle \Uorg \Uorg^\top \widehat{Z}_1VV^\top, T\Big \rangle + \Big \langle UU^\top \widehat{Z}_1 V_{\bot} V^\top_{\bot}, T \Big \rangle,
\end{equation}
and
\begin{equation}
\label{eq: main reminder}
    \Big \langle \Uorg\Uorg^\top \widehat{Z}_2VV^\top, T\Big \rangle + \Big \langle UU^\top \widehat{Z}_2 V_{\bot} V^\top_{\bot}, T \Big \rangle.
\end{equation}
Note that
    \begin{align*}
        & \mathbf{Var}\Big ( \sqrt{n}\Big \langle \Uorg \Uorg^\top \widehat{Z}_1VV^\top, T\Big \rangle + \Big \langle UU^\top \widehat{Z}_1 V_{\bot} V^\top_{\bot}, T \Big \rangle \Big|\mathcal{F}_{t-1}\Big) \\
        = & \frac{1}{n} \sum_{t=1}^n \mathbb{E} \Big [ \xi_t^2 \frac{I\{ a_t = 1\}}{\pi^2_t} \Big( \Big \langle \Uorg\Uorg^\top \bX_tVV^\top, T\Big \rangle + \Big \langle UU^\top \bX_t V_{\bot} V^\top_{\bot}, T \Big \rangle \Big)^2 \Big | \mathcal{F}_{t-1} \Big ]\\
        = &\frac{\sigma^2}{n}\sum_{t=1}^n  \underbrace{\int \frac{\Big (\Big \langle \Uorg\Uorg^\top X VV^\top, T\Big \rangle + \Big \langle UU^\top X V_{\bot} V^\top_{\bot}, T \Big \rangle\Big)^2}{\pi_t(X)} dP_X}_{S^2_t},
    \end{align*}
    by recalling that $\pi_t(X) = \bP(a_t=1|\mathcal{F}_{t-1}, X_t = X)$. As $\pi_t \geq p_0$ and $\pi_t(X) \xrightarrow{p} \pi_\infty(X)$, we have  $
        S_t^2/S^2 \xrightarrow{p} 1$ 
    as $t \rightarrow \infty$, where
    \begin{equation*}
        S^2 = \int \frac{\Big (\Big \langle \Uorg\Uorg^\top \bX VV^\top, T\Big \rangle + \Big \langle UU^\top \bX V_{\bot} V^\top_{\bot}, T \Big \rangle \Big )^2}{\pi_\infty(X)} dP_X.
    \end{equation*}
    Therefore, by the martingale central limit theorem, we thus have
    \begin{equation}
    \label{eq: main term clt}
            \frac{\sqrt{n} \Big (\Big \langle \Uorg\Uorg^\top \widehat{Z}_1VV^\top, T\Big \rangle + \Big \langle UU^\top \widehat{Z}_1 V_{\bot} V^\top_{\bot}, T \Big \rangle \Big )}{\sigma S} \xrightarrow{d} \mathcal{N}( 0, 1).
    \end{equation}
    Next, we evaluate \eqref{eq: main reminder}. By the definition of $\widehat{Z}_2$, we have 
    \begin{align*}
        &\Big \langle \Uorg\Uorg^\top \widehat{Z}_2VV^\top, T\Big \rangle=  \frac{1}{n}\sum_{t=1}^n \frac{I\{a_t=1\}}{\pi_t} \Big ( \langle \Delta_{t-1}, \bX_t \rangle \Big \langle \Uorg\Uorg^\top \bX_t VV^\top, T\Big \rangle - \Big \langle \Uorg\Uorg^\top \Delta_{t-1} VV^\top, T\Big \rangle \Big ).
    \end{align*}
    Note that conditional on $\mathcal{F}_{t-1}$,
    \begin{align*}
        &\mathbb{E}\Big[\frac{I\{a_t=1\}}{\pi^2_t} \langle \Delta_{t-1}, \bX_t \rangle^2\Big \langle \Uorg\Uorg^\top \bX_t VV^\top, T\Big \rangle^2  \Big| \mathcal{F}_{t-1}\Big]\lesssim  \frac{2}{\pmin} \|\Delta_{t-1}\|^2_{\mathrm{F}} \|V^\top T^\top \Uorg \|^2_{\mathrm{F}}.
    \end{align*}
    Note that $\pmin$ is a constant, and 
    \begin{align*}
        \vartheta^2_n :=& \sum_{t=1}^n \mathbb{E}\Big[\frac{I\{a_t=1\}}{\pi^2_t} \langle \Delta_{t-1}, \bX_t \rangle^2\Big \langle \Uorg\Uorg^\top \bX_t VV^\top, T\Big \rangle^2  \Big| \mathcal{F}_{t-1}\Big]\\
        \lesssim &\frac{2}{\pmin} \|V^\top T^\top \Uorg \|^2_{\mathrm{F}} \sum_{t=1}^n\|\Delta_{t-1}\|^2_{\mathrm{F}}.
    \end{align*}Therefore, following the same argument in the proof of Theorem \ref{thm: sgd consistent}, we have
    \begin{equation*}
        \Big \langle UU^\top \widehat{Z}_2 V_{\bot} V^\top_{\bot}, T \Big \rangle =O_p\Big(\sigma \| U^\top T \Vorg \|_{\mathrm{F}} \sqrt{\frac{dr\log^2d}{n^{1+\alpha}}}\Big).
    \end{equation*}
    
    Recall the definition of $S^2$, we can see that the lower bound for $S^2$ is given by 
    \begin{equation*}
        S^2 \ge  \mathbb{E}\left [\Big (\Big \langle \Uorg\Uorg^\top \bX VV^\top, T\Big \rangle + \Big \langle UU^\top \bX V_{\bot} V^\top_{\bot}, T \Big \rangle \Big )^2\right] = \| V^\top T^\top \Uorg \|^2_{\mathrm{F}}+\| U^\top T \Vorg \|^2_{\mathrm{F}},
    \end{equation*}
    and by Assumption \ref{assum: SNR condition}, 
    \begin{equation*}
        \frac{\sqrt{n} \left( \Big \langle \Uorg\Uorg^\top \widehat{Z}_2VV^\top, T\Big \rangle + \Big \langle UU^\top \widehat{Z}_2 V_{\bot} V^\top_{\bot}, T \Big \rangle \right)}{\sigma S} \xrightarrow{p} 0.
    \end{equation*}
    Together with equation \eqref{eq: main term clt}, we conclude the proof of the Lemma \ref{lm:main term}.

    \subsection{Proof of Lemma \ref{lm: outter nelig}}
    First, recall that $\Uhat$ and $\Vhat$ are the left and right top-$r$ singular vectors of $\Munbs_n$. We have $\| \Uhat\| = \|\Vhat \| = 1$, and thus 
    \begin{align*}
        &\Big \vert \left \langle \Uhat\Uhat^\top \widehat{Z} \Vhat\Vhat^\top, T \right \rangle \Big \vert  \\
        = &  \Big \vert\left \langle ( \Uhat\Uhat^\top - UU^\top) \widehat{Z} V , TV\right \rangle +  \left \langle ( \Uhat\Uhat^\top - UU^\top) \widehat{Z} (\Vhat\Vhat^\top - VV^\top) , T\right \rangle  \\
        & +  \left \langle U^\top \widehat{Z}  (\Vhat\Vhat^\top - VV^\top ), U^\top T \right \rangle + \left \langle UU^\top \widehat{Z} VV^\top, T \right \rangle\Big \vert \\
        \le & \left \| TV \right \|_{\mathrm{F}} \sqrt{r} \|\widehat{Z}\| \left \|\Uhat\Uhat^\top - UU^\top \right\| + \| U^\top T  \|_{\mathrm{F}} \sqrt{r} \|\widehat{Z}\| \left \|\Vhat\Vhat^\top - VV^\top \right\| \\
        & + \sqrt{r}\|T\|_{\mathrm{F}}  \|\widehat{Z}\| \left \|\Uhat\Uhat^\top - UU^\top \right\|  \left \|\Vhat\Vhat^\top - VV^\top \right\| + \left \vert \left \langle UU^\top \widehat{Z} VV^\top, T \right \rangle \right \vert.
    \end{align*}
    
    According to \cite{wedin1972perturbation}'s $\mathrm{sin}\Theta$ theorem, we have 
    \begin{equation}
    \label{eq: UV dist}
        \max \Big \{ \left \| \Uhat\Uhat^\top - UU^\top \right\|, \left \| \Vhat\Vhat^\top - VV^\top \right\|\Big \} \le \frac{\sqrt{2}\| \widehat{Z}\|}{\lambda_r},
    \end{equation}
    and thus according to Lemma \ref{lm:Opnorm}, we have 
    \begin{equation}
    \label{eq: op dist in neg 1}
        \| \widehat{Z} \| \left \| \Uhat\Uhat^\top - UU^\top \right\| =O_p\Big( \frac{1 }{\lambda_r} \| \widehat{Z} \|^2\Big).
    \end{equation}
    Therefore,
    \[
    \left \| TV \right \|_{\mathrm{F}} \sqrt{r} \|\widehat{Z}\| \left \|\Uhat\Uhat^\top - UU^\top \right\|=O_p\Big(\left \| TV \right \|_{\mathrm{F}} \frac{\sigma^2 }{\lambda_r} \frac{d\sqrt r}{n}\Big).
    \]
    A similar bound applies to $ \| U^\top T  \|_{\mathrm{F}} \sqrt{r} \|\widehat{Z}\|  \|\Vhat\Vhat^\top - VV^\top \|$. 
    In addition, we have
    \begin{equation*}
        \sqrt r\|T\|_{\mathrm{F}}  \|\widehat{Z}\| \left \|\Uhat\Uhat^\top - UU^\top \right\|  \left \|\Vhat\Vhat^\top - VV^\top \right\| =O_p\Big( \frac{\sqrt r }{\lambda_r^2} \|T \|_{\mathrm{F}}\| \widehat{Z} \|^3\Big).
    \end{equation*}
    By Assumption \ref{assum: null space}, we have 
    \[
     \frac{\sqrt r }{\lambda_r^2} \|T \|_{\mathrm{F}}\| \widehat{Z} \|^3=O_p\left(\Big(\left \| TV \right \|_{\mathrm{F}}  + \| U^\top T  \|_{\mathrm{F}}\Big)\frac{\sigma^3}{\lambda_r^2 }\frac{d^2}{\sqrt{n^3}}\right).
     \] 
    Note that $\widehat Z=\widehat Z_1+\widehat Z_2$. By Lemma \ref{lm:Opnorm},
      \begin{align*}
    &\left \langle UU^\top \widehat{Z}_2 VV^\top, T \right \rangle = O_p(\sqrt r\|\widehat Z_2\|\|U^\top TV\|_{\mathrm{F}})=O_p\left( \| U^\top TV  \|_{\mathrm{F}}\frac{\sigma dr\log d}{\sqrt{n^{1+\alpha}}}\right).
      \end{align*}
    Recall that $\widehat Z_1=\frac{1}{n} \sum^n_{t=1} I\{a_t = 1\} \xi_t \bX_t / \pi_t$. By Assumption \ref{assum: noise},
    \[
    \left \langle UU^\top \widehat{Z}_1 VV^\top, T \right \rangle= \left\langle \widehat{Z}_1, UU^\top TVV^\top \right \rangle =O_p\left(\frac{\sigma\|UU^\top TVV^\top\|_{\mathrm{F}}}{\sqrt{n}}\right).
    \]
     Combining above, 
    \begin{align*}
      \langle \Uhat\Uhat^\top \widehat{Z}& \Vhat\Vhat^\top, T  \rangle=O_p\left( \| U^\top TV  \|_{\mathrm{F}}\Big(\frac{\sigma}{\sqrt{n}}+\frac{\sigma d\sqrt{r}\log d}{\sqrt{n^{1+\alpha}}}\Big)\right) \\
      &+O_p\left(\big(\left \| TV \right \|_{\mathrm{F}}  + \| U^\top T  \|_{\mathrm{F}}\big)\Big(\frac{\sigma^2 d\sqrt r}{\lambda_r n} +\frac{\sigma^3}{\lambda_r^2 }\frac{d^2}{\sqrt{n^3}}\Big)\right).
     \end{align*}
     Note that $\|U^\top TV\|_{\mathrm{F}}/\|TV\|_{\mathrm{F}}\rightarrow 0$ from \eqref{eq: ratio lim}. By Assumption \ref{assum: SNR condition},  we thus conclude the proof for Lemma \ref{lm: outter nelig}.
     
    \subsection{Proof of Lemma \ref{lm: minor 1}}
        We first restate an observation in \cite{xia2021normal}.
    \begin{lemma}[\citealp{xia2021normal}]
    \label{lm:OpDist}
        Under Assumption of Theorem \ref{thm1}, for any $\ell\geq 1$, we have 
        \begin{equation*}
             \Big\| \sum_{k\ge \ell}^{\infty} \mathcal{S}_{A,k} (\widehat{E}) 
            \Big\| \lesssim  \Big (\frac{\|\widehat{E}\|}{\lambda_r}\Big)^{\ell}.
        \end{equation*}
    \end{lemma}
    By Lemma \ref{lm:Opnorm}, we have
    \begin{align*}
         \Big \langle  \sum_{k\ge 2}^{\infty} \mathcal{S}_{A,k}A \mathbf{\Theta}\mathbf{\Theta}^\top + \mathbf{\Theta}\mathbf{\Theta}^\top A \sum_{k\ge 2}^{\infty} \mathcal{S}_{A,k}, \Tilde{T} \Big\rangle  =O_p\left( \left( \|U^\top T\|_{\mathrm{F}} +\|TV\|_{\mathrm{F}} \right) \frac{\sigma^2 d\sqrt r}{\lambda_r^2 n}\right).
    \end{align*}
    
    \subsection{Proof of Lemma \ref{lm : minor 2}}
    Recall that $ \widehat{\mathbf{\Theta}}\widehat{\mathbf{\Theta}}^\top - \mathbf{\Theta}\mathbf{\Theta}^\top = \mathcal{S}_{A,1}(\widehat{E}) + \sum_{k\ge 2}^{\infty} \mathcal{S}_{A,k} (\widehat{E})$, 
    and $\mathcal{S}_{A,1}(\widehat{E}) = \mathfrak{B}^{-1} \widehat{E} \mathfrak{B}^{\perp} + \mathfrak{B}^{\perp} \widehat{E} \mathfrak{B}^{-1}$. 
    We can write 
    \begin{align*}
    &\left \langle (\widehat{\mathbf{\Theta}}\widehat{\mathbf{\Theta}}^\top - \mathbf{\Theta}\mathbf{\Theta}^\top) A (\widehat{\mathbf{\Theta}}\widehat{\mathbf{\Theta}}^\top - \mathbf{\Theta}\mathbf{\Theta}^\top), \Tilde{T} \right \rangle \\
        = &\underbrace{\left \langle \mathcal{S}_{A,1} A \mathcal{S}_{A,1}, \tilde{T}\right\rangle}_{I} + \underbrace{ \Big \langle \mathcal{S}_{A,1} A \mathcal{S}_{A,2}+ \mathcal{S}_{A,2} A \mathcal{S}_{A,1}, \tilde{T}\Big\rangle}_{II} \\
        &+ \underbrace{ \Big \langle  \sum_{k\ge 3}^{\infty} \mathcal{S}_{A,k}  A \mathcal{S}_{A,1}+\mathcal{S}_{A,1} A \sum_{k\ge 3}^{\infty} \mathcal{S}_{A,k}, \Tilde{T}\Big \rangle}_{III} +\underbrace{ \Big \langle  \sum_{k\ge 2}^{\infty} \mathcal{S}_{A,k}  A \sum_{k\ge 2}^{\infty} \mathcal{S}_{A,k}, \Tilde{T}\Big \rangle}_{IV}.
    \end{align*}
    For the term $(I)$. We have
    \[
    I=\big\langle \mathfrak{B}^{\perp} \widehat{E} \mathfrak{B}^{-1}A\mathfrak{B}^{-1} \widehat{E} \mathfrak{B}^{\perp},~\tilde T\rangle
    =\left \langle U_\perp U_\perp^\top \widehat{Z}V\Lambda^{-1}U^\top \widehat{Z} V_\perp V_\perp^\top, T\right \rangle.
    \]
    Assume that $U_\perp U_\perp^\top TV_\perp V_\perp^\top $ has the following SVD, $U_\perp U_\perp^\top TV_\perp V_\perp^\top=\sum_{k=1}^{r'} s_k\widetilde u_k\widetilde v_k^\top$, where $r'\leq d-r$, and $\sum_{k=1}^{r'} s_k^2=\|U_\perp^\top TV_\perp\|_{\mathrm{F}}^2$. By Cauchy-Schwarz inequality,
    \begin{align*}
    \sum_{k=1}^{r'}s_k\leq \sqrt{r'}\|U_\perp^\top TV_\perp\|_{\mathrm{F}}\leq \sqrt{d}\|U_\perp^\top TV_\perp\|_{\mathrm{F}}.
    \end{align*} 
    Let $\{u_\ell,v_\ell\}$ be the singular vectors corresponding to $U,V$. We can rewrite
    \begin{align}\label{eq:termI}
    I=&\Big\langle \widehat Z V\Lambda^{-1}U \widehat Z,~\sum_{k=1}^{r'}s_k\widetilde u_k\widetilde v_k^\top\Big\rangle
    =\sum_{k=1}^{r'}\sum_{\ell=1}^{r} s_k\widetilde u^\top_k \widehat Z v_\ell\lambda_\ell^{-1}u_\ell \widehat Z\widetilde v_k.
    \end{align}
    By Lemma \ref{lm:Opnorm} and Cauchy-Schwarz, we have 
    \begin{align}\label{eq:termI2}
    I\lesssim \sum_{k=1}^{r'}\sum_{\ell=1}^r \frac{s_k\sigma^2}{n\lambda_\ell}\lesssim \frac{r \sigma^2}{n}\sqrt{r'\sum_{k=1}^{r'} s_k^2}\le \frac{\sigma^2}{\lambda_r}\sqrt{\frac{dr^2}{n^{2}}}\|U_\perp^\top TV_\perp\|_{\mathrm{F}}.
    \end{align}
    According to Assumption \ref{assum: null space},
    \begin{align*}
    I=&O_p\Big( (\|U^\top T\|_{\mathrm{F}}+\|TV\|_{\mathrm{F}})\frac{\sigma^2}{\lambda_r}\sqrt{\frac{d^2r}{n^{2}}}\Big).
    \end{align*}
    
    Using similar arguments, with Lemma \ref{lm:OpDist},
    \[
     II+III+IV  = O_p\Big(\|U^\top T \|_\mathrm{F}+\|TV\|_{\mathrm{F}}) \frac{\sigma^3}{\lambda_r^2} \sqrt{\frac{d^4r}{n^3}}\Big). 
    \]
    Combining all the terms above, with Assumption \ref{assum: SNR condition}, we conclude the proof. 

\subsection{Proof of Lemma \ref{Lemma 1}}
Define
\[
A := M_1 - M_0,
\quad
B := \bigl(M_1 - \Msgd_{1,t-1}\bigr)\;+\;\bigl(\Msgd_{0,t-1} - M_0\bigr)
\,=\,
(M_1 - M_0)\;-\;\bigl(\Msgd_{1,t-1} - \Msgd_{0,t-1}\bigr).
\]
Note that $X_t$ is independent of $\mathcal{F}_{t-1}$, and define
\[
\Delta_{X_t} := \langle A,\;X_t\rangle 
\quad\text{and}\quad
\widehat{\Delta}_{X_t} := \langle \Msgd_{1,t-1} - \Msgd_{0,t-1},\;X_t\rangle \;=\;\langle A-B,\;X_t\rangle.
\]
Notice that 
\begin{equation*}
    I\{\widehat{\Delta}_{X_t} > 0\} = I\{\widehat{\Delta}_{X_t} +  \Delta_{X_t} - \Delta_{X_t}> 0 \} = I\{\Delta_{X_t} > \Delta_{X_t} - \widehat{\Delta}_{X_t} \}.
\end{equation*}
Therefore, we have  $
    I\{\widehat{\Delta}_{X_t} > 0\} = I\{\Delta_{X_t}>0\}$,
if and only if
\begin{equation}
\label{eq:lm1 (2)}
    \vert \Delta_{X_t}\vert > \vert \Delta_{X_t} - \widehat{\Delta}_{X_t}\vert.
\end{equation}
Therefore, we can rewrite the target probability as
\begin{align*}
    \bP \left( \hat{a}(X_t) = a^*(X_t)| \mathcal{F}_{t-1}\right) &= \bE\left[I\{\hat{a}(X_t) = a^*(X_t)\}|\mathcal{F}_{t-1}\right] \\
    &= \bE\left[I\left\{ I\{\widehat{\Delta}_{X_t} > 0\} = I\{\Delta_{X_t}>0\}\right\}|\mathcal{F}_{t-1}\right]\\
    & = \bE\left[I\left\{ \vert \Delta_{X_t}\vert > \vert \Delta_{X_t} - \widehat{\Delta}_{X_t}\vert \right\}|\mathcal{F}_{t-1}\right] \\
    & = \bP\left( \vert \langle A,X_t\rangle\vert > \vert \langle B,X_t\rangle\vert \big|\mathcal{F}_{t-1}\right).
\end{align*}
Given the above relationship, we focus on studying $\bP\left( \vert \langle A,X_t\rangle\vert > \vert \langle B,X_t\rangle\vert \big|\mathcal{F}_{t-1}\right)$. 
If we denote matrix $A = M_1 - M_0$, and matrix $B= M_1 - \Msgd_{1,t-1} + \Msgd_{0,t-1} - M_0$, and denote the Gaussian random variable $w_1 = \Delta_{X_t}$ while $w_2 = \Delta_{X_t} - \widehat{\Delta}_{X_t}$, then conditional on $\mathcal{F}_{t-1}$, we have $(w_1, w_2)$ is a joint Gaussian r.v. as 
\begin{equation*}
\begin{pmatrix}
    w_1 \\
    w_2
\end{pmatrix} \sim \mathcal{N}\left( 
\begin{pmatrix}
    0 \\
    0
\end{pmatrix} ,
    \begin{pmatrix}
\|A\|^2_{\mathrm{F}} & \langle A,B\rangle \\
\langle A,B\rangle & \|B\|^2_{\mathrm{F}}
\end{pmatrix}
\right),
\end{equation*}
It is easy to see that
\[
\bP\big(|w_1|\leq |w_2|\big|\mathcal{F}_{t-1}\big)\leq C_1 \|B\|_{\mathrm{F}} /\|A\|_{\mathrm{F}}.
\]
Then we have
\begin{equation*}
        \bP(\hat{a}(X_t) \ne a^*(X_t)|\mathcal{F}_{t-1}) \le C_1\|B\|_{\mathrm{F}} /\|A\|_{\mathrm{F}}\le C_1 \frac{\sum_{i=0}^1\| \Msgd_{i,t-1} - M_i\|_{\mathrm{F}}}{\Df}.
    \end{equation*}

\subsection{Proof of Lemma \ref{lemma 2}}

We first notice that 
\begin{equation*}
    \frac{1}{\sqrt{n}} \sum_{t=1}^n \left \vert \left\langle  M_{\hat{a}(X_t)} - M_{a^*(X_t)}, X_t\right\rangle \right\vert = \frac{1}{\sqrt{n}} \sum_{t=1}^n I\{ \hat{a}(X_t) \ne a^*(X_t)\} \left \vert \left \langle M_1- M_0, X_t\right\rangle\right\vert,
\end{equation*}
which is due to the fact that any item in the summation is not zero if and only if $\hat{a}(X_t) \ne a^*(X_t)$. Recall that in the proof of Lemma \ref{Lemma 1}, we have shown that 
\begin{equation*}
    I\{\hat{a}(X_t) \ne a^*(X_t) \} = I\left\{ I\{\widehat{\Delta}_{X_t} > 0\} \ne I\{\Delta_{X_t}>0\}\right\} = I\left\{ \vert \Delta_{X_t}\vert \le \vert \Delta_{X_t} - \widehat{\Delta}_{X_t}\vert \right\}.
\end{equation*}
Therefore, we have
\begin{align*}
\label{eq:lm2 (1)}
     \frac{1}{\sqrt{n}} \sum_{t=1}^n \left \vert \left\langle  M_{\hat{a}(X_t)} - M_{a^*(X_t)}, X_t\right\rangle \right\vert 
    =  &\frac{1}{\sqrt{n}} \sum_{t=1}^n  I\left\{ \vert \Delta_{X_t}\vert \le \vert \Delta_{X_t} - \widehat{\Delta}_{X_t}\vert \right\} \left \vert  \Delta_{X_t}\right\vert \\
    \le & \frac{1}{\sqrt{n}} \sum_{t=1}^n  I\left\{ \vert \Delta_{X_t}\vert \le \vert \Delta_{X_t} - \widehat{\Delta}_{X_t}\vert \right\} \vert \Delta_{X_t} - \widehat{\Delta}_{X_t}\vert. \numberthis
\end{align*}
 In addition, we note that it is easy to see that $\bE[\vert \Delta_{X_1} - \widehat{\Delta}_{X_1}\vert ] < \infty$ and that $$\bP\left(I\left\{ \vert \Delta_{X_t}\vert \le \vert \Delta_{X_t} - \widehat{\Delta}_{X_t}\vert \right\} \vert \Delta_{X_t} - \widehat{\Delta}_{X_t}\vert > x \right) \le \bP\left(\vert \Delta_{X_1} - \widehat{\Delta}_{X_1}\vert > x \right) $$ for any $x$. Therefore, we can apply Theorem 2.19 in \cite{hall2014martingale}, and have 
\begin{align*}
\label{eq:lm2 (2)}
    & \frac{1}{\sqrt{n}} \sum_{t=1}^n  I\left\{ \vert \Delta_{X_t}\vert \le \vert \Delta_{X_t} - \widehat{\Delta}_{X_t}\vert \right\} \vert \Delta_{X_t} - \widehat{\Delta}_{X_t}\vert \\
    \xrightarrow{p} & \frac{1}{\sqrt{n}} \sum_{t=1}^n \bE \left[I\left\{ \vert \Delta_{X_t}\vert \le \vert \Delta_{X_t} - \widehat{\Delta}_{X_t}\vert \right\} \vert \Delta_{X_t} - \widehat{\Delta}_{X_t}\vert \big | \mathcal{F}_{t-1} \right]. \numberthis
\end{align*}
We first note the fact that conditional on $\mathcal{F}_{t-1}$, both $\Delta_{X_t}$ and $\Delta_{X_t} - \widehat{\Delta}_{X_t}$ are Gaussian random variable. If we denote matrix $A = M_1 - M_0$, and matrix $B_{t-1} = M_1 - \Msgd_{1,t-1} + \Msgd_{0,t-1} - M_0$, and denote the Gaussian random variable $w_1 = \Delta_{X_t}$ while $w_2 = \Delta_{X_t} - \widehat{\Delta}_{X_t}$, then we have $(w_1, w_2)$ is a joint Gaussian r.v. as 
\begin{equation*}
\begin{pmatrix}
    w_1 \\
    w_2
\end{pmatrix} \sim \mathcal{N}\left( 
\begin{pmatrix}
    0 \\
    0
\end{pmatrix} ,
    \begin{pmatrix}
\|A\|^2_{\mathrm{F}} & \langle A,B_{t-1}\rangle \\
\langle A,B_{t-1}\rangle & \|B_{t-1}\|^2_{\mathrm{F}}
\end{pmatrix}
\right),
\end{equation*}
and we then know that 
\begin{equation*}
    w_1|w_2 \sim \mathcal{N}\left(w_2\frac{\langle A,B \rangle}{\|B\|^2_{\mathrm{F}}}, \|A\|^2_{\mathrm{F}}  - \frac{\langle A,B\rangle^2}{\|B\|^2_{\mathrm{F}}}\right),
\end{equation*}
where we use $B$ as the short notation for $B_{t-1}$. It is easy to see that
\begin{equation*}
     \mathbb{E}_{X_t}\left[I\{|w_1| \le |w_2|\} |w_2|\right] \le C_2' \frac{4\|B\|^2_{\mathrm{F}}}{\|A\|_{\mathrm{F}}},
\end{equation*}
for some positive constant $C_2'$. Then recall that $\|B\|_{\mathrm{F}} \le \|\Msgd_{1,t} - M_1\|_{\mathrm{F}} + \|\Msgd_{0,t} - M_0\|_{\mathrm{F}}$, and for some positive constant $C' = \max\{C_1', C_2'\}$, for $t>t_1$, we have 
\begin{align*}
    & \frac{1}{\sqrt{n}} \sum_{t=t_1 + 1}^n \bE \left[I\left\{ \vert \Delta_{X_t}\vert \le \vert \Delta_{X_t} - \widehat{\Delta}_{X_t}\vert \right\} \vert \Delta_{X_t} - \widehat{\Delta}_{X_t}\vert \big | \mathcal{F}_{t-1} \right] \\
    \le & \frac{1}{\sqrt{n}}\sum_{t=t_1 + 1}^n \frac{8C'\left(\|\Msgd_{1,t} - M_1\|^2_{\mathrm{F}} + \|\Msgd_{0,t} - M_0\|^2_{\mathrm{F}}\right)}{\Df}.
\end{align*}
Then by the results of Theorem \ref{thm: sgd consistent},  we have with probability $1 - \frac{4n}{d^\gamma}$,
\begin{equation*}
    \frac{1}{\sqrt{n}}\sum_{t=t_1 + 1}^n \frac{8C'\left(\|\Msgd_{1,t} - M_1\|^2_{\mathrm{F}} + \|\Msgd_{0,t} - M_0\|^2_{\mathrm{F}}\right)}{\Df} \le C\sigma_1 \sqrt{n}\frac{\sigma_1}{\Df} \frac{\gamma^2 dr\log^2(d)}{n^{\alpha - \beta}},
\end{equation*}
for some positive constant $C$. Then by Assumption \ref{assum:Optimal-Gap}, we first have 
\begin{equation*}
    \frac{1}{\sqrt{n}} \sum_{t=t_1 + 1}^n \bE \left[I\left\{ \vert \Delta_{X_t}\vert \le \vert \Delta_{X_t} - \widehat{\Delta}_{X_t}\vert \right\} \vert \Delta_{X_t} - \widehat{\Delta}_{X_t}\vert \big | \mathcal{F}_{t-1} \right] = o_p(\sigma_1),
\end{equation*}
for both case 1 and case 2. On the other hand, for the part that $t \le t_1$, 
\begin{equation}
\label{eq:zero to t1}
    \frac{1}{\sqrt{n}}\sum_{t=1}^{t_1} \mathbb{E}_{X_t}\left[I\{|w_1| \le |w_2|\} |w_2|\right] \le \frac{1}{\sqrt{n}}\tilde{C}t_1 \sigma_1,
\end{equation}
then as $n \rightarrow \infty$, we can easily see that the above term is $o_p(\sigma_1)$. Then if we combine above with \eqref{eq:lm2 (2)} and \eqref{eq:zero to t1}, we finally conclude that 
\begin{equation*}
    \frac{1}{\sqrt{n}} \sum_{t=1}^n  I\left\{ \vert \Delta_{X_t}\vert \le \vert \Delta_{X_t} - \widehat{\Delta}_{X_t}\vert \right\} \vert \Delta_{X_t} - \widehat{\Delta}_{X_t}\vert = o_p(\sigma_1).
\end{equation*}

\subsection{Proof of Lemma \ref{lemma 3}}

We note that 
\begin{equation*}
    \left| I\{\widehat{\Delta}_{X_t} > 0\} - I\{ \Delta_{X_t} >0\}\right| = I\{\hat{a}(X_t) \ne a^*(X_t) \}.
\end{equation*}
By Theorem 2.19 in \cite{hall2014martingale}, Lemma \ref{Lemma 1}, Assumption \ref{assum:Optimal-Gap}, and \eqref{lm3: eq1}, we have
\begin{align}
\label{lm3: eq1}
    \frac{1}{n}\sum_{t=1}^n\left| I\{\widehat{\Delta}_{X_t} > 0\} - I\{ \Delta_{X_t} >0\}\right| =\frac{1}{n}\sum_{t =1}^n I\{\hat{a}(X_t) \ne a^*(X_t) \} =o_p(1).
\end{align}

\subsection{Discussion on the Incoherence and SNR Conditions for Parameter Inference}
\setlength{\baselineskip}{0.95\baselineskip} 

We first note that the incoherence condition of Assumption \ref{assum: inco_assum} is not strictly necessary for establishing asymptotic normality; rather, it serves to simplify the expression of the asymptotic distribution. In our analysis, the sole instance in which this assumption is invoked for parameter inference is in \eqref{eq: ratio lim}. There, Assumption \ref{assum: inco_assum} is used to show that $\| \bV^\top T^\top \bU\|^2_{\mathrm{F}} + \| \bU^\top T \bV \|^2_{\mathrm{F}}$ is bounded by $\| TV\|^2_{\mathrm{F}} + \| U^\top T\|^2_{\mathrm{F}}$, which is a key step in the subsequent proof of Lemma \ref{lm: outter nelig} to establish that $ \langle UU^\top \widehat{Z}_1 VV^\top, T  \rangle$ is negligible. Absent the incoherence condition, this term will contribute an additional leading-order component in the asymptotic distribution—specifically, at the scale of $\langle UU^\top X VV^\top, T\rangle$. A comprehensive treatment of further relaxing this assumption is deferred to future study.

We next discuss how Assumption~\ref{assum: SNR condition} on the signal-to-noise ratio (SNR) may be relaxed by imposing an additional low-rank condition on the matrix $T$, which specifies the linear form under inference. In particular, if $\mathrm{rank}(T) = r_T$ is a constant, one could potentially weaken the SNR requirement with a more careful analysis. Here, we offer some preliminary insights into this direction, leaving a complete and rigorous derivation to future work. Specifically, one would need to refine the bounds for 
$
\big\langle \mathcal{S}_{A,k} A \mathcal{S}_{A,\ell}, \tilde{T}\big\rangle$ and $
\big\langle \mathcal{S}_{A,k} A \mathbf{\Theta}\mathbf{\Theta}^\top, \tilde{T}\big\rangle$
in Lemmas \ref{lm: minor 1} and \ref{lm : minor 2} by exploiting the low-rank structure of $T$.
Here we discuss improving the bound for 
$
\big\langle \mathcal{S}_{A,1} A \mathcal{S}_{A,1}, \tilde{T}\big\rangle$, 
i.e., the term $I$ in the proof of Lemma~\ref{lm : minor 2}, only, and postpone refining the other terms to future work. If $T$ were not assumed low-rank, one would use $r' \leq d$ in the bounds given in \eqref{eq:termI}--\eqref{eq:termI2}. Under the additional low-rank condition on $T$, $r'\leq r_T$, yielding
\[
I 
\;\lesssim\; 
\frac{\sigma^2 r \sqrt{r_T}\,\|T\|_{\mathrm{F}}}{n \lambda_r}
\;\lesssim\; 
\frac{\sigma^2 \sqrt{d r r_T}\,\|T V\|_{\mathrm{F}}}{n \lambda_r}
\;=\;
\frac{\sigma\bigl(\|U^\top T\|_{\mathrm{F}} + \|T V\|_{\mathrm{F}}\bigr)}{\sqrt{n}}
\,O_p\!\Bigl(\frac{\sigma}{\lambda_r}\sqrt{\frac{d r r_T}{n}}\Bigr).
\]
When $(\sigma_i / \lambda_r)\,\sqrt{d r r_T/n} = o(1)$, the term $I$ is then dominated by the main term in \eqref{eq:19}. A more thorough treatment of the remaining terms is deferred to future work.

\end{document}